\documentclass[fleqn,twoside]{article}
\usepackage{espcrc1}

\newcommand{\AmS}{{\protect\the\textfont2
  A\kern-.1667em\lower.5ex\hbox{M}\kern-.125emS}}
\title{Towards
 Integrated Glance To
Restructuring in Combinatorial Optimization}

\author{Mark Sh. Levin
\address{
 Inst. for Information Transmission Problems, Russian Academy of
 Sciences\\
 19 Bolshoj Karetny Lane, Moscow 127994, Russia\\
 }
\thanks{
  {\it E-mail address}: mslevin@acm.org
  }
 }

\begin{document}

\maketitle

\begin{abstract}
 The paper focuses on
 a new class of combinatorial problems
 which consists in restructuring of solutions
 (as sets/structures)
 in combinatorial optimization.
 Two main features of the restructuring process are examined:
 (i) a cost of the restructuring,
 (ii) a closeness to a goal solution.
  Three types of the restructuring problems are under study:
 (a) one-stage structuring,
 (b) multi-stage structuring, and
 (c)  structuring over changed element set.
 One-criterion and multicriteria problem formulations can be considered.
 The restructuring problems correspond to redesign (improvement, upgrade) of modular
 systems or solutions.
 The restructuring approach is described and illustrated
 (problem statements, solving schemes, examples)
 for the following combinatorial
 optimization problems:
 knapsack problem,
 multiple choice problem,
 assignment problem,
 spanning tree problems,
 clustering problem,
 multicriteria ranking (sorting) problem,
 morphological clique problem.
 Numerical examples
 illustrate the restructuring problems and solving schemes.

~~~~~~~~~~~

 {\it Keywords:}~
               Combinatorial optimization,
               restructuring,
               multicriteria decision making,
               framework, heuristics,
               artificial intelligence,
               knapsack problem,
                multiple choice problem,
                assignment problem,
               spanning trees,
              clustering,
               sorting problem,
               clique,
               applications.

\vspace{1pc}
\end{abstract}

\newcounter{cms}
\setlength{\unitlength}{1mm}

\tableofcontents

\section{Introduction}

 In recent decades, the following basic development directions for
 basic combinatorial optimization formulations have been  studied
 (Fig. 1):
  (i) multicriteria problem formulations
  (e.g., \cite{coello07,deb09,ehr10,talbi09,tk06}),
  (ii) problems under uncertainty
  (fuzzy combinatorial optimization problems, etc.)
  (e.g., \cite{lev15,lod10,pard13,yang07,yem08}),
  (iii) problems in dynamic environments
 (e.g., \cite{boj07,mor04,yang07,yang12}),
  and
  online  problems
  (e.g., \cite{albers03,albers99,borod98,hen09}).
 Evidently, the above-mentioned problem development
 directions can have intersections
 (e.g., multicriteria online problem under uncertainty).

\begin{center}
\begin{picture}(112,61)

\put(07,00){\makebox(0,0)[bl]{Fig. 1.
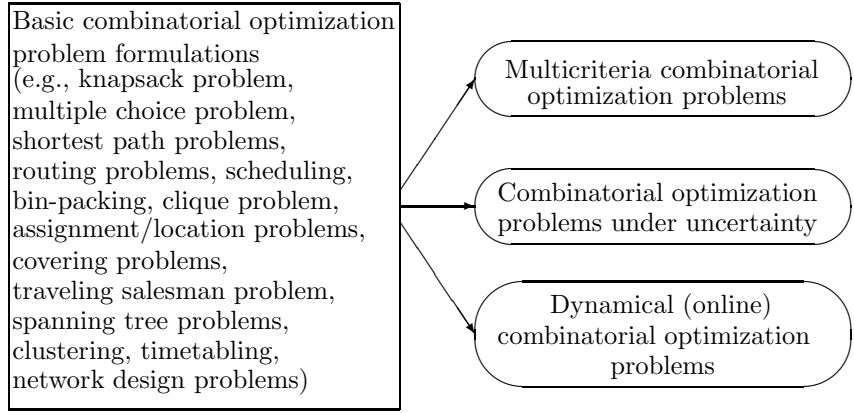
 Basic development directions for
 problem formulations}}


\put(00,06){\line(1,0){52}} \put(00,60){\line(1,0){52}}
\put(00,06){\line(0,1){54}} \put(52,06){\line(0,1){54}}

\put(0.5,56){\makebox(0,0)[bl]{Basic combinatorial optimization}}
\put(0.5,51.6){\makebox(0,0)[bl]{problem formulations}}

\put(0.5,48){\makebox(0,0)[bl]{(e.g., knapsack problem, }}
\put(0.5,44){\makebox(0,0)[bl]{multiple choice problem,}}
\put(0.5,40){\makebox(0,0)[bl]{shortest path problems,}}

\put(0.5,36){\makebox(0,0)[bl]{routing problems, scheduling,}}

\put(0.5,32){\makebox(0,0)[bl]{bin-packing, clique problem,}}

\put(0.5,28){\makebox(0,0)[bl]{assignment/location problems,}}

\put(0.5,24){\makebox(0,0)[bl]{covering problems,}}

\put(0.5,20){\makebox(0,0)[bl]{traveling salesman problem,}}

\put(0.5,16){\makebox(0,0)[bl]{spanning tree problems,}}
\put(0.5,12){\makebox(0,0)[bl]{clustering, timetabling,}}
\put(0.5,08){\makebox(0,0)[bl]{network design problems)}}


\put(52,33){\vector(1,0){10}}

\put(52,35){\vector(2,3){10}}

\put(52,31){\vector(2,-3){10}}


\put(87,50){\oval(50,10)}

\put(66,50){\makebox(0,0)[bl]{Multicriteria combinatorial}}
\put(69,46){\makebox(0,0)[bl]{optimization problems}}


\put(87,33){\oval(50,10)}

\put(65,33){\makebox(0,0)[bl]{Combinatorial optimization }}
\put(65,29){\makebox(0,0)[bl]{problems under uncertainty}}


\put(87,16){\oval(50,14)}

\put(72,18){\makebox(0,0)[bl]{Dynamical (online)}}
\put(65,14){\makebox(0,0)[bl]{combinatorial optimization}}
\put(80,10){\makebox(0,0)[bl]{problems}}

\end{picture}
\end{center}

 In this paper, combinatorial optimization problems with
 modifications of problem solutions are examined
 as a special new problem class.
 Generally, the following basic approaches for
 changing some solutions in combinatorial optimization problems
 are considered (Table 1):
 (1) modification of solution(s) as relinking, reassignment/relocation,
  rescheduling, repositioning, etc.
  (including editing problems, network modification/restructuring);
  (2) reoptimization (modification of a solution by a set of small
 change operations
   to improve of the solution objective function(s));
 (3) augmentation-type problems (addition/correction of solution
 components to obtain required solution properties);
 (4) restructuring (modification of a solution by set of change
  operations while taking into account two objectives/constraints:
 cost of the change operations and proximity to an optimal
 solution at the next time stage);
 (5) dynamic combinatorial optimization problems
 (including
  online problems, problems with changing requests); and
 (6) design of multistage dynamic restructuring trajectories for problem
 solution(s).

\begin{center}
 {\bf Table 1.} Basic types of reoptimization/restructuring approaches in combinatorial optimization   \\
\begin{tabular}{| c | l | l |}
\hline
 No.  & Direction  & Source \\
\hline

 1.& Modification of solution/structure (reassignment, relinking, &\\
 &rescheduling, repositioning, editing, recoloring, etc.)
 & \cite{aiex05,berm81,bose01,ho14,moran08,reeves98,vos96,zhu12}\\

 2.& Reoptimization (small correction of solution to improve its
 &\cite{archetti03,aus09,bilo08,boria10,esc09}\\

  & quality as improvement of the solution objective function(s)):&\\

 3.&Augmentation-type problems (addition/correction of solution
    &\cite{bock11,dam10,dehne06,esw76,guna13,guo09,mann10} \\
 &components to obtain required solution properties)   &\\

 4.& Restructuring problems (modification of solution while taking & \cite{lev11restr,lev15c}\\
 &into account two criteria: minimum modification cost,   &\\
 &minimum proximity to a next solution at the next time stage)&\\

 5.&Dynamic combinatorial optimization problems (including  &\cite{adib10,exp14,kho12,roh09,wzorek06}\\
  &online problems, problems with changing requests &\\

 6.& Design of multistage dynamic restructuring trajectories for &\\
  & problem  solution(s) &\cite{lev15c}\\

\hline
\end{tabular}
\end{center}


 This paper addresses
 a class of restructuring combinatorial problems.
 The examined restructuring problems correspond to
 redesign/reconfiguration (improvement, upgrade)
 of modular systems and
 the situations can be faced
 in many applied domains
 (e.g., complex software, algorithm systems,
 communication networks, computer networks,
 information systems, manufacturing systems,
  control systems, constructions)
     (e.g.,
  \cite{arain08,bi08,bon02,lev06,lev09,lev10a,levdan05,levsaf11,levand11,nolt99,pollock92,qui11}).
  In basic (one-stage) restructuring problem,
   an optimization problem is solved for two time moments:
 \(\tau_{0}\) and  \(\tau_{1}\) to obtain corresponding solutions
 \(S^{0}\) and \(S^{1}\).
 The problem consists in
 a ``cheap'' transformation (change) of solution \(S^{0}\) to a solution \(S^{*}\) that
 is very close to \(S^{1}\).
 Generally, the following restructuring problem types are examined:
 (i) basic one-stage restructuring  problem,
 (ii) multi-stage restructuring problem,
 (iii) restructuring over changed element set.
 The restructuring approach is described and illustrated for the following combinatorial
 optimization problems
 (e.g., \cite{gar79,lev09}):
%
 knapsack problem,
  multiple choice problem,
  assignment problem,
  spanning trees problems,
  clustering,
  sorting problem,
  morphological clique problem.

 Here, the following restructuring problem statement classification parameters are considered:
 ~(1) time-based problem type:
 (a) one-stage problems,
 (b) multi-stage problem;
 ~(2) types of criteria and/or estimates:
 (i) basic type,
 (ii) multicriteria problem,
 (iii) ordinal (or multiset-based) estimates.
 Numerical examples
 illustrate the restructuring processes.
 Some preliminary materials for the article were published
 in
  \cite{lev11restr,lev15,lev15c}.

\section{Modification problems types in combinatorial optimization}

 Modification of problem solutions is a well-known traditional technique
 for improvement/modification
 and is widely used in various heuristics,
 e.g., local optimization)
 (e.g.,
  \cite{aiex05,berm81,bose01,ho14,moran08,reeves98,vos96,zhu12}).


 In recent years, several combinatorial optimization problems have
 been examined under the reoptimization process (Fig. 2), for example:
 (i) travelling salesman problem (\cite{archetti03}),
 (ii) scheduling \cite{scha97},
 (iii) knapsack problem  \cite{arch06},
 (iv) shortest common superstring problems \cite{bilo11},
 (v) weighted graph and covering problems \cite{bilo08a},
 (vi) spanning tree problems \cite{qui11},
 and
 (vii) Steiner tree problems
 \cite{esc09}.

\begin{center}
\begin{picture}(78,49)
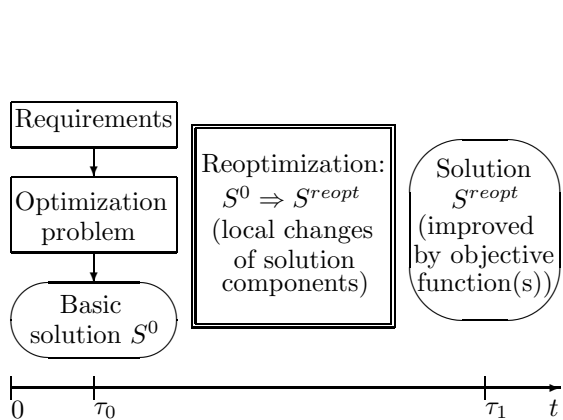
\put(1.2,00){\makebox(0,0)[bl]{Fig. 2. Framework for
 reoptimization  process}}

\put(00,09){\vector(1,0){73}}

\put(00,7.5){\line(0,1){3}} \put(11,7.5){\line(0,1){3}}
\put(63,7.5){\line(0,1){3}}

\put(00,05){\makebox(0,0)[bl]{\(0\)}}
\put(11,05){\makebox(0,0)[bl]{\(\tau_{0}\)}}
\put(63,05){\makebox(0,0)[bl]{\(\tau_{1}\)}}

\put(71.5,05.3){\makebox(0,0)[bl]{\(t\)}}


\put(00,41){\line(1,0){22}} \put(00,47){\line(1,0){22}}
\put(00,41){\line(0,1){06}} \put(22,41){\line(0,1){06}}

\put(0.5,43){\makebox(0,0)[bl]{Requirements}}


\put(11,41){\vector(0,-1){4}}

\put(00,27){\line(1,0){22}} \put(00,37){\line(1,0){22}}
\put(00,27){\line(0,1){10}} \put(22,27){\line(0,1){10}}

\put(0.5,32){\makebox(0,0)[bl]{Optimization}}
\put(4,28){\makebox(0,0)[bl]{problem}}


\put(11,27){\vector(0,-1){4}}

\put(11,18){\oval(22,10)}

\put(06.5,19){\makebox(0,0)[bl]{Basic}}
\put(02.5,15){\makebox(0,0)[bl]{solution \(S^{0}\)}}


\put(24,17){\line(1,0){27}} \put(24,44){\line(1,0){27}}
\put(24,17){\line(0,1){27}} \put(51,17){\line(0,1){27}}

\put(24.5,17.5){\line(1,0){26}} \put(24.5,43.5){\line(1,0){26}}
\put(24.5,17.5){\line(0,1){26}} \put(50.5,17.5){\line(0,1){26}}

\put(25.3,37){\makebox(0,0)[bl]{Reoptimization:}}
\put(28,33){\makebox(0,0)[bl]
 {\(S^{0} \Rightarrow S^{reopt} \)}}

\put(26.6,28){\makebox(0,0)[bl]{(local changes}}
\put(29.6,25){\makebox(0,0)[bl]{of solution}}
\put(28,21){\makebox(0,0)[bl]{components)}}


\put(63,30){\oval(20,24)}

\put(56.5,37){\makebox(0,0)[bl]{Solution}}
\put(58.5,33){\makebox(0,0)[bl]{\(S^{reopt}\)}}
\put(54,28.6){\makebox(0,0)[bl]{(improved}}
\put(53.5,25){\makebox(0,0)[bl]{by objective}}
\put(54,21){\makebox(0,0)[bl]{function(s))}}

\end{picture}
%
\begin{picture}(76,61)

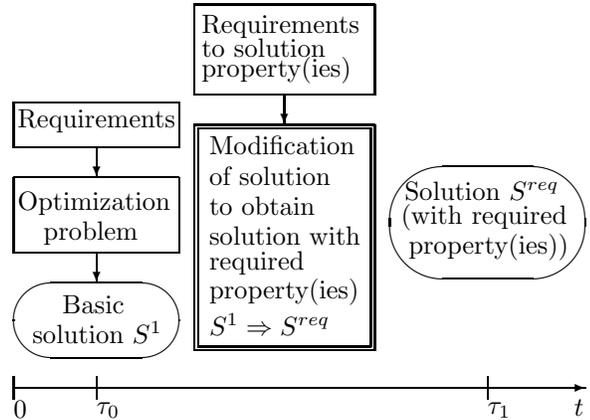
\put(02,00){\makebox(0,0)[bl]{Fig. 3. Framework for
 augmentation problem }}

\put(00,09){\vector(1,0){76}}

\put(00,7.5){\line(0,1){3}} \put(11,7.5){\line(0,1){3}}
\put(63,7.5){\line(0,1){3}}

\put(00,05){\makebox(0,0)[bl]{\(0\)}}
\put(11,05){\makebox(0,0)[bl]{\(\tau_{0}\)}}
\put(63,05){\makebox(0,0)[bl]{\(\tau_{1}\)}}

\put(74.5,05.3){\makebox(0,0)[bl]{\(t\)}}


\put(00,41){\line(1,0){22}} \put(00,47){\line(1,0){22}}
\put(00,41){\line(0,1){06}} \put(22,41){\line(0,1){06}}

\put(0.5,43){\makebox(0,0)[bl]{Requirements}}


\put(11,41){\vector(0,-1){4}}

\put(00,27){\line(1,0){22}} \put(00,37){\line(1,0){22}}
\put(00,27){\line(0,1){10}} \put(22,27){\line(0,1){10}}

\put(0.5,32){\makebox(0,0)[bl]{Optimization}}
\put(4,28){\makebox(0,0)[bl]{problem}}


\put(11,27){\vector(0,-1){4}}

\put(11,18){\oval(22,10)}

\put(06.5,19){\makebox(0,0)[bl]{Basic}}
\put(02.5,15){\makebox(0,0)[bl]{solution \(S^{1}\)}}


\put(24,48){\line(1,0){24}} \put(24,60){\line(1,0){24}}
\put(24,48){\line(0,1){12}} \put(48,48){\line(0,1){12}}

\put(25,55.5){\makebox(0,0)[bl]{Requirements}}
\put(25,53){\makebox(0,0)[bl]{to solution}}
\put(25,49.5){\makebox(0,0)[bl]{property(ies)}}

\put(36,48){\vector(0,-1){4}}


\put(24,14){\line(1,0){24}} \put(24,44){\line(1,0){24}}
\put(24,14){\line(0,1){30}} \put(48,14){\line(0,1){30}}

\put(24.5,14.5){\line(1,0){23}} \put(24.5,43.5){\line(1,0){23}}
\put(24.5,14.5){\line(0,1){29}} \put(47.5,14.5){\line(0,1){29}}

\put(26,40){\makebox(0,0)[bl]{Modification}}
\put(26,36){\makebox(0,0)[bl]{of solution}}
\put(26,32){\makebox(0,0)[bl]{to obtain }}
\put(26,28){\makebox(0,0)[bl]{solution with}}
\put(26,24){\makebox(0,0)[bl]{required }}
\put(26,20){\makebox(0,0)[bl]{property(ies)}}

\put(26,16){\makebox(0,0)[bl]{\(S^{1} \Rightarrow S^{req}  \)}}


\put(63,31){\oval(26,15)}

\put(52,34){\makebox(0,0)[bl]{Solution \(S^{req}\)}}
\put(51.5,30){\makebox(0,0)[bl]{(with required}}
\put(52.5,26){\makebox(0,0)[bl]{property(ies)) }}

\end{picture}
\end{center}

 The reoptimization problem describes the following scenario
 (Fig. 2):~

~~

 Given an instance
 of an optimization problem together with an optimal solution for
 it, we want to find a good solution for a locally modified
 instance (addition or removing links, etc.)
 (e.g., \cite{bilo11}).

~~

 Thus, reoptimization problems above
 are targeted to an improvement
 (``post-optimization'')
 of an obtained solution.
 Usually, the reoptimization problems are NP-hard \cite{bock08a}.
 In some simplified versions of reoptimization problems
 polynomial approximation schemes
 (PTAS) have been designed
 (e.g., \cite{bilo08a}).
 Evidently, the reoptimization approach
 is a contemporary step in the study of the
 problem solution modification processes.


 Augmentation problems are targeted to obtaining solution(s)
 with some required properties (Fig. 3), for example:
 (a) a required level of network connectivity in network topology design
 (e.g., bi-connected network, \(k\)-connected network
   \cite{bock11,dam10,dehne06,esw76,guna13,guo09,mann10});
 (b) a required structure type for the obtained graph/network
 (e.g.,
  a set of cliques/quasi-cliques
   \cite{bock11,dam10,dehne06,esw76,guna13,guo09,mann10},
   a tree/hierarchy with required property(ies)).


  Reload cost problems (and close changeover cost problems)
 are targeted to find a new structure
(e.g., paths, spanning trees, schedules, networks)
 with respect to reload costs
 \cite{amal11,gam12,goz14,wirth01}.


 Restructuring combinatorial problems
  are targeted to
 restructuring of an initial solution
 (e.g., a set of elements, a structure)
 in combinatorial optimization
 to obtain a new solution that is very close
 to a goal solution while taking into account
 a ``cheap'' modification of the initial solution.
 Here, our problem statement is described for  basic
 one-criterion and multicriteria problem formulations
 which are significant for real applications in dynamical environments.
 Two main features of the restructuring process are examined:
 (i) a cost of the initial problem solution restructuring,
 (ii) a closeness of the obtained restructured solution to a goal solution
 (the cost of restructuring and/or
  closeness to the goal solution may be used as  vector-like functions).
 Illustrations for one-stage restructuring problem are depicted in
 Fig. 4 and Fig. 5.

\begin{center}
\begin{picture}(80,52)

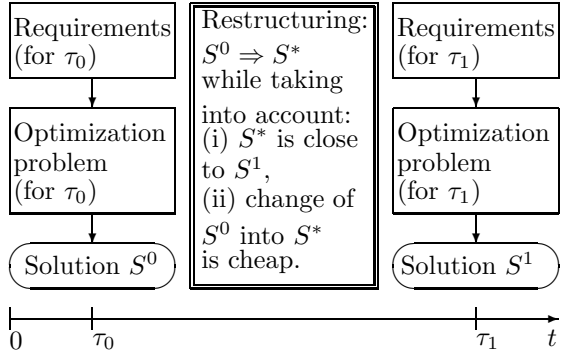
\put(01,00){\makebox(0,0)[bl]{Fig. 4. Framework for
 one-stage restructuring
 }}

\put(00,09){\vector(1,0){73}}

\put(00,7.5){\line(0,1){3}} \put(11,7.5){\line(0,1){3}}
\put(62,7.5){\line(0,1){3}}

\put(00,05){\makebox(0,0)[bl]{\(0\)}}
\put(11,05){\makebox(0,0)[bl]{\(\tau_{0}\)}}
\put(62,05){\makebox(0,0)[bl]{\(\tau_{1}\)}}

\put(71.5,05.3){\makebox(0,0)[bl]{\(t\)}}


\put(00,41){\line(1,0){22}} \put(00,51){\line(1,0){22}}
\put(00,41){\line(0,1){10}} \put(22,41){\line(0,1){10}}

\put(0.5,46){\makebox(0,0)[bl]{Requirements}}
\put(0.5,42){\makebox(0,0)[bl]{(for \(\tau_{0}\))}}


\put(11,41){\vector(0,-1){4}}

\put(00,23){\line(1,0){22}} \put(00,37){\line(1,0){22}}
\put(00,23){\line(0,1){14}} \put(22,23){\line(0,1){14}}

\put(0.5,32){\makebox(0,0)[bl]{Optimization}}
\put(0.5,28){\makebox(0,0)[bl]{problem}}
\put(0.5,24){\makebox(0,0)[bl]{(for \(\tau_{0})\)}}


\put(11,23){\vector(0,-1){4}}

\put(11,16){\oval(22,06)}

\put(02,15){\makebox(0,0)[bl]{Solution \(S^{0}\)}}


\put(24,13){\line(1,0){25}} \put(24,51){\line(1,0){25}}
\put(24,13){\line(0,1){38}} \put(49,13){\line(0,1){38}}

\put(24.5,13.5){\line(1,0){24}} \put(24.5,50.5){\line(1,0){24}}
\put(24.5,13.5){\line(0,1){37}} \put(48.5,13.5){\line(0,1){37}}

\put(25.5,47){\makebox(0,0)[bl]{Restructuring:}}
\put(25.5,43){\makebox(0,0)[bl]{\(S^{0} \Rightarrow S^{*}  \)}}
\put(25.5,39){\makebox(0,0)[bl]{while taking}}
\put(25.5,35){\makebox(0,0)[bl]{into account:}}
\put(25.5,31){\makebox(0,0)[bl]{(i) \(S^{*}\) is close}}
\put(25.5,27){\makebox(0,0)[bl]{to \(S^{1}\),}}
\put(25.5,23){\makebox(0,0)[bl]{(ii) change of }}
\put(25.5,19){\makebox(0,0)[bl]{\(S^{0}\) into \(S^{*}\)}}
\put(25.5,15){\makebox(0,0)[bl]{is cheap.}}


\put(51,41){\line(1,0){22}} \put(51,51){\line(1,0){22}}
\put(51,41){\line(0,1){10}} \put(73,41){\line(0,1){10}}

\put(51.5,46){\makebox(0,0)[bl]{Requirements}}
\put(51.5,42){\makebox(0,0)[bl]{(for \(\tau_{1}\))}}


\put(62,41){\vector(0,-1){4}}

\put(51,23){\line(1,0){22}} \put(51,37){\line(1,0){22}}
\put(51,23){\line(0,1){14}} \put(73,23){\line(0,1){14}}

\put(51.5,32){\makebox(0,0)[bl]{Optimization}}
\put(51.5,28){\makebox(0,0)[bl]{problem }}
\put(51.5,24){\makebox(0,0)[bl]{(for \(\tau_{1})\)}}


\put(62,23){\vector(0,-1){4}}

\put(62,16){\oval(22,06)}

\put(52,15){\makebox(0,0)[bl]{Solution \(S^{1}\)}}

\end{picture}
%
\begin{picture}(67,47)

\put(02.1,00){\makebox(0,0)[bl]{Fig. 5. One-stage restructuring
problem}}


\put(00,05){\vector(1,0){67}}

\put(65.5,06.3){\makebox(0,0)[bl]{\(t\)}}


\put(0.6,18){\makebox(0,0)[bl]{\(S^{0}\)}}
\put(3.5,17){\circle{1.6}}

\put(10,13){\makebox(0,0)[bl]{Initial}}
\put(10,10){\makebox(0,0)[bl]{solution}}
\put(10,06){\makebox(0,0)[bl]{(\(t=\tau^{0}\))}}

\put(09.5,12){\line(-4,3){5}}

\put(5,17.9){\vector(2,1){23.5}} \put(5,17.5){\vector(2,1){23.5}}


\put(40,35){\circle*{2.7}}

\put(54,39){\makebox(0,0)[bl]{Goal}}
\put(54,36){\makebox(0,0)[bl]{solution}}
\put(53.5,32){\makebox(0,0)[bl]{(\(t=\tau^{1}\)):}}
\put(57.5,29){\makebox(0,0)[bl]{\(S^{1}\)}}

\put(53,35){\line(-1,0){10.5}}

\put(40,35){\oval(12,10)} \put(40,35){\oval(17,17)}
\put(40,35){\oval(24,22)}


\put(40,35){\vector(-2,-1){9}} \put(30,30){\vector(2,1){9}}



\put(25,17){\makebox(0,0)[bl]{Proximity}}
\put(25,13){\makebox(0,0)[bl]{~\(\rho (S^{*},S^{1})\)}}

\put(36,20){\line(0,1){12.4}}


\put(43,19){\makebox(0,0)[bl]{Neighborhoods }}
\put(49,16){\makebox(0,0)[bl]{of ~\(S^{1}\)}}

\put(56,22){\line(-1,1){10}}

\put(53,22){\line(-2,1){8}}


\put(30,30){\circle{2}} \put(30,30){\circle*{1}}

\put(10,42.5){\makebox(0,0)[bl]{Obtained}}
\put(10,39.5){\makebox(0,0)[bl]{solution \(S^{*}\)}}

\put(21,39){\line(1,-1){8}}

\put(1,33.5){\makebox(0,0)[bl]{Solution }}
\put(1,29.5){\makebox(0,0)[bl]{change cost }}
\put(1,25.5){\makebox(0,0)[bl]{\(H(S^{0} \rightarrow S^{*})\)}}

\put(8,26){\line(1,-1){4}}


\end{picture}
\end{center}

 A brief description of a formal statement for the  restructuring problem
 is the following.
 Let \(P\) be a combinatorial optimization problem with a solution as
 structure
 \(S\)
 (i.e., subset, graph),
 \(\Omega\) be initial data (elements, element parameters, etc.),
 \(f(P)\) be objective function(s).
 Thus, \(S(\Omega)\) be a solution for initial data \(\Omega\),
 \(f(S(\Omega))\) be the corresponding objective function.
 Let \(\Omega^{0}\) be initial data at an initial stage,
  \(f(S(\Omega^{0}))\) be the corresponding objective function.
 \(\Omega^{1}\) be initial data at next stage,
  \(f(S(\Omega^{1}))\) be the corresponding objective function.
 As a result,
 the following solutions can be considered:

 (a) \( S^{0}=S(\Omega^{0})\) with \(f(S(\Omega^{0}))\) and
 (b) \( S^{1}=S(\Omega^{1})\) with \(f(S(\Omega^{1}))\).

 In addition it is reasonable to examine a cost of changing
 a solution into another one:~
 \( H(S^{\alpha} \rightarrow  S^{\beta})\).
 Let \(\rho ( S^{\alpha}, S^{\beta} )\) be a proximity between
 solutions
  \( S^{\alpha}\) and \( S^{\beta}\),
  for example,
 \(\rho ( S^{\alpha}, S^{\beta} ) = | f(S^{\alpha}) -  f(S^{\beta}) |\).
 Note function \(f(S)\) is often a vector function.
 Finally, the restructuring problem can be examine as follows (a basic version):

~~

 Find a solution \( S^{*}\) while taking into account the
 following:

  (i) \( H(S^{0} \rightarrow  S^{*}) \rightarrow \min \),
%
  ~(ii) \(\rho ( S^{*}, S^{1} )  \rightarrow \min  \) ~(or constraint).

~~


 Dynamic problems (including online problems,
  problems with changing requests)
 (i.e., while taking into account dynamically changing environment)
 are illustrated in Fig. 6.
 Here new requirements
 are obtaining in online mode
 and
 it is necessary to resolve the problem
 at each time moment
 (e.g.,
 \cite{adib10,exp14,kho12,roh09,wzorek06}).
 In Fig. 6, the resultant solution trajectory is:~
 \(\widehat{S} = <S^{0} \rightarrow S^{1} \rightarrow  S^{2} \rightarrow ...  >.\)

\begin{center}
\begin{picture}(122,51)

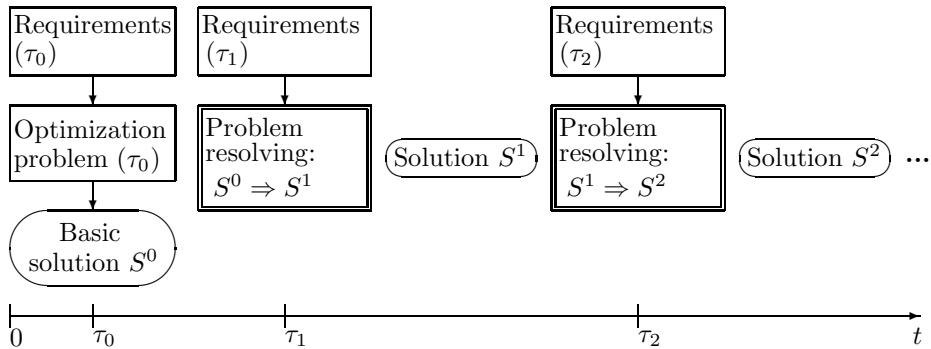
\put(09.3,00){\makebox(0,0)[bl]{Fig. 6. Illustration for dynamic
 (online) problem solving processes}}

\put(00,09){\vector(1,0){121}}

\put(00,7.5){\line(0,1){3}} \put(11,7.5){\line(0,1){3}}
\put(36.5,7.5){\line(0,1){3}}

\put(83.5,7.5){\line(0,1){3}}

\put(00,05){\makebox(0,0)[bl]{\(0\)}}
\put(11,05){\makebox(0,0)[bl]{\(\tau_{0}\)}}
\put(36.5,05){\makebox(0,0)[bl]{\(\tau_{1}\)}}
\put(83.5,05){\makebox(0,0)[bl]{\(\tau_{2}\)}}

\put(120,05.3){\makebox(0,0)[bl]{\(t\)}}


\put(00,41){\line(1,0){22}} \put(00,50){\line(1,0){22}}
\put(00,41){\line(0,1){09}} \put(22,41){\line(0,1){09}}

\put(0.5,46){\makebox(0,0)[bl]{Requirements}}
\put(0.5,42){\makebox(0,0)[bl]{(\(\tau_{0}\))}}


\put(11,41){\vector(0,-1){4}}

\put(00,27){\line(1,0){22}} \put(00,37){\line(1,0){22}}
\put(00,27){\line(0,1){10}} \put(22,27){\line(0,1){10}}

\put(0.5,32){\makebox(0,0)[bl]{Optimization}}
\put(0.5,28){\makebox(0,0)[bl]{problem (\(\tau_{0}\))}}


\put(11,27){\vector(0,-1){4}}

\put(11,18){\oval(22,10)}

\put(06.5,19){\makebox(0,0)[bl]{Basic}}
\put(02.5,15){\makebox(0,0)[bl]{solution \(S^{0}\)}}


\put(25,41){\line(1,0){23}} \put(25,50){\line(1,0){23}}
\put(25,41){\line(0,1){09}} \put(48,41){\line(0,1){09}}

\put(25.5,46){\makebox(0,0)[bl]{Requirements }}
\put(26,42){\makebox(0,0)[bl]{(\(\tau_{1}\))}}

\put(36.5,41){\vector(0,-1){4}}


\put(25,23){\line(1,0){23}} \put(25,37){\line(1,0){23}}
\put(25,23){\line(0,1){14}} \put(48,23){\line(0,1){14}}

\put(25.5,23.5){\line(1,0){22}} \put(25.5,36.5){\line(1,0){22}}
\put(25.5,23.5){\line(0,1){13}} \put(47.5,23.5){\line(0,1){13}}

\put(26,33){\makebox(0,0)[bl]{Problem }}
\put(26,29){\makebox(0,0)[bl]{resolving:}}
\put(27,25){\makebox(0,0)[bl]{\(S^{0} \Rightarrow S^{1}  \)}}


\put(60,30){\oval(20,05)}

\put(51,29){\makebox(0,0)[bl]{Solution \(S^{1}\)}}


\put(72,41){\line(1,0){23}} \put(72,50){\line(1,0){23}}
\put(72,41){\line(0,1){09}} \put(95,41){\line(0,1){09}}

\put(72.5,46){\makebox(0,0)[bl]{Requirements }}
\put(73,42){\makebox(0,0)[bl]{(\(\tau_{2}\))}}

\put(83.5,41){\vector(0,-1){4}}


\put(72,23){\line(1,0){23}} \put(72,37){\line(1,0){23}}
\put(72,23){\line(0,1){14}} \put(95,23){\line(0,1){14}}

\put(72.5,23.5){\line(1,0){22}} \put(72.5,36.5){\line(1,0){22}}
\put(72.5,23.5){\line(0,1){13}} \put(94.5,23.5){\line(0,1){13}}

\put(73,33){\makebox(0,0)[bl]{Problem }}
\put(73,29){\makebox(0,0)[bl]{resolving:}}
\put(74,25){\makebox(0,0)[bl]{\(S^{1} \Rightarrow S^{2}  \)}}


\put(107,30){\oval(20,05)}

\put(98,29){\makebox(0,0)[bl]{Solution \(S^{2}\)}}

\put(119,29.5){\makebox(0,0)[bl]{{\bf ...}}}

\end{picture}
\end{center}

 Fig. 7 illustrates a simplified general version of dynamic clustering
 (the scheme is similar to case-based reasoning)
 (e.g., \cite{jay14}).

\begin{center}
\begin{picture}(112,42)

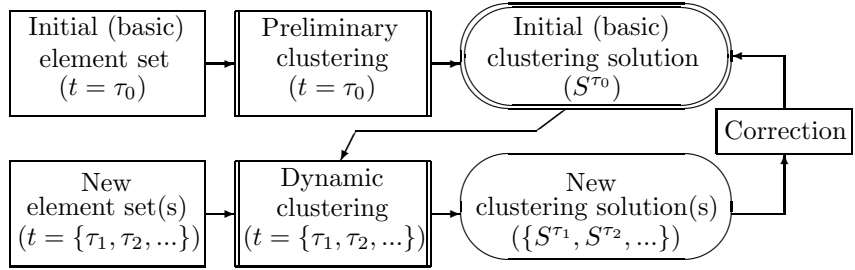
\put(016.5,00){\makebox(0,0)[bl]{Fig. 7. Scheme of dynamic
  clustering process}}


\put(00,26){\line(1,0){26}} \put(00,40){\line(1,0){26}}
\put(00,26){\line(0,1){14}} \put(26,26){\line(0,1){14}}

\put(02.5,36){\makebox(0,0)[bl]{Initial (basic)}}
\put(04,32.5){\makebox(0,0)[bl]{element set}}
\put(06.5,28){\makebox(0,0)[bl]{(\(t= \tau_{0}\))}}

\put(26,33){\vector(1,0){04}}


\put(30,26){\line(1,0){26}} \put(30,40){\line(1,0){26}}
\put(30,26){\line(0,1){14}} \put(56,26){\line(0,1){14}}
\put(30.5,26){\line(0,1){14}} \put(55.5,26){\line(0,1){14}}

\put(33.5,36){\makebox(0,0)[bl]{Preliminary}}
\put(35.5,32){\makebox(0,0)[bl]{clustering}}
\put(37,28){\makebox(0,0)[bl]{(\(t= \tau_{0}\))}}

\put(56,33){\vector(1,0){04}}


\put(78,34){\oval(36,14)} \put(78,34){\oval(35,13)}

\put(67,36){\makebox(0,0)[bl]{Initial (basic)}}
\put(63.5,32){\makebox(0,0)[bl]{clustering solution}}
\put(73.5,28){\makebox(0,0)[bl]{(\( S^{\tau_{0}}\)) }}

\put(74,27){\line(-4,-3){4}} \put(70,24){\line(-1,0){24}}
\put(46,24){\vector(-1,-2){2}}



\put(00,06){\line(1,0){26}} \put(00,20){\line(1,0){26}}
\put(00,06){\line(0,1){14}} \put(26,06){\line(0,1){14}}

\put(09.5,16){\makebox(0,0)[bl]{New}}
\put(02.3,12){\makebox(0,0)[bl]{element set(s)}}
\put(01,08){\makebox(0,0)[bl]{(\(t= \{\tau_{1},\tau_{2},...
 \}\))}}

\put(26,13){\vector(1,0){04}}


\put(30,06){\line(1,0){26}} \put(30,20){\line(1,0){26}}
\put(30,06){\line(0,1){14}} \put(56,06){\line(0,1){14}}
\put(30.5,06){\line(0,1){14}} \put(55.5,06){\line(0,1){14}}

\put(36,16){\makebox(0,0)[bl]{Dynamic}}
\put(35.5,12){\makebox(0,0)[bl]{clustering}}
\put(31,08){\makebox(0,0)[bl]{(\(t= \{\tau_{1},\tau_{2},...
 \}\))}}

\put(56,13){\vector(1,0){04}}


\put(78,14){\oval(36,14)}

\put(74,16){\makebox(0,0)[bl]{New}}
\put(62,12){\makebox(0,0)[bl]{clustering solution(s)}}
\put(66.6,08){\makebox(0,0)[bl]{(\(\{S^{\tau_{1}}, S^{\tau_{2}},
  ... \}\))}}


\put(94,21){\line(1,0){18}} \put(94,27){\line(1,0){18}}
\put(94,21){\line(0,1){6}} \put(112,21){\line(0,1){6}}

\put(95,23){\makebox(0,0)[bl]{Correction}}

\put(96,13){\line(1,0){7}} \put(103,13){\vector(0,1){08}}

\put(103,27){\line(0,1){7}} \put(103,34){\vector(-1,0){07}}

\end{picture}
\end{center}

 In multi-stage restructuring problems,
 a solution trajectory is designed (Fig. 8, Fig. 9).
 Thus, two  trajectories are examined:

 (a) \(n\)-stage trajectory of optimal solutions:~
 \(\overline{S}^{opt} =
 <S^{0} \rightarrow S^{1} \rightarrow  S^{2} \rightarrow ... \rightarrow  S^{n} >\),

 (b) \(n\)-stage trajectory of restructured solutions:~
 \(\overline{S}^{rest} =
 <S^{0} \rightarrow S^{1*} \rightarrow  S^{2*} \rightarrow ... \rightarrow S^{n*}>\).

 Here, the restructuring problem can be examine as follows (a basic version):

~~

 Find a solution \( \overline{S}^{rest}\) while taking into account the
 following:

  (i) \( \overline{H} (\overline{S}^{rest} \rightarrow  \overline{S}^{opt}) \rightarrow \min  \),
%
  ~(ii) \( \overline{\rho} ( \overline{S}^{rest}, \overline{S}^{opt} )  \rightarrow \min  \)
   ~(or constraint),

 where

  \( \overline{H} (\overline{S}^{rest} \rightarrow\overline{S}^{opt})
  =(
   H (S^{0} \rightarrow S^{1*}),
   H (S^{1*} \rightarrow S^{2*}),
   ...,
   H (S^{(n-1)*} \rightarrow S^{n*}))
   \),

 \(\overline{\rho} ( \overline{S}^{rest}, \overline{S}^{opt} )
  =(
   \rho (S^{1*}, S^{1}),
   \rho (S^{2*}, S^{2*}),
   ...,
    \rho (S^{(n*}, S^{n}))
   \).

~~

 Note,
 minimization (maximization) of a vector function
  corresponds to searching for Pareto-efficient solutions.
 The corresponding optimization model can be examined as the following:

%
   \[\min \overline{\rho} ( \overline{S}^{rest}, \overline{S}^{opt} ) ~~~~ s.t.
 ~~~ \overline{H} (\overline{S}^{rest} \rightarrow  \overline{S}^{opt})  \leq \widehat{\overline{h}}, \]
 where
 \(\widehat{\overline{h}} = (\widehat{h}_{1}, \widehat{h}_{2},...,\widehat{h}_{n},)\)
 is a set (vector) of constraints for costs of the solution changes
 (i.e., a vector component corresponds to each stage).

\begin{center}
\begin{picture}(78,63)

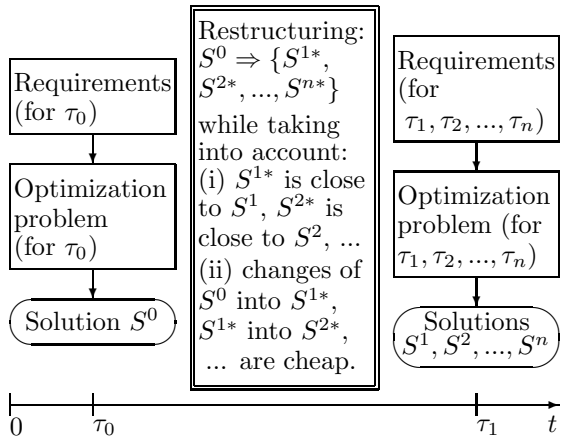
\put(02,00){\makebox(0,0)[bl]{Fig. 8. Framework for
 \(n\)-stage restructuring}}

\put(00,09){\vector(1,0){73}}

\put(00,7.5){\line(0,1){3}} \put(11,7.5){\line(0,1){3}}
\put(62,7.5){\line(0,1){3}}

\put(00,05){\makebox(0,0)[bl]{\(0\)}}
\put(11,05){\makebox(0,0)[bl]{\(\tau_{0}\)}}
\put(62,05){\makebox(0,0)[bl]{\(\tau_{1}\)}}

\put(71.5,05.3){\makebox(0,0)[bl]{\(t\)}}


\put(00,45){\line(1,0){22}} \put(00,55){\line(1,0){22}}
\put(00,45){\line(0,1){10}} \put(22,45){\line(0,1){10}}

\put(0.5,50){\makebox(0,0)[bl]{Requirements}}
\put(0.5,46){\makebox(0,0)[bl]{(for \(\tau_{0}\))}}


\put(11,45){\vector(0,-1){4}}

\put(00,27){\line(1,0){22}} \put(00,41){\line(1,0){22}}
\put(00,27){\line(0,1){14}} \put(22,27){\line(0,1){14}}

\put(0.5,36){\makebox(0,0)[bl]{Optimization}}
\put(0.5,32){\makebox(0,0)[bl]{problem}}
\put(0.5,28){\makebox(0,0)[bl]{(for \(\tau_{0})\)}}


\put(11,27){\vector(0,-1){4}}

\put(11,20){\oval(22,06)}

\put(02,19){\makebox(0,0)[bl]{Solution \(S^{0}\)}}


\put(24,11){\line(1,0){25}} \put(24,62){\line(1,0){25}}
\put(24,11){\line(0,1){51}} \put(49,11){\line(0,1){51}}

\put(24.5,11.5){\line(1,0){24}} \put(24.5,61.5){\line(1,0){24}}
\put(24.5,11.5){\line(0,1){50}} \put(48.5,11.5){\line(0,1){50}}

\put(25,57){\makebox(0,0)[bl]{Restructuring:}}
\put(25,53){\makebox(0,0)[bl]{\(S^{0} \Rightarrow \{S^{1*},\)}}

\put(25,49){\makebox(0,0)[bl]{\( S^{2*},...,S^{n*}\}\)}}

\put(25,44.5){\makebox(0,0)[bl]{while taking}}
\put(25,41.5){\makebox(0,0)[bl]{into account:}}
\put(25,37){\makebox(0,0)[bl]{(i) \(S^{1*}\) is close}}
\put(25,33.4){\makebox(0,0)[bl]{to \(S^{1}\), \(S^{2*}\) is}}

\put(25,29.4){\makebox(0,0)[bl]{close to \(S^{2}\), ...}}

\put(25,25){\makebox(0,0)[bl]{(ii) changes of }}
\put(25,21){\makebox(0,0)[bl]{\(S^{0}\) into \(S^{1*}\), }}

\put(25,17){\makebox(0,0)[bl]{\(S^{1*}\) into \(S^{2*}\), }}
\put(26,13){\makebox(0,0)[bl]{... are cheap.}}


\put(51,44){\line(1,0){22}} \put(51,58){\line(1,0){22}}
\put(51,44){\line(0,1){14}} \put(73,44){\line(0,1){14}}

\put(51.5,53){\makebox(0,0)[bl]{Requirements}}
\put(51.5,49){\makebox(0,0)[bl]{(for }}

\put(52.5,45){\makebox(0,0)[bl]{\(\tau_{1},\tau_{2},...,\tau_{n}\))}}


\put(62,44){\vector(0,-1){4}}

\put(51,26){\line(1,0){22}} \put(51,40){\line(1,0){22}}
\put(51,26){\line(0,1){14}} \put(73,26){\line(0,1){14}}

\put(51.5,35){\makebox(0,0)[bl]{Optimization}}
\put(51.5,31){\makebox(0,0)[bl]{problem (for}}
\put(51.5,27){\makebox(0,0)[bl]{\(\tau_{1},\tau_{2},...,\tau_{n}\))}}


\put(62,26){\vector(0,-1){4}}

\put(62,18){\oval(22,08)}

\put(55,19){\makebox(0,0)[bl]{Solutions }}
\put(52,15){\makebox(0,0)[bl]{\(S^{1},S^{2},...,S^{n}\)}}

\end{picture}
%
\begin{picture}(80,47)

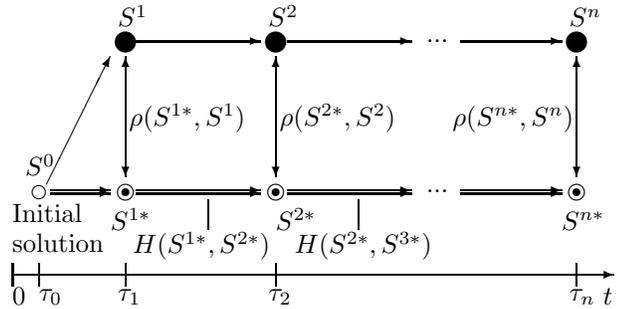
\put(07,00){\makebox(0,0)[bl]{Fig. 9. Multi-stage restructuring
 problem}}


\put(00,09){\vector(1,0){80}}

\put(78.5,05){\makebox(0,0)[bl]{\(t\)}}

\put(00,7.5){\line(0,1){3}} \put(3.5,7.5){\line(0,1){3}}
\put(15,7.5){\line(0,1){3}} \put(35,7.5){\line(0,1){3}}
\put(75,7.5){\line(0,1){3}}

\put(00,05){\makebox(0,0)[bl]{\(0\)}}
\put(3.5,05){\makebox(0,0)[bl]{\(\tau_{0}\)}}
\put(14,05){\makebox(0,0)[bl]{\(\tau_{1}\)}}
\put(34,05){\makebox(0,0)[bl]{\(\tau_{2}\)}}
\put(74,05){\makebox(0,0)[bl]{\(\tau_{n}\)}}


\put(1.7,22.2){\makebox(0,0)[bl]{\(S^{0}\)}}
\put(3.5,20){\circle{1.6}}

\put(00,16){\makebox(0,0)[bl]{Initial}}
\put(00,12){\makebox(0,0)[bl]{solution}}

\put(5,20.2){\vector(1,0){8}} \put(5,19.8){\vector(1,0){8}}

\put(4.5,21){\vector(1,2){8.5}}


\put(14,42){\makebox(0,0)[bl]{\(S^{1}\)}}

\put(15,40){\circle*{2.7}}

\put(16.5,40){\vector(1,0){16.5}}


\put(15,20){\circle{2}} \put(15,20){\circle*{1}}

\put(16.5,20.2){\vector(1,0){16.5}}
\put(16.5,19.8){\vector(1,0){16.5}}

\put(13,15){\makebox(0,0)[bl]{\(S^{1*}\)}}


\put(15,22){\vector(0,1){16}} \put(15,38){\vector(0,-1){16}}

\put(15.5,28){\makebox(0,0)[bl]{\(\rho(S^{1*},S^{1})\)}}


\put(26,15){\line(0,1){4}}

\put(16,11){\makebox(0,0)[bl]{\(H(S^{1*},S^{2*})\)}}


\put(34,42){\makebox(0,0)[bl]{\(S^{2}\)}}

\put(35,40){\circle*{2.7}}

\put(36.5,40){\vector(1,0){16.5}}


\put(35,20){\circle{2}} \put(35,20){\circle*{1}}

\put(36.5,20.2){\vector(1,0){16.5}}
\put(36.5,19.8){\vector(1,0){16.5}}

\put(35,15){\makebox(0,0)[bl]{\(S^{2*}\)}}


\put(35,22){\vector(0,1){16}} \put(35,38){\vector(0,-1){16}}

\put(35.5,28){\makebox(0,0)[bl]{\(\rho(S^{2*},S^{2})\)}}


\put(46,15){\line(0,1){4}}

\put(37.5,11){\makebox(0,0)[bl]{\(H(S^{2*},S^{3*})\)}}

\put(74,42){\makebox(0,0)[bl]{\(S^{n}\)}}

\put(75,40){\circle*{2.7}}

\put(59.5,40){\vector(1,0){13.5}}

\put(55,40){\makebox(0,0)[bl]{...}}


\put(75,20){\circle{2}} \put(75,20){\circle*{1}}

\put(59.5,20.2){\vector(1,0){13.5}}
\put(59.5,19.8){\vector(1,0){13.5}}

\put(55,20){\makebox(0,0)[bl]{...}}

\put(73,15){\makebox(0,0)[bl]{\(S^{n*}\)}}


\put(75,22){\vector(0,1){16}} \put(75,38){\vector(0,-1){16}}

\put(58.5,28){\makebox(0,0)[bl]{\(\rho(S^{n*},S^{n})\)}}


\end{picture}
\end{center}

 Clearly, the multi-stage restructuring problems
 are very complicated.
 The problems consist of a combination of NP-hard combinatorial
 problems.
 Thus,  it is necessary to use composite
 heuristic solving schemes for the multi-stage restructuring problems.

 Table 2 contains an integrated list on basic research directions
 on the considered six types of modification problems in
 combinatorial optimization.

\newpage
\begin{center}
{\bf Table 2.} Basic research reoptimization/restructuring directions in combinatorial optimization   \\
\begin{tabular}{| c | l | l |}
\hline
 No.  & Direction  & Source \\

\hline
 1.& Modification of solution/structure (reassignment, relinking, &\\
 &rescheduling, repositioning, editing, recoloring, etc.):&\\

 1.1.& Reassignment/relocation/repositioning  &\cite{berm81,brot03,cox73,ho14,lee01,mas13,mel00,vos96,wang05,wangz16}\\

 1.2.& Rescheduling & \cite{chur92,jain97,li93,ran04,vier03,zweb93}\\

 1.3.& Path relinking:
 routing, TSP, orienteering, network design
   &\cite{aiex05,alva12,far05,gham04,glov00,ho06,lag99,mes12,moraes12,ped12}\\
 &(multi-layer optimization, load balancing, topology control) &\cite{reeves98,res05,sor12,souf10,usb12,val10,yag06}\\

 1.4.&Reconnecting network partitions& \cite{dini08} \\

 1.5.&Hotlink assignment problems (addition of direct links into
  &\cite{bose01,czy03,fuhr01,lev12hier,lev15}\\
  &hierarchical/tree-like information structure)&\\

 1.6.& Recoloring of graphs (e.g., paths, strings, trees)
 &\cite{lev09,lima11,moran08}\\

 1.7.& Vehicle relocation problem & \cite{sch15}\\

 1.8.& Block relocation problem (container relocation problem)&
 \cite{cas10,cas11,cas12,jov14,kim06,pet13,zeh15,zhu12}\\


 2.& Reoptimization (small correction of solution to improve its    & \\
  &quality as
   the objective function(s)):&\\

 2.1.& Minimum spanning tree problem&\cite{boria10}\\
 2.2.& Traveling salesman problems
 (TSP), postman problem, etc.&\cite{archetti03,arch13,aus09}\\
 2.3. &Steiner tree problems&\cite{bilo08,esc09}\\
 2.4.& Covering problems& \cite{bilo08a}\\
 2.5.& Shortest common superstring problem& \cite{bilo11}\\


 3.&Augmentation-type problems (addition/correction of    & \\
 &solution components to obtain required solution properties):   &\\

 3.1.&Augmentation network problems (addition of links to obtain & \cite{esw76}\\
 & required network properties (e.g., connectivity level)  &\\

 3.2.& Social network restructuring
 (node/link addition/deletion)
 & \cite{guna13}\\

 3.3.& Cluster editing problem (edge addition/deletion in graph
 & \cite{bock09,bock11,dam10,dehne06,guo09,mann10}\\
 & to obtain a disjoint union of cliques)&\cite{rah07,sham04}\\


 4.&Reload cost problems, changeover cost problems:&\\

 4.1&Reload cost spanning trees, networks &\cite{gam12,goz14,wirth01}\\

 4.2&Reload cost paths, tours, flows &\cite{amal11}\\

 4.3&Spectrum switching scheduling  &\cite{goz13}\\


 5.& Restructuring problems (modification of solution with  & \\
 & two criteria: minimum modification cost,  minimum  &\\
 &proximity to a next solution at the next time stage):&\\

 5.1& Knapsack problem & \cite{lev11restr,lev15}\\
 5.2& Multiple choice problem  & \cite{lev11restr}\\
 5.3& Spanning tree problems  & \cite{lev11restr}\\
 5.4& Clustering problem & \cite{lev15c}\\
 5.5& Assignment/location problems & this paper \\


 6.&Dynamic combinatorial optimization problems:&\\

 6.1.& Dynamic knapsack problem &\cite{kley98,kley01,roh09}\\

 6.2.& Dynamic clustering& \cite{chaud94,chen04,de06,jay14,mcdon99,omran06,papa08,yu07}\\

 6.3.& Dynamic scheduling (e.g.,
 rescheduling strategies) & \cite{adib10,ames01,ayd00,aytug05,cow02,exp14,man98,ou09,vier03}\\

 6.4. &Online bin-packing &  \cite{asch99,gal95}\\

 6.5. & Dynamic routing (e.g.,
 VRP with changing requests)
 &\cite{ash97,chenh98,kho12,savel98}\\

 6.6.& Dynamic path replanning for UAVs & \cite{wzorek06}\\


 7.& Multistage dynamic restructuring problems  &\\

 7.1.& Knapsack problem  &this paper \\

 7.2.& Classification, clustering, sorting &\cite{lev15c}, this paper\\

 7.3.&Morphological clique problem  & this paper\\

\hline
\end{tabular}
\end{center}

\section{Basic Assessment Scales}

 The list of basic considered assessment scales
(for system parts/ components, for final system)
 involves the following
(e.g., \cite{lev12a,lev15}):
 (i) quantitative scale,
 (ii) ordinal scale,
 (iii) multicriteria description or vector estimate,
 (iv) poset-like scales
 (based on ordinal vectors, based on multiset
 estimates).
 The descriptions for the scales is presented in
  \cite{lev12a,lev15}.
 Some illustrations for the scales above are shown in
 Fig. 10,  Fig. 11, Fig. 12.
 Let us consider  illustrations for the above-mentioned
 basic assessment scales.

 In the case of vector scales,
 domination is illustrated in Fig. 10c:
 \(\alpha_{2} \succ  \beta_{2}\),
 \(\alpha_{2} \succ  \beta_{3}\),
 \(\alpha_{2} \succ  \beta_{4}\).
 In the case of domination by Pareto-rule
 (e.g., \cite{mirkin79,pareto71}),
 the basic domination binary relation is extended by cases as
 \(\alpha_{2} \succ^{P}  \beta_{1}\).
 Here,
 the following ordered layers of quality
 can be considered
 (as a special system ordinal scale \(D\),
 by illustration in Fig. 10c):
 (i) the ideal point (the best point)
 \(\alpha^{I}\),
 (ii) a layer of Pareto-efficient points
 (e.g., points: ~\(\{\alpha_{1},\alpha_{2},\alpha_{3},\alpha_{4}\}\)),
 (iii) near Pareto-efficient points
 (the points are close to the Pareto-layer,
 e.g., points: ~\(\{\beta_{1},\beta_{2},\beta_{3},\beta_{4},\beta_{5}\}\)),
 (iv) a next layer of quality
 (i.e, between near Pareto-efficient points and the worst point,
 e.g., points: ~\(\{\gamma_{1},\gamma_{2}\}\)),
 and
 ~(v) the worst point.

 The description of poset-like scales
  (or lattices) for quality of
 composite (modular) systems
 (based on ordinal estimates of DAs and their compatibility)
  was suggested within framework of HMMD approach
 (e.g., \cite{lev98,lev06,lev15})
 Here, two cases have to be examined:
 (1) scale for system quality
 based on system components ordinal estimates
 (\(\iota = \overline{1,l}\);
      \(1\) corresponds to the best one);
 (2) scale for system quality
  while taking into account
 system components ordinal estimates
 and ordinal compatibility
 estimates between the system components
  (\(w=\overline{1,\nu}\); \(\nu\) corresponds to the best level).

\begin{center}
\begin{picture}(54,49)

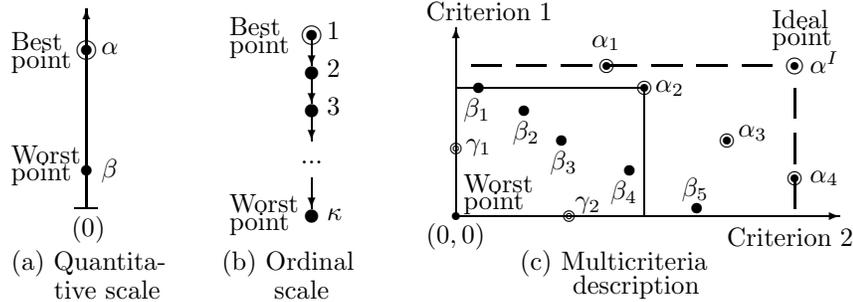
\put(00,00){\makebox(0,0)[bl]{Fig. 10. Quantitative scale,
 ordinal scale, multicriteria description
  \cite{lev12a,lev15} }}

\put(00,9){\makebox(0,0)[bl]{(a) Quantita-}}
\put(05.7,6){\makebox(0,0)[bl]{tive scale}}

\put(08.5,18){\line(1,0){3}}

\put(08,13.5){\makebox(0,0)[bl]{(\(0\))}}

\put(10,18){\vector(0,1){26.5}}


\put(12,38.5){\makebox(0,0)[bl]{\(\alpha\)}}

\put(10,39){\circle*{1.4}} \put(10,39){\circle{2.4}}

\put(00,39){\makebox(0,0)[bl]{Best}}
\put(00,36){\makebox(0,0)[bl]{point}}

\put(12,21.5){\makebox(0,0)[bl]{\(\beta\)}}

\put(10,23){\circle*{1.5}}

\put(00,24){\makebox(0,0)[bl]{Worst}}
\put(00,21){\makebox(0,0)[bl]{point}}

\put(28,9){\makebox(0,0)[bl]{(b) Ordinal}}
\put(35,6){\makebox(0,0)[bl]{scale}}

\put(29,41){\makebox(0,0)[bl]{Best}}
\put(29,38){\makebox(0,0)[bl]{point}}

\put(40,41){\circle*{1.4}} \put(40,41){\circle{2.4}}
\put(42,40.5){\makebox(0,0)[bl]{\(1\)}}

\put(40,41){\vector(0,-1){4}}

\put(42,35.5){\makebox(0,0)[bl]{\(2\)}}

\put(40,36){\circle*{1.9}} \put(40,36){\vector(0,-1){4}}

\put(42,30.5){\makebox(0,0)[bl]{\(3\)}}

\put(40,31){\circle*{1.9}} \put(40,31){\vector(0,-1){4}}

\put(38.5,24){\makebox(0,0)[bl]{...}}


\put(40,22){\vector(0,-1){4}}


\put(42,16.5){\makebox(0,0)[bl]{\(\kappa\)}}

\put(40,17){\circle*{1.9}}

\put(29,17.5){\makebox(0,0)[bl]{Worst}}
\put(29,14.5){\makebox(0,0)[bl]{point}}

\end{picture}
%
\begin{picture}(58,49)

\put(12.5,9){\makebox(0,0)[bl]{(c) Multicriteria}}
\put(19.5,6){\makebox(0,0)[bl]{description}}

\put(04,17){\circle*{0.9}}

\put(05,20.2){\makebox(0,0)[bl]{Worst}}
\put(05,17.2){\makebox(0,0)[bl]{point}}

\put(00,12.5){\makebox(0,0)[bl]{\((0,0)\)}}

\put(04,17){\vector(0,1){25}} \put(04,17){\vector(1,0){51}}

\put(40,13){\makebox(0,0)[bl]{Criterion 2}}
\put(00,43){\makebox(0,0)[bl]{Criterion 1}}


\put(06,37){\line(1,0){4}} \put(12,37){\line(1,0){4}}
\put(18,37){\line(1,0){4}} \put(24,37){\line(1,0){4}}
\put(30,37){\line(1,0){4}} \put(36,37){\line(1,0){4}}
\put(42,37){\line(1,0){4}}



\put(49,18){\line(0,1){4}} \put(49,24){\line(0,1){4}}
\put(49,30){\line(0,1){4}}


\put(49,37){\circle*{1}} \put(49,37){\circle{2}}

\put(46,42){\makebox(0,0)[bl]{Ideal}}
\put(46,39){\makebox(0,0)[bl]{point}}
\put(51,36){\makebox(0,0)[bl]{\(\alpha^{I}\)}}


\put(24,37){\circle*{1.1}} \put(24,37){\circle{1.9}}
\put(22,39){\makebox(0,0)[bl]{\(\alpha_{1}\)}}

\put(29,34){\circle*{1.1}} \put(29,34){\circle{1.9}}
\put(30.5,33){\makebox(0,0)[bl]{\(\alpha_{2}\)}}
\put(29,34){\line(-1,0){25}} \put(29,34){\line(0,-1){17}}

\put(40,27){\circle*{1.1}} \put(40,27){\circle{1.9}}
\put(41.5,27){\makebox(0,0)[bl]{\(\alpha_{3}\)}}

\put(49,22){\circle*{1.1}} \put(49,22){\circle{1.9}}
\put(51,21){\makebox(0,0)[bl]{\(\alpha_{4}\)}}


\put(07,34){\circle*{1.4}}
\put(05,29.5){\makebox(0,0)[bl]{\(\beta_{1}\)}}

\put(13,31){\circle*{1.4}}
\put(11.5,26.5){\makebox(0,0)[bl]{\(\beta_{2}\)}}

\put(18,27){\circle*{1.4}}
\put(16.5,22.5){\makebox(0,0)[bl]{\(\beta_{3}\)}}

\put(27,23){\circle*{1.4}}
\put(24.5,18.5){\makebox(0,0)[bl]{\(\beta_{4}\)}}

\put(36,18){\circle*{1.4}}
\put(34,19){\makebox(0,0)[bl]{\(\beta_{5}\)}}


\put(04,26){\circle{0.75}} \put(04,26){\circle{1.4}}
\put(5.3,25){\makebox(0,0)[bl]{\(\gamma_{1}\)}}

\put(19,17){\circle{0.75}} \put(19,17){\circle{1.4}}
\put(19.8,17.3){\makebox(0,0)[bl]{\(\gamma_{2}\)}}

\end{picture}
\end{center}

 For the system consisting of \(m\) parts/components,
  a discrete space (poset, lattice) of the system quality (excellence)
   on the basis of the following vector is used:
 ~\(N(S)=(w(S);n(S))\),
 ~where \(w(S)\) is the minimum of pairwise compatibility
 between DAs which correspond to different system components
 (i.e.,
 \(~\forall ~P_{j_{1}}\) and \( P_{j_{2}}\),
 \(1 \leq j_{1} \neq j_{2} \leq m\))
 in \(S\),
 ~\(n(S)=(\eta_{1},...,\eta_{r},...,\eta_{k})\),
 ~where ~\(\eta_{r}\) is the number of DAs of the \(r\)th quality in ~\(S\)
 ~(\(\sum^{k}_{r=1} n_{r} = m \)).

 An example of the  three-component system
 ~\(S = X \star Y \star Z\)
 is considered.
 The following ordinal scales are used:
 (a) ordinal scale for elements (priorities) is \([1,2,3]\),
 (b) ordinal scale for compatibility is \([0,1,2,3]\).
 For this case,
 Fig. 11a depicts the poset of system quality
  by components and
 Fig. 11b depicts an integrated poset with compatibility
 (each triangle corresponds to the poset from Fig. 11a).
   Generally, the following layers of system excellence can be considered
    (Fig. 11, this corresponds to the resultant system scale \(D\) in Fig. 11b):

  {\it 1.} The ideal point \(N(S^{I})\) (\(S^{I}\) is the ideal system solution).

  {\it 2.} A layer of Pareto-efficient solutions:
  \( \{ S_{1}^{p}, S_{2}^{p}, S_{3}^{p} \} \);
  estimates are:
 \(N(S^{p}_{1}) = (2;3,0,0)\),
 \(N(S^{p}_{2})=(3;1,1,1)\), and
 \(N(S^{p}_{3})=(3;0,3,0)\).

  {\it 3.} A next layer of quality
 (e.g., neighborhood of Pareto-efficient solutions layer):
 \( \{ S'_{1}, S'_{2}, S'_{3} \} \);
 estimates are:
  \(N(S'_{1}) = (1;3,0,0)\),
 \(N(S'_{2})=(2;1,1,1)\), and
 \(N(S'_{3})=(3;0,2,1)\);
  a composite solution of this set can be
  transformed into a Pareto-efficient solution
  on the basis of a
 simple improvement action(s)
 (e.g., as modification of the only one  element).

  {\it 4.} A next layer of quality \(S''\);
  estimate is:
  \(N(S'')=(1;0,3,0)\).

 {\it 5.} The worst point
 \(S_{0}\);
 estimate is:
 \(N(S_{0}) = (1;0,0,3)\).

 Note,
 the compatibility component of vector ~\(N(S)\)
 can be considered on the basis of a poset-like scale too
 (as \(n(S)\))
  \cite{lev06}.
 In this case, the discrete space of system excellence
 will be an analogical lattice.

\begin{center}
\begin{picture}(58,87)

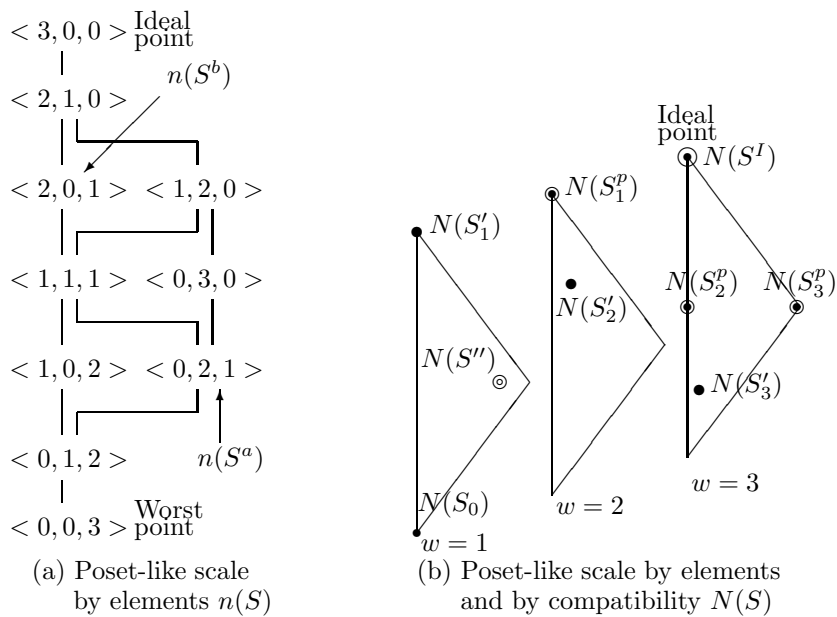
\put(06,00){\makebox(0,0)[bl] {Fig. 11. Poset-like scale
 based on ordinal estimates
 (based on \cite{lev15}) }}

\put(08,09){\makebox(0,0)[bl] {(a) Poset-like scale}}
\put(13.5,05){\makebox(0,0)[bl]{by elements \(n(S)\)}}

\put(05,81){\makebox(0,0)[bl]{\(<3,0,0>\) }}

\put(21.6,83){\makebox(0,0)[bl]{Ideal}}
\put(21.6,80){\makebox(0,0)[bl]{point}}

\put(12,77){\line(0,1){3}}
\put(05,72){\makebox(0,0)[bl]{\(<2,1,0>\)}}

\put(26,75){\makebox(0,0)[bl]{\(n(S^{b})\)}}
\put(25,74){\vector(-1,-1){10}}

\put(12,65){\line(0,1){6}}
\put(05,60){\makebox(0,0)[bl]{\(<2,0,1>\) }}

\put(12,53){\line(0,1){6}}
\put(05,48){\makebox(0,0)[bl]{\(<1,1,1>\) }}

\put(12,41){\line(0,1){6}}
\put(05,36){\makebox(0,0)[bl]{\(<1,0,2>\) }}


\put(12,29){\line(0,1){6}}
\put(05,24){\makebox(0,0)[bl]{\(<0,1,2>\) }}

\put(30,24){\makebox(0,0)[bl]{\(n(S^{a})\)}}
\put(33,28){\vector(0,1){7}}

\put(12,20){\line(0,1){3}}

\put(05,15){\makebox(0,0)[bl]{\(<0,0,3>\) }}

\put(21.6,18){\makebox(0,0)[bl]{Worst}}
\put(21.6,15){\makebox(0,0)[bl]{point}}

\put(14,68){\line(0,1){3}} \put(30,68){\line(-1,0){16}}
\put(30,65){\line(0,1){3}}

\put(23,60){\makebox(0,0)[bl]{\(<1,2,0>\) }}

\put(30,59){\line(0,-1){3}} \put(30,56){\line(-1,0){16}}
\put(14,56){\line(0,-1){3}}
\put(32,53){\line(0,1){6}}
\put(23,48){\makebox(0,0)[bl]{\(<0,3,0>\) }}

\put(14,44){\line(0,1){3}} \put(30,44){\line(-1,0){16}}
\put(30,41){\line(0,1){3}}

\put(32,41){\line(0,1){6}}
\put(23,36){\makebox(0,0)[bl]{\(<0,2,1>\) }}

\put(30,35){\line(0,-1){3}} \put(30,32){\line(-1,0){16}}
\put(14,32){\line(0,-1){3}}

\end{picture}
%
\begin{picture}(56,59)

\put(00,09){\makebox(0,0)[bl]{(b) Poset-like scale by elements}}
\put(05.8,05){\makebox(0,0)[bl]{and by compatibility \(N(S)\)}}

\put(00,16){\circle*{0.9}}
\put(0.1,18){\makebox(0,0)[bl]{\(N(S_{0})\)}}

\put(00,16){\line(0,1){40}} \put(00,16){\line(3,4){15}}
\put(00,56){\line(3,-4){15}}

\put(18,21){\line(0,1){40}} \put(18,21){\line(3,4){15}}
\put(18,61){\line(3,-4){15}}

\put(36,26){\line(0,1){40}} \put(36,26){\line(3,4){15}}
\put(36,66){\line(3,-4){15}}



\put(18,61){\circle*{1.1}} \put(18,61){\circle{1.9}}
\put(19.6,60){\makebox(0,0)[bl]{\(N(S^{p}_{1})\)}}

\put(36,46){\circle*{1.1}} \put(36,46){\circle{1.9}}
\put(32.9,47.4){\makebox(0,0)[bl]{\(N(S^{p}_{2})\)}}

\put(50.5,46){\circle*{1.1}} \put(50.5,46){\circle{1.9}}
\put(46,47.4){\makebox(0,0)[bl]{\(N(S^{p}_{3})\)}}


\put(00,56){\circle*{1.5}}
\put(01.6,55){\makebox(0,0)[bl]{\(N(S'_{1})\)}}

\put(20.5,49){\circle*{1.5}}
\put(18.4,44){\makebox(0,0)[bl]{\(N(S'_{2})\)}}

\put(37.5,35){\circle*{1.5}}
\put(39,33.8){\makebox(0,0)[bl]{\(N(S'_{3})\)}}


\put(11,36){\circle{0.75}} \put(11,36){\circle{1.7}}
\put(00.5,37.1){\makebox(0,0)[bl]{\(N(S'')\)}}


\put(36,66){\circle*{1}} \put(36,66){\circle{2.5}}

\put(32,70.5){\makebox(0,0)[bl]{Ideal}}
\put(32,67.5){\makebox(0,0)[bl]{point}}

\put(38,64){\makebox(0,0)[bl]{\(N(S^{I})\)}}

\put(00.5,13.5){\makebox(0,0)[bl]{\(w=1\)}}
\put(18.5,18.5){\makebox(0,0)[bl]{\(w=2\)}}
\put(36.5,21.5){\makebox(0,0)[bl]{\(w=3\)}}

\end{picture}
\end{center}


 The  poset-like scales based on interval multiset estimates
 have been suggested in \cite{lev12a,lev15}.
 Analogically, two cases have to be considered:
 (i) system estimate by components,
 (ii) system estimate by components and by component compatibility.
%
%
 Fig. 12a illustrates the scale-poset and estimates for problem
 \(P_{3,3}\)
 (assessment over scale \([1,3]\)
 with three elements, estimates \((2,0,2\) and \((1,0,2)\) are not used)
 \cite{lev12a,lev15}.
 Evidently, the above-mentioned resultant special system ordinal scale \(D\)
  can used here as well.
 For evaluation of multi-component system,
 multi-component poset-like scale
 (as in Fig. 11b)
 composed from several poset-like scale
 (as in Fig. 12)
  may be used
 \cite{lev12a,lev15}.
 Fig. 12b depicts
 the integrated poset-like scale for tree-component system
 (compatibility scale is \([0,1,2,3]\)).

\begin{center}
\begin{picture}(84,102)

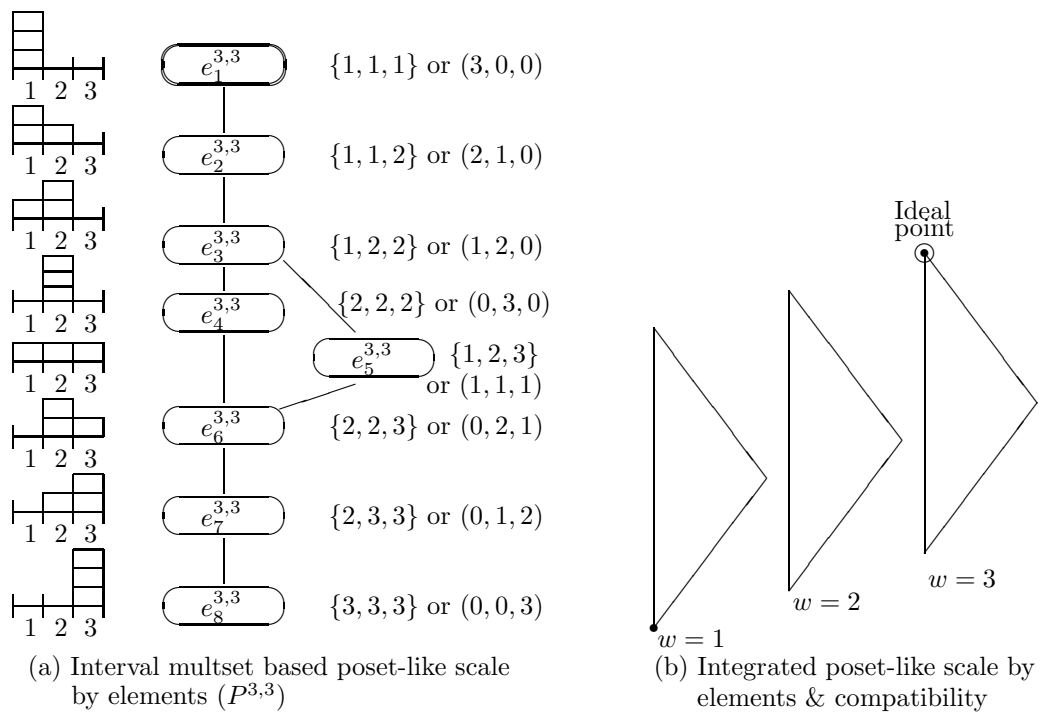
\put(22,00){\makebox(0,0)[bl]{Fig. 12. Multiset based
 scale, estimates (based on \cite{lev12a,lev15}) }}

\put(02,09){\makebox(0,0)[bl]{(a) Interval multset based
poset-like scale}}

\put(07.8,05){\makebox(0,0)[bl]{by elements (\(P^{3,3}\))}}

\put(25,88.7){\makebox(0,0)[bl]{\(e^{3,3}_{1}\) }}

\put(28,91){\oval(16,5)} \put(28,91){\oval(16.5,5.5)}


\put(42,89){\makebox(0,0)[bl]{\(\{1,1,1\}\) or \((3,0,0)\) }}

\put(00,90.5){\line(0,1){07.5}} \put(04,90.5){\line(0,1){07.5}}

\put(00,93){\line(1,0){04}} \put(00,95.5){\line(1,0){04}}
\put(00,98){\line(1,0){4}}

\put(00,90.5){\line(1,0){12}}

\put(00,89){\line(0,1){3}} \put(04,89){\line(0,1){3}}
\put(08,89){\line(0,1){3}} \put(12,89){\line(0,1){3}}

\put(01.5,86.5){\makebox(0,0)[bl]{\(1\)}}
\put(05.5,86.5){\makebox(0,0)[bl]{\(2\)}}
\put(09.5,86.5){\makebox(0,0)[bl]{\(3\)}}


\put(28,82){\line(0,1){6}}

\put(25,76.7){\makebox(0,0)[bl]{\(e^{3,3}_{2}\) }}

\put(28,79){\oval(16,5)}


\put(42,77){\makebox(0,0)[bl]{\(\{1,1,2\}\) or \((2,1,0)\) }}

\put(00,80.5){\line(0,1){05}} \put(04,80.5){\line(0,1){05}}
\put(8,80.5){\line(0,1){02.5}}

\put(00,83){\line(1,0){8}} \put(00,85.5){\line(1,0){4}}

\put(00,80.5){\line(1,0){12}}

\put(00,79){\line(0,1){3}} \put(04,79){\line(0,1){3}}
\put(08,79){\line(0,1){3}} \put(12,79){\line(0,1){3}}

\put(01.5,76.5){\makebox(0,0)[bl]{\(1\)}}
\put(05.5,76.5){\makebox(0,0)[bl]{\(2\)}}
\put(09.5,76.5){\makebox(0,0)[bl]{\(3\)}}


\put(28,70){\line(0,1){6}}

\put(25,64.7){\makebox(0,0)[bl]{\(e^{3,3}_{3}\) }}

\put(28,67){\oval(16,5)}


\put(42,65){\makebox(0,0)[bl]{\(\{1,2,2\}\) or \((1,2,0)\) }}

\put(00,70.5){\line(0,1){02.5}} \put(04,70.5){\line(0,1){05}}
\put(8,70.5){\line(0,1){05}}

\put(00,73){\line(1,0){8}} \put(04,75.5){\line(1,0){4}}

\put(00,70.5){\line(1,0){12}}

\put(00,69){\line(0,1){3}} \put(04,69){\line(0,1){3}}
\put(08,69){\line(0,1){3}} \put(12,69){\line(0,1){3}}

\put(01.5,66.5){\makebox(0,0)[bl]{\(1\)}}
\put(05.5,66.5){\makebox(0,0)[bl]{\(2\)}}
\put(09.5,66.5){\makebox(0,0)[bl]{\(3\)}}


\put(28,61){\line(0,1){3}}

\put(25,55.7){\makebox(0,0)[bl]{\(e^{3,3}_{4}\) }}

\put(28,58){\oval(16,5)}


\put(43,57.4){\makebox(0,0)[bl]{\(\{2,2,2\}\) or \((0,3,0)\) }}

\put(04,59.5){\line(0,1){06}} \put(08,59.5){\line(0,1){06}}

\put(04,61.5){\line(1,0){04}} \put(04,63.5){\line(1,0){04}}
\put(04,65.5){\line(1,0){04}}

\put(00,59.5){\line(1,0){12}}

\put(00,58){\line(0,1){3}} \put(04,58){\line(0,1){3}}
\put(08,58){\line(0,1){3}} \put(12,58){\line(0,1){3}}

\put(01.5,55.5){\makebox(0,0)[bl]{\(1\)}}
\put(05.5,55.5){\makebox(0,0)[bl]{\(2\)}}
\put(09.5,55.5){\makebox(0,0)[bl]{\(3\)}}


\put(28,46){\line(0,1){9}}

\put(25,40.7){\makebox(0,0)[bl]{\(e^{3,3}_{6}\) }}

\put(28,43){\oval(16,5)}


\put(42,41){\makebox(0,0)[bl]{\(\{2,2,3\}\) or \((0,2,1)\) }}

\put(04,41.5){\line(0,1){05}} \put(08,41.5){\line(0,1){05}}
\put(12,41.5){\line(0,1){02.5}}

\put(04,44){\line(1,0){8}} \put(04,46.5){\line(1,0){4}}

\put(00,41.5){\line(1,0){12}}

\put(00,40){\line(0,1){3}} \put(04,40){\line(0,1){3}}
\put(08,40){\line(0,1){3}} \put(12,30){\line(0,1){3}}

\put(01.5,37.5){\makebox(0,0)[bl]{\(1\)}}
\put(05.5,37.5){\makebox(0,0)[bl]{\(2\)}}
\put(09.5,37.5){\makebox(0,0)[bl]{\(3\)}}


\put(28,34){\line(0,1){6}}

\put(25,28.7){\makebox(0,0)[bl]{\(e^{3,3}_{7}\) }}

\put(28,31){\oval(16,5)}


\put(42,29){\makebox(0,0)[bl]{\(\{2,3,3\}\) or \((0,1,2)\) }}

\put(04,31.5){\line(0,1){02.5}} \put(08,31.5){\line(0,1){05}}
\put(12,31.5){\line(0,1){05}}

\put(04,34){\line(1,0){8}} \put(08,36.5){\line(1,0){4}}

\put(00,31.5){\line(1,0){12}}

\put(00,30){\line(0,1){3}} \put(04,30){\line(0,1){3}}
\put(08,30){\line(0,1){3}} \put(12,30){\line(0,1){3}}

\put(01.5,27.5){\makebox(0,0)[bl]{\(1\)}}
\put(05.5,27.5){\makebox(0,0)[bl]{\(2\)}}
\put(09.5,27.5){\makebox(0,0)[bl]{\(3\)}}


\put(28,22){\line(0,1){6}}

\put(25,16.7){\makebox(0,0)[bl]{\(e^{3,3}_{8}\) }}

\put(28,19){\oval(16,5)}


\put(42,17){\makebox(0,0)[bl]{\(\{3,3,3\}\) or \((0,0,3)\) }}

\put(08,19){\line(0,1){07.5}} \put(12,19){\line(0,1){07.5}}
\put(08,21.5){\line(1,0){4}} \put(08,24){\line(1,0){4}}
\put(08,26.5){\line(1,0){4}}

\put(00,19){\line(1,0){12}}

\put(00,017.5){\line(0,1){3}} \put(04,017.5){\line(0,1){3}}
\put(08,017.5){\line(0,1){3}} \put(12,017.5){\line(0,1){3}}

\put(01.5,15){\makebox(0,0)[bl]{\(1\)}}
\put(05.5,15){\makebox(0,0)[bl]{\(2\)}}
\put(09.5,15){\makebox(0,0)[bl]{\(3\)}}


\put(45.5,55.5){\line(-1,1){09.5}}

\put(45.5,48.5){\line(-3,-1){10}}

\put(45,49.7){\makebox(0,0)[bl]{\(e^{3,3}_{5}\) }}

\put(48,52){\oval(16,5)}


\put(58,50.5){\makebox(0,0)[bl]{\(\{1,2,3\}\) }}
\put(55,46.5){\makebox(0,0)[bl]{or \((1,1,1)\) }}

\put(00,51.5){\line(0,1){02.5}} \put(04,51.5){\line(0,1){02.5}}
\put(8,51.5){\line(0,1){02.5}} \put(12,51.5){\line(0,1){02.5}}

\put(00,54){\line(1,0){12}}

\put(00,51.5){\line(1,0){12}}

\put(00,50){\line(0,1){3}} \put(04,50){\line(0,1){3}}
\put(08,50){\line(0,1){3}} \put(12,50){\line(0,1){3}}

\put(01.5,47.5){\makebox(0,0)[bl]{\(1\)}}
\put(05.5,47.5){\makebox(0,0)[bl]{\(2\)}}
\put(09.5,47.5){\makebox(0,0)[bl]{\(3\)}}


\end{picture}
%
\begin{picture}(56,59)

\put(00,09){\makebox(0,0)[bl]{(b) Integrated poset-like scale by}}

\put(05.8,05){\makebox(0,0)[bl]{elements \& compatibility}}

\put(00,16){\circle*{0.9}}

\put(00,16){\line(0,1){40}} \put(00,16){\line(3,4){15}}
\put(00,56){\line(3,-4){15}}

\put(18,21){\line(0,1){40}} \put(18,21){\line(3,4){15}}
\put(18,61){\line(3,-4){15}}

\put(36,26){\line(0,1){40}} \put(36,26){\line(3,4){15}}
\put(36,66){\line(3,-4){15}}



\put(36,66){\circle*{1}} \put(36,66){\circle{2.5}}

\put(32,70.5){\makebox(0,0)[bl]{Ideal}}
\put(32,67.5){\makebox(0,0)[bl]{point}}


\put(00.5,13.5){\makebox(0,0)[bl]{\(w=1\)}}
\put(18.5,18.5){\makebox(0,0)[bl]{\(w=2\)}}
\put(36.5,21.5){\makebox(0,0)[bl]{\(w=3\)}}

\end{picture}
\end{center}

\section{Restructuring Problems}

\subsection{One-stage restructuring}

 The basic one-stage restructuring problem was illustrated in
 Fig. 4 and Fig. 5.
 Let \(P\) be a combinatorial optimization problem with a solution as
 structure
 \(S\)
 (i.e., subset, graph),
 \(\Omega\) be initial data (elements, element parameters, etc.),
 \(f(P)\) be objective function(s).
 Thus, \(S(\Omega)\) be a solution for initial data \(\Omega\),
 \(f(S(\Omega))\) be the corresponding objective function.
 Let \(\Omega^{1}\) be initial data at an initial stage,
  \(f(S(\Omega^{1}))\) be the corresponding objective function.
 \(\Omega^{2}\) be initial data at next stage,
  \(f(S(\Omega^{2}))\) be the corresponding objective function.
 As a result,
 the following solutions can be considered:
 ~(a) \( S^{1}=S(\Omega^{1})\) with \(f(S(\Omega^{1}))\) and
 ~(b) \( S^{2}=S(\Omega^{2})\) with \(f(S(\Omega^{2}))\).

 In addition it is reasonable to examine a cost of changing
 a solution into another one:~
 \( H(S^{\alpha} \rightarrow  S^{\beta})\).
 Let \(\rho ( S^{\alpha}, S^{\beta} )\) be a proximity between solutions
  \( S^{\alpha}\) and \( S^{\beta}\),
  for example,
 \(\rho ( S^{\alpha}, S^{\beta} ) = | f(S^{\alpha}) -  f(S^{\beta}) |\).
 Note function \(f(S)\) is often a vector function.
 Finally, the restructuring problem can be examine as follows (a basic version):

~~

 Find a solution \( S^{*}\) while taking into account the
 following:

  (i) \( H(S^{1} \rightarrow  S^{*}) \rightarrow \min \),
%
  ~(ii) \(\rho ( S^{*}, S^{2} )  \rightarrow \min  \) ~(or constraint).

~~

 The corresponding basic optimization model is:~~
   \(\min \rho ( S^{*}, S^{2} ) ~~s.t.
 ~ H(S^{1} \rightarrow  S^{*})  \leq \widehat{h} \),
 where \(\widehat{h}\) is a constraint for cost of the solution
 change.
%
 In a simple case, this problem can be
 formulated as knapsack problem for selection of a subset of
 change operations \cite{lev11restr,lev15}:
%
%
 \[\max\sum_{i=1}^{n} c^{1}_{i} x_{i}
 ~~~ s.t.~ \sum_{i=1}^{n} a^{1}_{i} x_{i} ~\leq~ b^{1},
 ~~~ x_{i} \in \{0,1\}.  \]
 In the case of interconnections between change operations,
 it is reasonable to consider combinatorial synthesis problem
 (i.e., while taking into account compatibility between the operations).


 Now let us consider multicriteria restructuring problems.

 First, the initial combinatorial optimization problem can by a
 multicriteria one.
 As a result, a set of Pareto-efficient solutions have
 to be considered.

 Second, the proximity function  ~\(\rho (S^{*},S^{2}) \)~
 (or  ~\(\rho (S^{*j}, \{S^{21},S^{22},S^{23}\} \))
  can be examined as a vector function as well
 (analogically for the solution change cost).

 The situation will lead to a multicriteria restructuring problem
 (and to searching for Pareto-efficient solution(s))
 (Fig. 13):

~~

 Find a solution \( S^{*}\) while taking into account the
 following:

  (i) \( \overline{H}(S^{1} \rightarrow  S^{*}) \rightarrow \min \),
  ~(ii) \( \overline{ \rho} ( S^{*}, S^{2} )  \rightarrow \min  \) ~(or constraint).

~~

 The corresponding  multicriteria optimization is:~~~
   \(\min \overline{\rho} ( S^{*}, S^{2} ) ~~~s.t.
 ~ \overline{ H} (S^{1} \rightarrow  S^{*})  \leq \widehat{ \overline{h} }
 \),
 where vector \(\widehat{ \overline{ h}}\) is a vector constraint for cost of the solution
 change.
 In a simple case of the multicriteria restructuring,
  problem can be formulated as a multicriteria knapsack problem for selection of a subset of
 change operations:
 \[\max\sum_{i=1}^{n} \overline{c}^{1}_{i} x_{i}
 ~~~ s.t.~ \sum_{i=1}^{n} \overline{a}^{1}_{i} x_{i} ~\leq~ \overline{b}^{1},
 ~~~ x_{i} \in \{0,1\}.  \]
 In the case of interconnections between change operations,
 it is reasonable to consider combinatorial synthesis problem
 (i.e., while taking into account compatibility between the operations).


 In the case of ordinal estimates and/or multiset estimates,
  restructuring problems
 (i.e., searching for Pareto-efficient solution(s) at posets
 for \(\overline{H}\) and for \(\overline{ \rho}\)
  based on ordinal scale and/or multiset scale;
   as in Fig. 11, Fig. 12) are:

~~

 Find a solution \( S^{*}\) while taking into account the
 following:

  (i) \( \widetilde{H}(S^{1} \rightarrow  S^{*}) \rightarrow \min \),
  ~(ii) \( \widetilde{ \rho} ( S^{*}, S^{2} )  \rightarrow \min  \) ~(or constraint),

  where estimates of \(\widetilde{H}(S^{1}\) and \( \widetilde{ \rho}\)
 are based on ordinal and/or multiset scale (as in Fig. 11, Fig. 12).

 ~~

 The kinds of optimization problems are described in
 \cite{lev11restr,lev15}.

\begin{center}
\begin{picture}(80,53)
\put(07,00){\makebox(0,0)[bl]{Fig. 13. Multicriteria
 restructuring  \cite{lev11restr,lev15}}}

\put(00,05){\vector(1,0){80}}

\put(78.5,06.3){\makebox(0,0)[bl]{\(t\)}}


\put(0.6,18){\makebox(0,0)[bl]{\(S^{11}\)}}
\put(0.6,10){\makebox(0,0)[bl]{\(S^{12}\)}}

\put(5,17){\circle{1.7}} \put(5,15){\circle{1.7}}

\put(12,12.5){\makebox(0,0)[bl]{Initial}}
\put(12,09.5){\makebox(0,0)[bl]{Pareto-efficient}}
\put(12,06){\makebox(0,0)[bl]{solutions (\(t=\tau^{1}\))}}

\put(11,10){\line(-4,3){4}}

\put(6,16){\vector(1,1){23}}

\put(6,18){\line(3,1){21}} \put(27,25){\vector(2,1){07}}

\put(35,29){\circle{2}} \put(35,29){\circle*{1}}

\put(33,19){\makebox(0,0)[bl]{\(S^{*2}\)}}
\put(35,22){\line(0,1){5}}


\put(36.5,29.5){\vector(2,1){6}}
\put(42.5,32.5){\vector(-2,-1){6}}


\put(40,35){\circle*{2.7}} \put(43,35){\circle*{2.7}}
\put(46,35){\circle*{2.7}}

\put(59,42){\makebox(0,0)[bl]{Goal}}
\put(59,39){\makebox(0,0)[bl]{Pareto-efficient}}
\put(59,36){\makebox(0,0)[bl]{solutions}}
\put(59,32){\makebox(0,0)[bl]{(\(t=\tau^{2}\)):}}
\put(59,28){\makebox(0,0)[bl]{\(S^{21},S^{22},S^{23}\)}}

\put(58.5,35){\line(-1,0){10}}

\put(43,35){\oval(18,10)} \put(43,35){\oval(23,17)}
\put(43,35){\oval(30,22)}


\put(40,35){\vector(-2,1){9}} \put(30,40){\vector(2,-1){9}}

\put(30,48){\makebox(0,0)[bl]{Proximity \(\rho
(S^{*1},\{S^{21},S^{22},S^{23}\})\)}}

\put(34,48){\line(0,-1){9}}


\put(46,19){\makebox(0,0)[bl]{Neighborhoods }}
\put(46,15){\makebox(0,0)[bl]{of ~\(\{S^{21},S^{22},S^{23}\}\)}}

\put(56,22){\line(-1,1){7}}

\put(53,22){\line(-2,1){6}}


\put(30,40){\circle{2}} \put(30,40){\circle*{1}}

\put(11,47){\makebox(0,0)[bl]{Obtained}}
\put(11,44){\makebox(0,0)[bl]{solution \(S^{*1}\)}}

\put(20,43){\line(4,-1){7}}

\put(1,39){\makebox(0,0)[bl]{Solution }}
\put(1,35){\makebox(0,0)[bl]{change cost }}
\put(1,31){\makebox(0,0)[bl]{\(H(S^{12} \rightarrow S^{*1})\)}}

\put(6,31){\line(1,-1){7}}

\end{picture}
\end{center}

\subsection{Multi-stage restructuring}

 In multi-stage restructuring problems were illustrated
 in Fig. 7 and Fig. 8.
 Two basic trajectories are:

 (a) \(n\)-stage trajectory of optimal solutions:~
 \(\overline{S}^{opt} =
 <S^{0} \rightarrow S^{1} \rightarrow  S^{2} \rightarrow ... \rightarrow  S^{n} >\),

 (b) \(n\)-stage trajectory of restructured solutions:~
 \(\overline{S}^{rest} =
 <S^{0} \rightarrow S^{1*} \rightarrow  S^{2*} \rightarrow ... \rightarrow S^{n*}>\).

 As a result, the problem is:

~~

 Find Pareto-efficient solution(s) \( \overline{S}^{rest}\) while taking into account the
 following:

  (i) \( \overline{H} (\overline{S}^{rest} \rightarrow  \overline{S}^{opt}) \rightarrow \min  \),
  ~(ii) \( \overline{\rho} ( \overline{S}^{rest}, \overline{S}^{opt} )  \rightarrow \min  \)
   ~(or constraint),

 where ~~
  \( \overline{H} (\overline{S}^{rest} \rightarrow\overline{S}^{opt})
  =(
   H (S^{0} \rightarrow S^{1*}),
   H (S^{1*} \rightarrow S^{2*}),
   ...,
   H (S^{(n-1)*} \rightarrow S^{n*}))
   \),

 \(\overline{\rho} ( \overline{S}^{rest}, \overline{S}^{opt} )
  =(
   \rho (S^{1*}, S^{1}),
   \rho (S^{2*}, S^{2*}),
   ...,
    \rho (S^{(n*}, S^{n}))
   \).

~~

 Here, two corresponding simplified optimization models can be examined as the following:

 (a)   \(\min \overline{\rho} ( \overline{S}^{rest}, \overline{S}^{opt} ) ~~~~ s.t.
 ~~~ \overline{H} (\overline{S}^{rest} \rightarrow  \overline{S}^{opt})  \leq \widehat{\overline{h}}
 \),

 (b)   \(\min \overline{\rho} ( \overline{S}^{rest},\overline{S}^{opt})~~~~
 s.t.~~~ \overline{\rho} ( \overline{S}^{rest}, \overline{S}^{opt} )
  \leq \widehat{\overline{\rho}}\),
 where
 \(\widehat{\overline{h}} = (\widehat{h}_{1}, \widehat{h}_{2},...,\widehat{h}_{n})\)
 is a set (vector) of constraints for costs of the solution changes
 (i.e., a vector component corresponds to each stage),
 \(\widehat{\overline{\rho}} = (\widehat{\rho}_{1}, \widehat{\rho}_{2},...,\widehat{\rho}_{n})\)
 is a set (vector) of constraints for proximities of the solutions
 (i.e., a vector component corresponds to each stage).

 The following heuristic solving schemes (frameworks)
 can be considered:

~~~

 {\bf Scheme 1} (series solving process):

 {\it Step 1.} Solving the optimization problem at each stage 1
 (i.e., \(\tau_{1}\)):

 (1.1.) Finding the optimization solution \(S^{1}\)
 (basic optimization).

 (1.2.) Finding the restructuring solution \(S^{1*}\)
 (i.e., \(S^{0} \rightarrow S^{1*}\)).

 {\it Step 2.} Solving the optimization problem at each stage 2
 (i.e., \(\tau_{2}\)):

 (a) Finding the optimization solution \(S^{2}\)
 (basic optimization).

 (b) Finding the restructuring solution \(S^{2*}\)
 (i.e., \(S^{1*} \rightarrow S^{2*}\)).

  {\bf .~.~.}

 {\it Step n.} Solving the optimization problem at each stage 2
 (i.e., \(\tau_{n}\)):

 (a) Finding the optimization solution \(S^{n}\)
 (basic optimization).

 (b) Finding the restructuring solution \(S^{n*}\)
(i.e., \(S^{(n-1)*} \rightarrow S^{n*}\)).

~~~

 {\bf Scheme 2} (``composition'' solving process):

 {\it Step 1.} Solving the optimization problems:

 (1.1.) Finding the optimization solution \(S^{1}\)
 (basic optimization at stage 1) (i.e., \(\tau_{1}\)).

 (1.2.) Finding the optimization solution \(S^{2}\)
 (basic optimization at stage 2) (i.e., \(\tau_{2}\)).

  {\bf .~.~.}

 (1.n.) Finding the optimization solution \(S^{n}\)
 (basic optimization at stage n) (i.e., \(\tau_{n}\)).

 {\it Step 2.} Solving the one-stage restructuring problems for each
 stage to obtain several ``good'' solutions:

 (2.1.) Finding the ``good'' solutions at stage \(1\):~
  \(S^{0} \rightarrow \{ S^{1*}_{1}, S^{1*}_{2},...,S^{1*}_{q_{1}} \}\)).

 (2.2.) Finding the ``good'' solutions at stage \(2\):~
  \(\{ S^{1*}_{1}, S^{1*}_{2},...,S^{1*}_{q_{1}}\} \rightarrow \{ S^{2*}_{1}, S^{2*}_{2},...,S^{2*}_{q_{2}}\} \)).

 {\bf .~.~.}

 (2.n.) Finding the ``good' solutions at stage \(n\):~
 \(\{ S^{(n-1)*}_{1}, S^{(n-1)*}_{2},...,S^{(n-1)*}_{q_{n-1}}\} \rightarrow \{ S^{n*}_{1}, S^{n*}_{2},...,S^{n*}_{q_{n}}\} \)).

 {\it Step 3.} Composition of multi-stage restructuring solution trajectory
 (i.e., selection of a restructuring solution at each stage for solving the multi-stage restructuring problem above)
 (Fig. 14) (the initial point of the trajectory corresponds to
 \(S^{0}\)):~
 \(\overline{S}^{rest} =
 <S^{0} \rightarrow S^{1*}_{\xi_{1}} \rightarrow  S^{2*}_{\xi_{2}} \rightarrow ... \rightarrow S^{n*}_{\xi_{n}}>\),
 where
  \( \xi_{1} \in \{1,...,q_{1}\}\),
 \( \xi_{2} \in \{1,...,q_{2}\}\), ... ,
  \(\xi_{n} \in \{1,...,q_{n}\}\).

 ~~

 The solving scheme 3 extends scheme 2
 by finding several good solution trajectories and selection of
 the best final solution trajectory:

~~~

 {\bf Scheme 3} (``composition\&selection'' solving process):

 {\it Step 1.} Solving the optimization problems:

 (1.1.) Finding the optimization solution \(S^{1}\)
 (basic optimization at stage 1) (i.e., \(\tau_{1}\)).

 (1.2.) Finding the optimization solution \(S^{2}\)
 (basic optimization at stage 2) (i.e., \(\tau_{2}\)).

  {\bf .~.~.}

 (1.n.) Finding the optimization solution \(S^{n}\)
 (basic optimization at stage n) (i.e., \(\tau_{n}\)).

 {\it Step 2.} Solving the one-stage restructuring problems for each
 stage to obtain several ``good'' solutions
 (as in Scheme 2).

 {\it Step 3.} Composition of \(k (k >1)\)
 multi-stage restructuring solution trajectories:
 (e.g., selection of a restructuring solution at each stage for solving the multi-stage restructuring problem above)
 (Fig. 14)
 (the initial point of each trajectory corresponds to \(S^{0}\)):

 (3.1.)  \(\overline{S}^{rest}_{1} =
 <S^{0} \rightarrow S^{1*}_{\xi_{1}^{1}} \rightarrow  S^{2*}_{\xi_{2}^{1}} \rightarrow ... \rightarrow S^{n*}_{\xi_{n}^{1}}>\),

 where
  \( \xi_{1}^{1} \in \{1,...,q_{1}\}\),
 \( \xi_{2}^{1} \in \{1,...,q_{2}\}\), ... ,
  \(\xi_{n}^{1} \in \{1,...,q_{n}\}\);

 (3.2) \(\overline{S}^{rest}_{2} =
 <S^{0} \rightarrow S^{1*}_{\xi_{1}^{2}} \rightarrow  S^{2*}_{\xi_{2}^{2}} \rightarrow ... \rightarrow S^{n*}_{\xi_{n}^{2}}>\),

 where
  \( \xi_{1}^{2} \in \{1,...,q_{1}\}\),
 \( \xi_{2}^{2} \in \{1,...,q_{2}\}\), ... ,
  \(\xi_{n}^{2} \in \{1,...,q_{n}\}\);

 {\bf .~.~.}

 (3.k) \(\overline{S}^{rest}_{k} =
 <S^{0} \rightarrow S^{1*}_{\xi_{1}^{k}} \rightarrow  S^{2*}_{\xi_{2}^{k}} \rightarrow ... \rightarrow S^{n*}_{\xi_{n}^{k}}>\),

 where
  \( \xi_{1}^{k} \in \{1,...,q_{1}\}\),
 \( \xi_{2}^{k} \in \{1,...,q_{2}\}\), ... ,
  \(\xi_{n}^{k} \in \{1,...,q_{n}\}\).

 {\it Step 4.} Selection of the best restructuring trajectory
 \(\overline{S}^{rest}_{*}\) (Fig. 15):~
 \(\{ \overline{S}^{rest}_{1},\overline{S}^{rest}_{2},...,\overline{S}^{rest}_{k} \} \Longrightarrow  \overline{S}^{rest}_{*} \)

 ~~

\begin{center}
\begin{picture}(83,51)

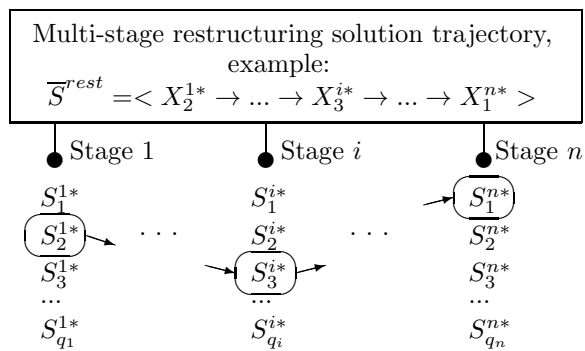
\put(00,00){\makebox(0,0)[bl] {Fig. 14. Composition of
 restructuring trajectory }}

\put(4,5){\makebox(0,8)[bl]{\(S^{1*}_{q_{1}}\)}}
\put(4,11){\makebox(0,8)[bl]{\(. . .\)}}
\put(4,13){\makebox(0,8)[bl]{\(S^{1*}_{3}\)}}
\put(4,18){\makebox(0,8)[bl]{\(S^{1*}_{2}\)}}
\put(4,23){\makebox(0,8)[bl]{\(S^{1*}_{1}\)}}
\put(06,20){\oval(8,5.6)}

\put(10,20){\vector(3,-1){4}}

\put(32,5){\makebox(0,8)[bl]{\(S^{i*}_{q_{i}}\)}}
\put(32,11){\makebox(0,8)[bl]{\(. . .\)}}
\put(32,13){\makebox(0,8)[bl]{\(S^{i*}_{3}\)}}
\put(32,18){\makebox(0,8)[bl]{\(S^{i*}_{2}\)}}
\put(32,23){\makebox(0,8)[bl]{\(S^{i*}_{1}\)}}
\put(34,15){\oval(8,5.6)}

\put(26,16){\vector(4,-1){4}} \put(38,15){\vector(4,1){4}}

\put(61,5){\makebox(0,8)[bl]{\(S^{n*}_{q_{n}}\)}}
\put(61,11){\makebox(0,8)[bl]{\(. . .\)}}
\put(61,13){\makebox(0,8)[bl]{\(S^{n*}_{3}\)}}
\put(61,18){\makebox(0,8)[bl]{\(S^{n*}_{2}\)}}
\put(61,23){\makebox(0,8)[bl]{\(S^{n*}_{1}\)}}
\put(63,25){\oval(8,5.6)}

\put(55,24){\vector(4,1){4}}


\put(6,30){\circle*{2}} \put(34,30){\circle*{2}}
\put(63,30){\circle*{2}}

\put(06,31){\line(0,1){4}} \put(34,31){\line(0,1){4}}
\put(63,31){\line(0,1){4}}

\put(08,29.5){\makebox(0,8)[bl]{Stage \(1\)}}
\put(36,29.5){\makebox(0,8)[bl]{Stage \(i\)}}
\put(64.5,29.5){\makebox(0,8)[bl]{Stage \(n\)}}


\put(16,20){\makebox(0,8)[bl]{ .~.~.}}
\put(44,20){\makebox(0,8)[bl]{ .~.~.}}


\put(00,35){\line(1,0){76}} \put(00,50){\line(1,0){76}}
\put(00,35){\line(0,1){15}} \put(76,35){\line(0,1){15}}

\put(03,45){\makebox(0,8)[bl]{Multi-stage restructuring solution
 trajectory,}}

\put(029,41){\makebox(0,8)[bl]{example:}}

\put(05,36.5){\makebox(0,8)[bl]{\(\overline{S}^{rest}=
 <X^{1*}_{2} \rightarrow ... \rightarrow X^{i*}_{3} \rightarrow ... \rightarrow X^{n*}_{1}>\)}}

\end{picture}
%
\begin{picture}(70,45)
\put(03.4,00){\makebox(0,0)[bl] {Fig. 15.
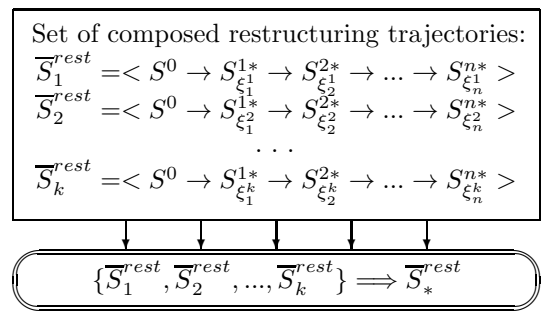
 Selection of the best trajectory }}

\put(15,17){\vector(0,-1){4}} \put(25,17){\vector(0,-1){4}}
\put(35,17){\vector(0,-1){4}} \put(45,17){\vector(0,-1){4}}
\put(55,17){\vector(0,-1){4}}



\put(00,17){\line(1,0){70}} \put(00,45){\line(1,0){70}}
\put(00,17){\line(0,1){28}} \put(70,17){\line(0,1){28}}

\put(02.5,40){\makebox(0,8)[bl]{Set of composed
 restructuring trajectories:}}

\put(03,33.5){\makebox(0,8)[bl]{\(\overline{S}^{rest}_{1} =
 <S^{0} \rightarrow S^{1*}_{\xi_{1}^{1}} \rightarrow  S^{2*}_{\xi_{2}^{1}} \rightarrow ... \rightarrow
 S^{n*}_{\xi_{n}^{1}}>\)}}


\put(03,28.5){\makebox(0,8)[bl]{\(\overline{S}^{rest}_{2} =
 <S^{0} \rightarrow S^{1*}_{\xi_{1}^{2}} \rightarrow  S^{2*}_{\xi_{2}^{2}} \rightarrow ... \rightarrow
 S^{n*}_{\xi_{n}^{2}}>\)}}

\put(31,26){\makebox(0,8)[bl]{ .~.~.}}

\put(03,19){\makebox(0,8)[bl]{\(\overline{S}^{rest}_{k} =
 <S^{0} \rightarrow S^{1*}_{\xi_{1}^{k}} \rightarrow  S^{2*}_{\xi_{2}^{k}} \rightarrow ... \rightarrow
 S^{n*}_{\xi_{n}^{k}}>\)}}


\put(35,09){\oval(70,08)} \put(35,09){\oval(69,07)}

\put(10.5,07){\makebox(0,8)[bl]{\(\{\overline{S}^{rest}_{1},\overline{S}^{rest}_{2},...,\overline{S}^{rest}_{k}
 \}
 \Longrightarrow  \overline{S}^{rest}_{*}\)}}

\end{picture}
\end{center}

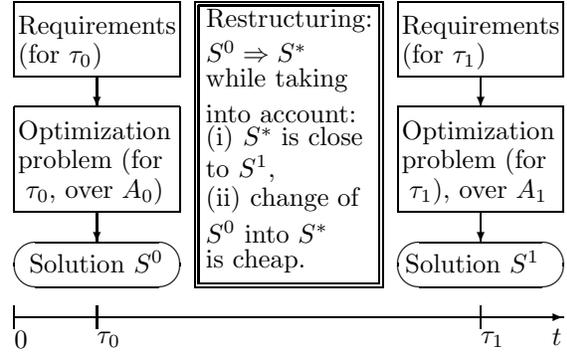
\subsection{Restructuring over changed element set}

 Let us consider restructuring over changed element set
 for knapsack problem (i.e., combinatorial optimization problem
 over one element set).
 The following element sets are examined (Fig. 16):
 ~(i) initial set \(A_{0}\),
 ~(ii) new set \(A_{1}\),
 ~(iii) added (new) set \(A^{+}\),
 ~(iv) deleted set
 \(A^{-} =
   A_{0} \backslash \{ A_{0} \bigcap A_{1} \} \),
   and
 ~(v) fixed (non-changed) element set
  \(\widehat{A}=\{ A_{0} \bigcap A_{1} \} \).

 Here, the restructuring problem is considered as a one-stage
 restructuring (Fig. 17):

~~

 Find a solution \( S^{*}\) while taking into account the
 following:

  (i) \( H(S^{1} \rightarrow  S^{*}) \rightarrow \min \),
  ~(ii) \(\rho ( S^{*}, S^{2} )  \rightarrow \min  \) ~(or constraint),

  where cost \(\rho ( S^{*}, S^{2} )\) involves the following components:

 (a) cost of deletion of elements
 \(A^{-} = A_{0} \backslash \{ A_{0} \bigcap A_{1} \} \),

 (b) cost of processing fixed elements
 \(\widehat{A}=\{ A_{0} \bigcap A_{1} \} \),

 (c) cost for processing new elements \(A^{+}\).

~~

 Thus, the correction problem (as a basic correction problem) is solved over
 elements
  \(\widehat{A}  \bigcup  A^{+}\)
 while taking into account cost of deletion of elements \(A^{-}\).
 The problem can be extended for multi-stage case.

\begin{center}
\begin{picture}(73,54)
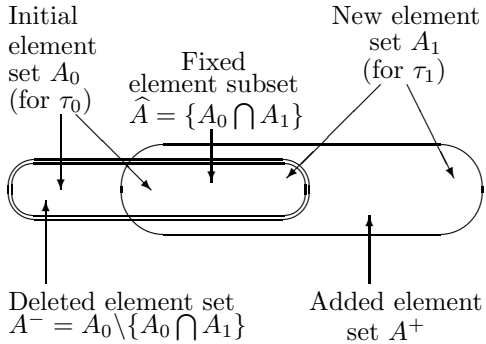
\put(02.6,00){\makebox(0,0)[bl]{Fig. 16. Illustration of changing
 sets}}

\put(07,36){\vector(0,-1){10}} \put(09,36){\vector(1,-1){10}}

\put(20,26){\oval(40,08)} \put(20,26){\oval(39,07)}

\put(00,48){\makebox(0,0)[bl]{Initial}}
\put(00,44){\makebox(0,0)[bl]{element}}
\put(00,40){\makebox(0,0)[bl]{set \(A_{0}\)}}
\put(00,36){\makebox(0,0)[bl]{(for \(\tau_{0}\))}}


\put(39,26){\oval(48,12)}

\put(43,48){\makebox(0,0)[bl]{New element }}
\put(48,44){\makebox(0,0)[bl]{set \(A_{1}\) }}
\put(47,40){\makebox(0,0)[bl]{(for \(\tau_{1}\))}}

\put(53,40){\vector(1,-2){06.25}}
\put(49.5,40){\vector(-1,-1){12.5}}


\put(23,42){\makebox(0,0)[bl]{Fixed}}

\put(16,39){\makebox(0,0)[bl]{element subset  }}
\put(16,34){\makebox(0,0)[bl]{\(\widehat{A}=\{ A_{0}
 \bigcap A_{1} \} \)}}

\put(27,34){\vector(0,-1){07}}


\put(40,10){\makebox(0,0)[bl]{Added element}}
\put(45,06){\makebox(0,0)[bl]{set \(A^{+}\) }}

\put(48,13.5){\vector(0,1){09}}


\put(00,10){\makebox(0,0)[bl]{Deleted element set}}
\put(00,06){\makebox(0,0)[bl]{\(A^{-} =
   A_{0} \backslash \{ A_{0} \bigcap A_{1} \} \) }}

\put(05,13.5){\vector(0,1){11}}

\end{picture}
%
\begin{picture}(73,54)
\put(00,00){\makebox(0,0)[bl]{Fig. 17.
 Restructuring over changed element set}}

\put(00,09){\vector(1,0){73}}

\put(00,7.5){\line(0,1){3}} \put(11,7.5){\line(0,1){3}}
\put(62,7.5){\line(0,1){3}}

\put(00,05){\makebox(0,0)[bl]{\(0\)}}
\put(11,05){\makebox(0,0)[bl]{\(\tau_{0}\)}}
\put(62,05){\makebox(0,0)[bl]{\(\tau_{1}\)}}

\put(71.5,05.3){\makebox(0,0)[bl]{\(t\)}}


\put(00,41){\line(1,0){22}} \put(00,51){\line(1,0){22}}
\put(00,41){\line(0,1){10}} \put(22,41){\line(0,1){10}}

\put(0.5,46){\makebox(0,0)[bl]{Requirements}}
\put(0.5,42){\makebox(0,0)[bl]{(for \(\tau_{0}\))}}


\put(11,41){\vector(0,-1){4}}

\put(00,23){\line(1,0){22}} \put(00,37){\line(1,0){22}}
\put(00,23){\line(0,1){14}} \put(22,23){\line(0,1){14}}

\put(0.5,32){\makebox(0,0)[bl]{Optimization}}
\put(0.5,28){\makebox(0,0)[bl]{problem (for}}
\put(1.5,24){\makebox(0,0)[bl]{\(\tau_{0}\),
 over \(A_{0}\)) }}


\put(11,23){\vector(0,-1){4}}

\put(11,16){\oval(22,06)}

\put(02,15){\makebox(0,0)[bl]{Solution \(S^{0}\)}}


\put(24,13){\line(1,0){25}} \put(24,51){\line(1,0){25}}
\put(24,13){\line(0,1){38}} \put(49,13){\line(0,1){38}}

\put(24.5,13.5){\line(1,0){24}} \put(24.5,50.5){\line(1,0){24}}
\put(24.5,13.5){\line(0,1){37}} \put(48.5,13.5){\line(0,1){37}}

\put(25.5,47){\makebox(0,0)[bl]{Restructuring:}}
\put(25.5,43){\makebox(0,0)[bl]{\(S^{0} \Rightarrow S^{*}  \)}}
\put(25.5,39){\makebox(0,0)[bl]{while taking}}
\put(25.5,35){\makebox(0,0)[bl]{into account:}}
\put(25.5,31){\makebox(0,0)[bl]{(i) \(S^{*}\) is close}}
\put(25.5,27){\makebox(0,0)[bl]{to \(S^{1}\),}}
\put(25.5,23){\makebox(0,0)[bl]{(ii) change of }}
\put(25.5,19){\makebox(0,0)[bl]{\(S^{0}\) into \(S^{*}\)}}
\put(25.5,15){\makebox(0,0)[bl]{is cheap.}}


\put(51,41){\line(1,0){22}} \put(51,51){\line(1,0){22}}
\put(51,41){\line(0,1){10}} \put(73,41){\line(0,1){10}}

\put(51.5,46){\makebox(0,0)[bl]{Requirements}}
\put(51.5,42){\makebox(0,0)[bl]{(for \(\tau_{1}\))}}


\put(62,41){\vector(0,-1){4}}

\put(51,23){\line(1,0){22}} \put(51,37){\line(1,0){22}}
\put(51,23){\line(0,1){14}} \put(73,23){\line(0,1){14}}

\put(51.5,32){\makebox(0,0)[bl]{Optimization}}
\put(51.5,28){\makebox(0,0)[bl]{problem (for}}
\put(52.5,24){\makebox(0,0)[bl]{\(\tau_{1})\), over \(A_{1}\)}}


\put(62,23){\vector(0,-1){4}}

\put(62,16){\oval(22,06)}

\put(52,15){\makebox(0,0)[bl]{Solution \(S^{1}\)}}

\end{picture}
\end{center}
%

\section{Restructuring in Combinatorial Optimization Problems }

\subsection{Knapsack problem}

 Let us present the restructuring approach for basic knapsack problem
 from \cite{lev11restr}.
 Let \(A=\{1,...,i,...,n\}\) be a basic initial set of elements.
  Knapsack problem is considered
  for two time moments \(\tau_{0}\) and \(\tau_{1}\)
 (for \(\tau_{1}\) parameters
 \(\{c^{1}_{i}\}\), \(\{a^{1}_{i}\}\), and \(b^{1}\) are used):
 \[\max\sum_{i=1}^{n} c^{0}_{i} x_{i}
 ~~~ s.t.~ \sum_{i=1}^{n} a^{0}_{i} x_{i} ~\leq~ b^{0},
 ~~~ x_{i} \in \{0,1\}.  \]
 The corresponding solutions are:
 \(S^{0} \subseteq A \)
 (\(t=\tau_{0}\))
 and
 \(S^{1} \subseteq A \)
 (\(t=\tau_{1}\))
 (\(S^{0} \neq S^{1}\)).
%


 {\bf Illustrative numerical example is:}~
 \(A = \{1,2,3,4,5,6,7\}\),
 \(S^{0} = \{1,3,4,5\}\),
 \(S^{1} = \{2,3,5,7\}\),
  \(S^{*} = \{2,3,4,6\}\).
  The change (restructuring) process (i.e., \(S^{0} \Rightarrow S^{*}\)) is based
  on the following (Fig. 18):
 (a) deleted elements:
 \( S^{*-} =  S^{0} \backslash S^{*} =  \{1,5\}\),
 (b) added elements:
  \(S^{*+} = S^{*} \backslash S^{0} = \{2,6\}\).

\begin{center}
\begin{picture}(80,37)

\put(013.4,00){\makebox(0,0)[bl]{Fig. 18. Example for
restructuring}}

\put(00,10){\vector(1,0){80}}

\put(00,8.5){\line(0,1){3}} \put(8,8.5){\line(0,1){3}}
\put(72,8.5){\line(0,1){3}}

\put(00,06){\makebox(0,0)[bl]{\(0\)}}
\put(8,06){\makebox(0,0)[bl]{\(\tau_{0}\)}}
\put(71,06){\makebox(0,0)[bl]{\(\tau_{1}\)}}

\put(79,06.3){\makebox(0,0)[bl]{\(t\)}}


\put(16,25){\vector(1,1){4}} \put(16,23){\vector(1,-1){4}}

\put(8,24){\oval(16,10)}

\put(3.5,25){\makebox(0,0)[bl]{\(S^{0} =\)}}
\put(0.5,21){\makebox(0,0)[bl]{\(\{1,3,4,5\}\)}}


\put(20,26){\line(1,0){21}} \put(20,36){\line(1,0){21}}
\put(20,26){\line(0,1){10}} \put(41,26){\line(0,1){10}}

\put(22,32){\makebox(0,0)[bl]{Deletion of }}
\put(20.4,27.5){\makebox(0,0)[bl]{\(S^{*-}=\{1,5\}\)}}


\put(20,12){\line(1,0){21}} \put(20,22){\line(1,0){21}}
\put(20,12){\line(0,1){10}} \put(41,12){\line(0,1){10}}


\put(21.4,18){\makebox(0,0)[bl]{Addition of }}
\put(20.4,13.5){\makebox(0,0)[bl]{\(S^{*+}=\{2,6\}\)}}


\put(41,29){\vector(1,-1){4}} \put(41,19){\vector(1,1){4}}



\put(54,24){\oval(16,10)} \put(54,24){\oval(17,11)}

\put(49.5,25){\makebox(0,0)[bl]{\(S^{*} =\)}}
\put(46.5,21){\makebox(0,0)[bl]{\(\{2,3,4,6\}\)}}


\put(72,24){\oval(16,10)}

\put(67.5,25){\makebox(0,0)[bl]{\(S^{1} =\)}}
\put(64.5,21){\makebox(0,0)[bl]{\(\{2,3,5,7\}\)}}


\end{picture}
\end{center}

 Note the following exists
 at the start stage of the solving process:
 ~\( S^{*-} = S^{0} \) and
 \( S^{*+} = A \backslash  S^{0}   \).
 The restructuring problem can be considered as the following:
  \[\min \rho ( S^{*} , S^{1})
  ~~~s.t. ~~
   H(S^{0} \Rightarrow S^{*}) = ( \sum_{i \in  S^{*-} } h^{-}_{i} +  \sum_{i \in S^{*+}} h^{+}_{i} ~) \leq \widehat{h},
   ~\sum_{i \in S^{*}}  a^{2}_{i} \leq b^{2}, \]
 where \(\widehat{h}\) is a constraint for the change cost,
 \(h^{-}(i)\) is a cost of deletion of element \(i \in A\), and
  \(h^{+}(i)\) is a cost of addition of element \(i \in A\).
  On the other hand, an equivalent problem can be examined:
 \[\max \sum_{i \in S^{*}} x_{i} c_{i}^{1}
  ~~~s.t. ~~~
  H ( S^{0} \Rightarrow S^{*} ) =
  ( \sum_{i \in S^{*-}} h^{-}_{i} + \sum_{i \in S^{*+}} h^{+}_{i} ~) \leq \widehat{h},
   ~\sum_{i \in S^{*} }  a^{1}_{i} \leq b^{1},\]
 because
  ~\( \max \sum_{i \in S^{*}} x_{i} c_{i}^{1}
   \leq
     \max \sum_{i \in {S}^{2}} x_{i} c_{i}^{1} \)~
 while taking into account constraint:
   ~\( \sum_{i \in {S}^{*}}  a^{1}_{i} \leq b^{1} \).
%
%
  The obtained problem is a modified knapsack-like problem as well.
  At the same time, it is possible to use a simplified solving
  scheme (by analysis of {\it change elements} for addition/deletion):
 (a) generation  of candidate elements for deletion
 (i.e., selection of \(S^{*-}\) from \(S^{0}\)),
 (b) generation of candidate elements for addition
 (i.e., selection of \(S^{*+}\) from
 \( A \backslash  S^{0}  \)).
 The selection processes may be based on multicriteria ranking.
 As a result, a problem with sufficiently
 decreased dimension will be obtained.


 In the case of multicriteria knapsack problem,
 the restructuring process is the same
 (i.e., selection of deletion and addition operations).
 Thus,
 the restructuring problem can be examined as
 multicriteria knapsack problem.
 Analogical situation exists in the case of
 ordinal or multiset  estimates
  \cite{lev12a,lev15}.


 {\bf Applied three-stage example}
 for three-stage restructuring
 (\(t\in\{\tau_{0},\tau_{1},\tau_{2}  \}\))
  of modular educational course  is considered
  (Table 3, educational topics/items
 are  \(A = \{1,...,i,...,13\}\)).

\begin{center}
 {\bf Table 3.} Components of modular course on combinatorial optimization, their parameters   \\
\begin{tabular}{| c | l | l c c | l c c c c |l c c c c |}
\hline
 \(i\)&Topic (item)&\(\tau_{0}:\)&\(c_{i}^{0}\)&\(a_{i}^{0}\)
  &\(\tau_{1}:\)&\(c_{i}^{1}\)&\(a_{i}^{1}\)&\(h_{i}^{1-}\)&\(h_{i}^{1+}\)
  &\(\tau_{2}:\)&\(c_{i}^{2}\)&\(a_{i}^{2}\)&\(h_{i}^{2-}\)&\(h_{i}^{2+}\)\\

\hline

 1.& Complexity,
  &&\(4.0\)&\(1.5\)
      &&\(5.0\)&\(2.0\)&\(0.5\)&\(1.0\)
      &&\(5.0\)&\(2.0\)&\(0.5\)&\(1.0\)\\

  &algorithms
  & &&&&&&&&&&&&\\

 2.& Knapsack
  &&\(4.0\)&\(3.0\)
      &&\(5.0\)&\(4.0\)&\(0.6\)&\(1.5\)
      &&\(5.0\)&\(4.0\)&\(0.6\)&\(1.5\)\\

 3.& Routing
  &&\(1.0\)&\(3.5\)
      &&\(5.0\)&\(4.0\)&\(0.8\)&\(1.4\)
      &&\(6.0\)&\(5.0\)&\(0.8\)&\(1.4\)\\

 4.& Assignment/
   &&\(4.0\)&\(2.5\)
      &&\(5.0\)&\(3.0\)&\(0.7\)&\(1.5\)
      &&\(5.0\)&\(3.0\)&\(0.7\)&\(1.5\)\\

 &allocation
  & &&&&&&&&&&&&\\

 5.& Scheduling
    &&\(1.5\)&\(5.0\)
      &&\(2.0\)&\(4.0\)&\(0.4\)&\(1.5\)
      &&\(3.0\)&\(5.0\)&\(0.4\)&\(1.5\)\\

 6.& Packing
    &&\(1.0\)&\(3.0\)
      &&\(3.0\)&\(3.0\)&\(0.3\)&\(1.0\)
      &&\(3.5\)&\(4.0\)&\(0.3\)&\(1.0\)\\

 7.& Covering
    &&\(1.0\)&\(1.5\)
      &&\(2.5\)&\(2.0\)&\(0.4\)&\(1.0\)
      &&\(3.0\)&\(3.5\)&\(0.4\)&\(1.0\)\\

 8.& Spanning trees
    &&\(3.0\)&\(2.0\)
      &&\(4.0\)&\(2.5\)&\(1.0\)&\(1.5\)
           &&\(5.0\)&\(3.5\)&\(1.0\)&\(1.5\)\\

 9.& Clique-based
    &&\(1.0\)&\(1.5\)
      &&\(1.5\)&\(2.0\)&\(0.3\)&\(0.8\)
      &&\(2.0\)&\(2.0\)&\(0.3\)&\(0.8\)\\

 10.& Graph coloring
    &&\(1.0\)&\(1.5\)
      &&\(2.0\)&\(1.7\)&\(0.2\)&\(0.7\)
      &&\(2.5\)&\(2.0\)&\(0.2\)&\(0.7\)\\

 11.& Clustering,
   &&\(4.0\)&\(2.0\)
      &&\(5.0\)&\(2.5\)&\(1.0\)&\(1.7\)
      &&\(5.5\)&\(2.5\)&\(1.0\)&\(1.7\)\\

 &sorting
 &&&&&&&&&&&&&\\

 12.& Alignment
    &&\(1.0\)&\(0.8\)
      &&\(0.9\)&\(1.0\)&\(0.2\)&\(0.3\)
      &&\(0.9\)&\(1.0\)&\(0.2\)&\(0.3\)\\

 13.& Satisfiability
   &&\(2.0\)&\(2.0\)
      &&\(1.5\)&\(2.0\)&\(0.2\)&\(0.3\)
      &&\(1.0\)&\(2.0\)&\(0.2\)&\(0.3\)\\

\hline
\end{tabular}
\end{center}

 The following parameters of each item \(i\) are examined
  (\(\gamma\) is the number of stage, \(\gamma=0,1,2\)):
  (a) profit (utility) \(c^{\gamma}_{i}\),
  (b) required resource \(a^{\gamma}_{i}\),
  (c) cost of deletion for item \(i\)
   \(h^{\gamma -}_{i}\),
 (d) cost of addition
   \(h^{\gamma +}_{i}\).

 First, knapsack problems for each stage
 (\(\gamma=0,1,2\)) are considered
  (\(b^{0} = 14\), \(b^{1} = 20\), \(b^{2} = 23\)):
 \[\max \sum_{i=1}^{13} c^{\gamma}_{i} x^{\gamma}_{i} ~~s.t. ~ \sum _{i=1}^{13} a^{\gamma}_{i} \leq b^{\gamma}\]
 The obtained
 solutions are:
 \(S^{0}=\{1,2,4,8,11,12,13\}\), \(c(S^{0})=22.0\), \(b(S^{0})=13.8\);
 \(S^{1}=\{1,2,4,8,10,11\}\), \(c(S^{1})=31.0\), \(b(S^{1})=19.7\);
 \(S^{2}=\{1,2,3,4,7,8,11\}\), \(c(S^{2})=29.5\), \(b(S^{2})=22.5\).
 Note, the assumption is:
  items \( B =\{1,2,4,8,11 \}\) are included in solutions at each stage.
  Thus, set \(\widehat{A} = \{3,5,6,7,9,10,12,13\}\) is under change process.
 Further,
 two series restructuring problems
  (as deletion/addition knapsack problems) are examined (as in {\it Scheme 1}):
  \(S^{0} \rightarrow  S^{*1}\) and
  \(S^{*1} \rightarrow  S^{*2}\).
 The local restructuring problem
 for \(\tau_{1}\) is
 (\(\rho (S^{*1},S^{1})=|c(S^{*1}) - c(S^{*1})|\), \(D^{1} = 1.6\) is constraint for total change
 cost):
\[ \min \rho (S^{*1},S^{1})
 ~~~~ s.t. ~
 (\sum_{ i \in  (\widehat{A} \bigcap S^{0}) }   h^{1-}_{i}+
 \sum_{ i \in (\widehat{A} \backslash (\widehat{A} \bigcap S^{0})) }   h^{1+}_{i})
  \leq D^{1},
 ~~\sum_{i\in (B \bigcup ( \widehat{A} \backslash (\widehat{A} \bigcap S^{0}) ))} a_{i}
  \leq b^{1}.
 \]
 The examined restructuring solutions are
 (problem for \(\tau_{2}\) is analogical, \(D^{2} = 1.6\)):

 \(S^{*1}=\{1,2,3,4,8,11\}\), \(c(S^{*1})=29.0\), \(b(S^{*1})=19.0\);

 \(S^{*2}=\{1,2,3,4,8,11\}\), \(c(S^{*2})=29.5\), \(b(S^{*2})=22.5\) (here,  \(S^{*2}= S^{2}\)).

 The final trajectory is:~
 \(\overline{S}^{rest} =  <S^{0},S^{*1},S^{*2}>\).

\subsection{Multiple choice problem}

 The description of restructuring for multiple choice problem
 is based on \cite{lev11restr}
 (\(t = \{\tau_{1},\tau_{2}\}\)).
 Basic multiple choice problem is for \(t=\tau_{1}\)
 (for \(t=\tau_{2}\) parameters \(\{c^{2}_{ij}\}\), \(\{a^{2}_{ij}\}\), and \(b^{2}\)
 are used):
 \[\max\sum_{i=1}^{m} \sum_{j=1}^{q_{i}} c^{1}_{ij} x_{ij}
 ~~~s.t.~\sum_{i=1}^{m} \sum_{j=1}^{q_{i}} a^{1}_{ij} x_{ij} \leq b^{1},
 ~\sum_{j=1}^{q_{i}} x_{ij} \leq 1 ~~ \forall i=\overline{1,m},
 ~~~x_{ij} \in \{0,1\}.\]
 Here initial element set  \(A\) is divided into
 \(m\) subsets (without intersection):
 ~\(A = \bigcup_{i=1}^{m}  A_{i}\), where  ~\(A_{i} =
 \{1,...,j,...,q_{i}\}\) (\(i=\overline{1,m}\)).
 Thus, each element is denoted by \((i,j)\).
 An equivalent problem is:
 \[ \max \sum_{(i,j) \in S^{1}} c_{ij}^{1}~~~~~ s.t.  \sum_{(i,j) \in S^{1}} a_{ij}^{1} \leq b^{1},
  ~~ | S^{1} \bigcap A_{i} | \leq 1 ~ \forall i =  \overline{1,m}. \]
 For \(t=\tau_{2}\) the problem is the same.

 {\bf Illustrative numerical example:}~
 \(A = \{1,2,3,4,5,6,7,8,9,10,11,12,13\}\),
 \(A_{1} = \{1,3,5,12\}\), \(A_{2} = \{2,7,9\}\),
 \(A_{3} = \{4,8,13\}\), \(A_{4} = \{6,10,11\}\),
 \(S^{1} = \{1,7,8,11\}\),
 \(S^{2} = \{3,7,8,10\}\),
  \(S^{*} = \{1,2,8,6\}\).
  The change (restructuring) process (i.e., \(S^{1} \Rightarrow S^{*}\)) is based
  on the following (Fig. 6):
 (a) deleted elements:
 \( S^{1*-} =  S^{1} \backslash S^{*} =  \{7,11\}\),
 (b) added elements:
  \(S^{1*+} = S^{*} \backslash S^{1} = \{2,6\}\).

 Further, the restructuring problem can be considered as the following:
  \[\min \rho ( S^{*} , S^{2})\]
  \[s.t. ~~~
   H (S^{1} \Rightarrow S^{*}) = ( \sum_{(i,j) \in  S^{1*-} } h^{-}_{ij} +  \sum_{(i,j) \in S^{1*+}} h^{+}_{ij} ~~) \leq \widehat{h},
   ~~\sum_{(i,j) \in S^{*}}  a^{2}_{ij} \leq b^{2},
%
 ~~ | S^{*} \bigcap A_{i} | \leq 1 ~ \forall i =  \overline{1,m}.
   \]
 where \(\widehat{h}\) is a constraint for the change cost,
 \(h^{-}(ij)\) is a cost of deletion of element \((i,j) \in A\), and
  \(h^{+}(ij)\) is a cost of addition of element \((i,j) \in A\).
 An equivalent problem is:
  \[\max  \sum_{(i,j) \in S^{*}}  c^{2}_{ij} \]
  \[s.t. ~~~
   H (S^{1} \Rightarrow S^{*}) = ( \sum_{(i,j) \in  S^{1*-} } h^{-}_{ij} +  \sum_{(i,j) \in S^{1*+}} h^{+}_{ij} ~~) \leq \widehat{h},
   ~~\sum_{(i,j) \in S^{*}}  a^{2}_{ij} \leq b^{2},
%
 ~~ | S^{*} \bigcap A_{i} | \leq 12 ~ \forall i =  \overline{1,m}.
   \]

 In the case of multicriteria multiple choice problem,
 the restructuring process is the same
 (i.e., selection of deletion and addition operations).
 Thus,
 the restructuring problem can be examined as
 multicriteria multiple choice  problem.
 Analogical situation exists in
 the case of the usage of ordinal or multiset-based estimates.
 Here,
 the corresponding restructuring multiple choice problem
 is based on multi-sate estimates
 (as in \cite{lev12a,lev15}).

 Further, a realistic applied  example for configuration of modular system
  is examined (from \cite{lev11restr}).


  {\bf Applied example.} Reconfiguration of ``microelectronic components
  part''
  in wireless sensor (multiple choice problem)
 ~\(M = R \star P \star D \star Q\) \cite{levfim10}:

    {\it 1.} Radio~  \(R\):~
    10 mw 916 MHz Radio~  \(R_{1}(3)\),
    1 mw 916 MHz Radio~  \(R_{2}(2)\),
    10 mw 600 MHz Radio~  \(R_{3}(2)\),
    1 mw 600 MHz Radio~ \(R_{4}(1)\).

    {\it 2.} Microprocessor~  \(P\):~
    MAXQ 2000~ \(P_{1}(1)\),
    AVR with embedded DAC/ ADC~  \(P_{2}(2)\),
    MSP~ \(P_{3}(3)\).

    {\it 3.} DAC/ADC~  \(D\):~
    Motorola~  \(D_{1}(2)\),
    AVR embedded DAC/ADC~  \(D_{2}(1)\),
    Analog Devices 1407~  \(D_{3}(2)\).

    {\it 4.} Memory~  \(Q\):~
     512 byte RAM~  \(Q_{1}(3)\),
     512 byte EEPROM~  \(Q_{2}(3)\),
     8 KByte Flash~ \(Q_{3}(2)\),
     1 MByte Flash~  \(Q_{4}(1)\).

 Here it is assumed that solutions
 are based on multiple choice problem
 (in \cite{levfim10} the solving process was based on morphological clique problem
 while taking into account compatibility of selected DAs).
 Thus,
 two solutions \(M^{1}\) (for \(t=\tau_{1}\), Fig. 19) and
 \(M^{2}\) (for \(t=\tau_{2}\), Fig. 20) are examined
 (in \cite{levfim10} the solutions correspond to trajectory
 design: stage 1 and stage 3).
 Table 4 contains estimates of DAs (expert judgment).
 Estimates of cost (Table 4) and priorities (Fig. 19, Fig. 20, in parentheses)
 correspond to examples in \cite{levfim10}.
 Here \(c_{ij} = 4-p_{ij}\).
 Two possible change operations can be considered
  (\( M^{1} \Rightarrow M^{*}\), \(M^{*}\) is close to \(M^{2}\) ):

 (a) \(R_{4} \rightarrow R_{2}  \),
 \(h_{a}^{-} = 2\), \(h_{a}^{+} = 1\)
 (corresponding Boolean variable \(x_{a} \in \{0,1\}\)),

 (b) \(Q_{4} \rightarrow Q_{1}  \),
 \(h_{b}^{-} = 1\), \(h_{b}^{+} = 1\)
 (corresponding Boolean variable \(x_{b} \in \{0,1\}\)).

 As a result, the following simplified knapsack problem can be used:
   \[  \max ~( ~(c^{2}(R_{2}) - c^{2}(R_{4}) )~ x_{a} +  (c^{2}(Q_{1}) - c^{2}(Q_{4}) ) ~x_{b}    ~ ) \]
   \[s.t. ~~ H (M^{*} \rightarrow M^{2})  =  (h^{-} (R_{4} \rightarrow R_{2}) + h^{+} (R_{4} \rightarrow R_{2} )) ~x_{a}
   +
    (h^{-}(Q_{4} \rightarrow Q_{1} )  + h^{+}(Q_{4} \rightarrow Q_{1} ))~ x_{b}  \leq
   \widehat{h}.
   \]

 Finally, the restructuring solutions are:
 ~(i) \(\widehat{h} = 2\):
 ~\(M^{*1} = R_{4} \star P_{2} \star D_{2} \star Q_{1}\),
 ~(ii) \(\widehat{h} = 3\):
 ~\(M^{*2} = R_{2} \star P_{2} \star D_{2} \star Q_{4}\),
 ~(iii) \(\widehat{h} = 5\):
 ~\(M^{*3} = M^{2} = R_{2} \star P_{2} \star D_{2} \star Q_{1}\).
 Evidently, real restructuring problems can be more complicated.

\begin{center}
 {\bf Table 4.} Estimates of DAs \\
\begin{tabular}{| c| c | l cc |l c c|}
\hline
 DAs  & Cost (\(a_{ij}\))& Change costs: & \(h^{-}_{ij}\)& \(h^{-}_{ij}\)&
    Priorities: & \(c^{1}_{ij}\)& \(c^{2}_{ij}\)\\
\hline

 \(R_{1}\)&6 &&2&2&&  1&1\\
 \(R_{2}\)&5 &&1&1&&  2&3\\
 \(R_{3}\)&3 &&2&1&&  2&1\\
 \(R_{4}\)&2 &&2&2&&  3&2\\

 \(P_{1}\)&5 &&2&3&&  3&2\\
 \(P_{2}\)&10&&2&2&&  2&3\\
 \(P_{3}\)&30&&3&2&&  1&2\\

 \(D_{1}\)&2 &&2&3&&  2&3\\
 \(D_{2}\)&1 &&2&2&&  3&2\\
 \(D_{3}\)&2 &&1&1&&  2&1\\

 \(Q_{1}\)&3 &&2&1&&  1&3\\
 \(Q_{2}\)&2 &&2&2&&  1&3\\
 \(Q_{3}\)&3 &&1&2&&  2&2\\
 \(Q_{4}\)&3 &&1&1&&  3&2\\

\hline
\end{tabular}
\end{center}

\begin{center}
\begin{picture}(56,33)

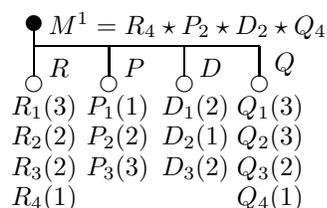
\put(00,00){\makebox(0,0)[bl]{Fig. 19. Structure of \(M^{1}\)}}

\put(01,17){\makebox(0,8)[bl]{\(R_{1}(3)\)}}
\put(01,13){\makebox(0,8)[bl]{\(R_{2}(2)\)}}
\put(01,09){\makebox(0,8)[bl]{\(R_{3}(2)\)}}
\put(01,05){\makebox(0,8)[bl]{\(R_{4}(1)\)}}

\put(11,17){\makebox(0,8)[bl]{\(P_{1}(1)\)}}
\put(11,13){\makebox(0,8)[bl]{\(P_{2}(2)\)}}
\put(11,09){\makebox(0,8)[bl]{\(P_{3}(3)\)}}

\put(21,17){\makebox(0,8)[bl]{\(D_{1}(2)\)}}
\put(21,13){\makebox(0,8)[bl]{\(D_{2}(1)\)}}
\put(21,09){\makebox(0,8)[bl]{\(D_{3}(2)\)}}

\put(31,17){\makebox(0,8)[bl]{\(Q_{1}(3)\)}}
\put(31,13){\makebox(0,8)[bl]{\(Q_{2}(3)\)}}
\put(31,09){\makebox(0,8)[bl]{\(Q_{3}(2)\)}}
\put(31,05){\makebox(0,8)[bl]{\(Q_{4}(1)\)}}


\put(04,27){\line(1,0){30}}

\put(04,27){\line(0,-1){04}} \put(14,27){\line(0,-1){04}}
\put(24,27){\line(0,-1){04}} \put(34,27){\line(0,-1){04}}


\put(04,22){\circle{2}} \put(14,22){\circle{2}}
\put(24,22){\circle{2}} \put(34,22){\circle{2}}

\put(06,23){\makebox(0,8)[bl]{\(R\) }}
\put(16,23){\makebox(0,8)[bl]{\(P\) }}
\put(26,23){\makebox(0,8)[bl]{\(D\) }}
\put(36,23){\makebox(0,8)[bl]{\(Q\) }}


\put(04,27){\line(0,1){3}}

\put(04,30){\circle*{2}}

\put(06,28){\makebox(0,8)[bl]{\(M^{1} = R_{4} \star P_{2} \star
D_{2} \star Q_{4}\) }}

\end{picture}
%
\begin{picture}(46,33)

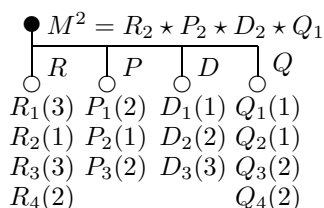
\put(00,0){\makebox(0,0)[bl]{Fig. 20. Structure of  \(M^{2}\)}}

\put(01,17){\makebox(0,8)[bl]{\(R_{1}(3)\)}}
\put(01,13){\makebox(0,8)[bl]{\(R_{2}(1)\)}}
\put(01,9){\makebox(0,8)[bl]{\(R_{3}(3)\)}}
\put(01,5){\makebox(0,8)[bl]{\(R_{4}(2)\)}}

\put(11,17){\makebox(0,8)[bl]{\(P_{1}(2)\)}}
\put(11,13){\makebox(0,8)[bl]{\(P_{2}(1)\)}}
\put(11,9){\makebox(0,8)[bl]{\(P_{3}(2)\)}}

\put(21,17){\makebox(0,8)[bl]{\(D_{1}(1)\)}}
\put(21,13){\makebox(0,8)[bl]{\(D_{2}(2)\)}}
\put(21,9){\makebox(0,8)[bl]{\(D_{3}(3)\)}}

\put(31,17){\makebox(0,8)[bl]{\(Q_{1}(1)\)}}
\put(31,13){\makebox(0,8)[bl]{\(Q_{2}(1)\)}}
\put(31,9){\makebox(0,8)[bl]{\(Q_{3}(2)\)}}
\put(31,5){\makebox(0,8)[bl]{\(Q_{4}(2)\)}}


\put(04,27){\line(1,0){30}}

\put(04,27){\line(0,-1){04}} \put(14,27){\line(0,-1){04}}
\put(24,27){\line(0,-1){04}} \put(34,27){\line(0,-1){04}}


\put(04,22){\circle{2}} \put(14,22){\circle{2}}
\put(24,22){\circle{2}} \put(34,22){\circle{2}}

\put(06,23){\makebox(0,8)[bl]{\(R\) }}
\put(16,23){\makebox(0,8)[bl]{\(P\) }}
\put(26,23){\makebox(0,8)[bl]{\(D\) }}
\put(36,23){\makebox(0,8)[bl]{\(Q\) }}


\put(04,27){\line(0,1){3}}

\put(04,30){\circle*{2}}




\put(06,28){\makebox(0,8)[bl]{\(M^{2} = R_{2} \star P_{2} \star
D_{2} \star Q_{1}\) }}

\end{picture}
\end{center}

\subsection{Assignment problem}

 The description of restructuring for assignment  problem
 is based on \cite{lev11restr}
 (\(t = \{\tau_{1},\tau_{2}\}\)).
 The simplest version of algebraic assignment problem is:
%
 \[ \max  \sum_{i=1}^{m} \sum_{j=1}^{n} c^{1}_{i,j} x_{i,j}
 ~~s.t. \sum_{i=1}^{m} x_{i,j} \leq 1, j=\overline{1,n};
 ~ \sum_{j=1}^{n} x_{i,j} \leq 1, i=\overline{1,m};
 ~ x_{i,j} \in \{0,1\}. \]
%
 This problem is polynomially solvable.
 Let us consider \(n=m\).
 In this case, a solution can be considered
 as a permutation of elements
 ~\(A=\{1,...,i,..,n\}\):  \( S = < s[1],...,s[i],..,s[n] > \),
 where \(s[i]\) defines the position of element
 \(i\) in the resultant permutation \(S\).
 Let ~\(c(i,s[i]) \geq 0\)~ (\(i=\overline{1,n}\))~
 be a ``profit'' of assignment of element \(i\) into
 position \(s[i]\)
 (i.e., \(\| c(i,s[i]) \|\) is a ``profit'' matrix).

 The combinatorial formulation of assignment problem  is:

~~

 Find permutation \(S\) such that
 ~~\( \sum_{i=1}^{n} c(i,s[i]) \rightarrow  \max \).

~~

 Now let us consider three solutions (permutations):

 (a) ~\(S^{1} = < s^{1}[1],...,s^{1}[i],..,s^{1}[n] > \)~
 for \(t=\tau_{1}\),

 (b)
 ~\(S^{2} = < s^{2}[1],...,s^{2}[i],..,s^{2}[n] >\)~
 for \(t=\tau_{2}\),
 and

 (c)
  ~\(S^{*}= < s^{*}[1],...,s^{*}[i],..,s^{*}[n] >\)~  (the restructured
  solution).

~~

 {\bf Illustrative numerical example:}~
 \(A = \{1,2,3,4,5,6,7\}\),

 \(S^{1} = \{2,4,5,1,3,7,6\}\),
 ~\(S^{2} = \{4,1,3,7,5,2,6\}\),
 ~\(S^{*} = \{2,4,3,1,5,7,6\}\).

 Here the following changes are made in \(S^{1}\):
 ~\(5 \rightarrow 3 \), \(3 \rightarrow 5 \).
 Clearly, the changes can be based on typical exchange operations:
 {\it 2-exchange}, {\it 3-exchange}, etc.

~~

 Further, let us consider a vector of structural difference
 (by components) for two permutations \(S^{\alpha}\) and \(S^{\beta}\):
  ~\(\{ s^{\alpha}[i] - s^{\beta}[i], i=\overline{1,n} \} \)
  and
  a change cost matrix
  ~\(\| d(i,j) \|\)~ (\(i=\overline{1,n}, j=\overline{1,n}\)).
 Here ~\( d(i,i) =0\) ~ \(\forall i=\overline{1,n}\).
 Evidently, the cost for restructuring
 solution \(S^{1}\) into solution  \(S^{*}\) is:
 ~\( H (S^{1} \rightarrow S^{*}) =  \sum_{i=1}^{n} h(s^{1}[i],s^{*}[i]) \).
 Proximity (by ``profit'') for two permutations \(S^{\alpha}\) and \(S^{\beta}\)
 may be considered as follows:
  \(\rho ( S^{\alpha},S^{\beta}) =
  | \sum_{i=1}^{n} c^{\alpha}(i,s^{\alpha}[i]) - \sum_{i=1}^{n} c^{\beta}(i,s^{\beta}[i])|
  \).
 Finally, the restructuring of assignment is (a simple version):
 \[\min \rho ( S^{*},S^{2} )
 ~~~~s.t.~~ H (S^{1} \rightarrow S^{*}) =  \sum_{i=1}^{n} h(s^{1}[i],s^{*}[i]) \leq \widehat{h}.\]

 In the case of multicriteria assignment problem,
 the restructuring process is the same.
 Thus,
 the presented restructuring of assignment
 can be examined as well
 (multicriteria case).

~~

 {\bf Example of reassignment of users to access points}
 \cite{lev10a,lev11restr,levpet10}.
 Here the initial multicriteria assignment problem involves
 21 users and 6 access points.
 Table 5 and Table 6 contain
 some parameters for users (\(A\))
 (coordinates (\(x_{i},y_{i},z_{i}\)),
  required frequency spectrum \(f_{j}\),
  required level of reliability \(r_{j}\), etc.)
 and
 some parameters for 6 access points (\(B = \{j\} =\{1,2,3,4,5,6\}\))
 (coordinates (\(x_{j},y_{j},z_{j}\)),
  frequency spectrum \(f_{j}\),
  number of connections \(n_{j}\),
  level of reliability \(r_{j}\))
  (\cite{lev10a}, \cite{levpet10}).

\begin{center}
 {\bf Table 5.} Access points\\
\begin{tabular}{| c| c | c|c| c | c| c|}
\hline
 \(j\)&\(x_{j}\)&\(y_{j}\)&\(z_{j}\)&\(f_{j}\)&
  \(n_{j}\)&\(r_{j}\)\\

\hline

 \(1\)&\(50\) &\(157\)&\(10\)&\(30\)&\(4\)&\(10\)\\
 \(2\)&\(72\) &\(102\)&\(10\)&\(42\)&\(6\)&\(10\)\\
 \(3\)&\(45\) &\(52\) &\(10\)&\(45\)&\(10\)&\(10\)\\
 \(4\)&\(150\)&\(165\)&\(10\)&\(30\)&\(5\)&\(15\)\\
 \(5\)&\(140\)&\(112\)&\(10\)&\(32\)&\(5\)&\(8\)\\
 \(6\)&\(147\) &\(47\)&\(10\)&\(30\)&\(5\)&\(15\)\\

\hline
\end{tabular}
\end{center}

\begin{center}
 {\bf Table 6.} Users\\
\begin{tabular}{| c| c | c|c| c | c|}
\hline
 \(i\)&\(x_{i}\)&\(y_{i}\)&\(z_{i}\)&\(f_{j}\)&\(r_{j}\)\\

\hline

 \(1\) &\(30\)&\(165\)&\(5\)&\(10\)&\(5\)\\
 \(2\) &\(58\)&\(174\)&\(5\)&\(5\)&\(9\)\\
 \(3\) &\(95\)&\(156\)&\(0\)&\(6\)&\(6\)\\
 \(4\) &\(52\)&\(134\)&\(5\)&\(6\)&\(8\)\\
 \(5\) &\(85\)&\(134\)&\(3\)&\(6\)&\(7\)\\
 \(6\) &\(27\)&\(109\)&\(7\)&\(8\)&\(5\)\\

 \(7\) &\(55\) &\(105\)&\(2\)&\(7\)&\(10\)\\
 \(8\) &\(98\) &\(89\) &\(3\)&\(10\)&\(10\)\\
 \(9\) &\(25\) &\(65\) &\(2\)&\(7\)&\(5\)\\
 \(10\)&\(52\)&\(81\)&\(1\)&\(10\)&\(8\)\\
 \(11\)&\(65\)&\(25\)&\(7\)&\(6\)&\(9\)\\
 \(12\)&\(93\)&\(39\)&\(1\)&\(10\)&\(10\)\\
 \(13\)&\(172\) &\(26\)&\(2\)&\(10\)&\(7\)\\
 \(14\)&\(110\) &\(169\)&\(5\)&\(7\)&\(5\)\\
 \(15\)&\(145\) &\(181\) &\(3\)&\(5\)&\(4\)\\
 \(16\)&\(150\)&\(150\)&\(5\)&\(7\)&\(4\)\\
 \(17\)&\(120\)&\(140\)&\(6\)&\(4\)&\(6\)\\
 \(18\)&\(150\)&\(136\)&\(3\)&\(6\)&\(7\)\\
 \(19\)&\(135\) &\(59\)&\(4\)&\(13\)&\(4\)\\
 \(20\)&\(147\) &\(79\)&\(5\)&\(7\)&\(16\)\\
 \(21\)&\(127\) &\(95\) &\(5\)&\(7\)&\(5\)\\

\hline
\end{tabular}
\end{center}

 A simplified version of assignment problem from \cite{lev10a} is considered.
 Two regions are examined:
 an initial region and an additional region (Fig. 21).
 In \cite{lev10a} the problem was solved for two cases:
 (i) separated assignment \(S^{1}\) (Fig. 21),
 (ii) joint assignment \(S^{2}\) (Fig. 22).

\begin{center}
\begin{picture}(55,81)

\put(03,00){\makebox(0,0)[bl]{Fig. 21.
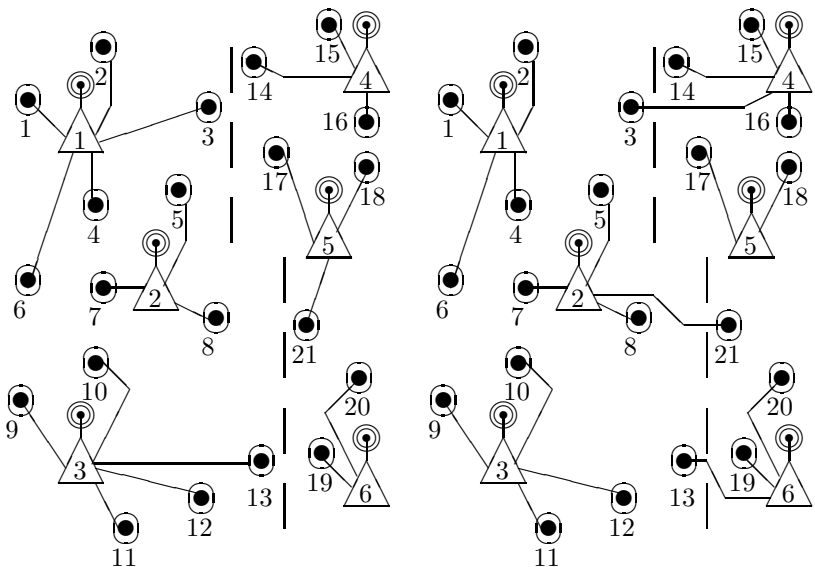
 Separated assignment \(S^{1}\)}}


\put(14,32){\oval(3.2,4)} \put(14,32){\circle*{2}}
\put(12,27){\makebox(0,0)[bl]{\(10\)}}

\put(15,42){\oval(3.2,4)} \put(15,42){\circle*{2}}
\put(13,37){\makebox(0,0)[bl]{\(7\)}}

\put(19,39){\line(1,0){6}} \put(19,39){\line(1,2){3}}
\put(25,39){\line(-1,2){3}} \put(22,45){\line(0,1){3}}
\put(22,48){\circle*{1}} \put(21,39.5){\makebox(0,0)[bl]{\(2\)}}

\put(22,48){\circle{2.5}} \put(22,48){\circle{3.5}}
\put(24.5,40){\line(2,-1){6}}
\put(23,42.3){\line(1,2){3}} \put(26,48.3){\line(0,1){4.8}}
\put(20.5,42){\line(-1,0){4.5}}

\put(05,43){\oval(3.2,4)} \put(05,43){\circle*{2}}
\put(03,38){\makebox(0,0)[bl]{\(6\)}}

\put(11,60){\line(-1,-3){6}}

\put(05,67){\oval(3.2,4)} \put(05,67){\circle*{2}}
\put(03.8,62){\makebox(0,0)[bl]{\(1\)}}

\put(15,74){\oval(3.2,4)} \put(15,74){\circle*{2}}
\put(13.8,69){\makebox(0,0)[bl]{\(2\)}}

\put(09,60){\line(1,0){6}} \put(09,60){\line(1,2){3}}
\put(15,60){\line(-1,2){3}} \put(12,66){\line(0,1){3}}
\put(12,69){\circle*{1}} \put(11,60.5){\makebox(0,0)[bl]{\(1\)}}

\put(12,69){\circle{2.5}} \put(12,69){\circle{3.5}}
\put(10,62){\line(-1,1){5}}
\put(13.5,60){\line(0,-1){6}}
\put(14,62.3){\line(1,2){2}} \put(16,66.3){\line(0,1){5.8}}

\put(14,53){\oval(3.2,4)} \put(14,53){\circle*{2}}
\put(12.8,48){\makebox(0,0)[bl]{\(4\)}}

\put(04,27){\oval(3.2,4)} \put(04,27){\circle*{2}}
\put(02,22){\makebox(0,0)[bl]{\(9\)}}

\put(18,10){\oval(3.2,4)} \put(18,10){\circle*{2}}
\put(16,5){\makebox(0,0)[bl]{\(11\)}}

\put(28,14){\oval(3.2,4)} \put(28,14){\circle*{2}}
\put(26,9){\makebox(0,0)[bl]{\(12\)}}

\put(30,38){\oval(3.2,4)} \put(30,38){\circle*{2}}
\put(28,33){\makebox(0,0)[bl]{\(8\)}}

\put(38,60){\oval(3.2,4)} \put(38,60){\circle*{2}}
\put(36,55){\makebox(0,0)[bl]{\(17\)}}

\put(50,58){\oval(3.2,4)} \put(50,58){\circle*{2}}
\put(49,53){\makebox(0,0)[bl]{\(18\)}}

\put(42,46){\line(1,0){6}} \put(42,46){\line(1,2){3}}
\put(48,46){\line(-1,2){3}} \put(45,52){\line(0,1){3}}
\put(45,55){\circle*{1}} \put(44,46.5){\makebox(0,0)[bl]{\(5\)}}

\put(45,55){\circle{2.5}} \put(45,55){\circle{3.5}}
\put(43,48){\line(-1,3){4}}

\put(46,49.4){\line(1,2){4.5}}

\put(49,30){\oval(3.2,4)} \put(49,30){\circle*{2}}
\put(47,25){\makebox(0,0)[bl]{\(20\)}}

\put(44,20){\oval(3.2,4)} \put(44,20){\circle*{2}}
\put(42,15){\makebox(0,0)[bl]{\(19\)}}

\put(42,37){\oval(3.2,4)} \put(42,37){\circle*{2}}
\put(40,32){\makebox(0,0)[bl]{\(21\)}}

\put(42,37){\line(1,3){3}}

\put(47,13){\line(1,0){6}} \put(47,13){\line(1,2){3}}
\put(53,13){\line(-1,2){3}} \put(50,19){\line(0,1){3}}
\put(50,22){\circle*{1}} \put(49,13.5){\makebox(0,0)[bl]{\(6\)}}

\put(50,22){\circle{2.5}} \put(50,22){\circle{3.5}}
\put(48,15.5){\line(-1,1){4.5}}
\put(48.7,17){\line(-1,2){4.2}} \put(44.5,25.4){\line(1,1){4}}

\put(25,55){\oval(3.2,4)} \put(25,55){\circle*{2}}
\put(24,50){\makebox(0,0)[bl]{\(5\)}}
\put(29,66){\oval(3.2,4)} \put(29,66){\circle*{2}}
\put(28,61){\makebox(0,0)[bl]{\(3\)}}
\put(28.5,66){\line(-3,-1){14}}

\put(35,72){\oval(3.2,4)} \put(35,72){\circle*{2}}
\put(34,67){\makebox(0,0)[bl]{\(14\)}}

\put(45,77){\oval(3.2,4)} \put(45,77){\circle*{2}}
\put(43,72){\makebox(0,0)[bl]{\(15\)}}

\put(50,64){\oval(3.2,4)} \put(50,64){\circle*{2}}
\put(44,63){\makebox(0,0)[bl]{\(16\)}}

\put(50,64){\line(0,1){4}}

\put(47,68){\line(1,0){6}} \put(47,68){\line(1,2){3}}
\put(53,68){\line(-1,2){3}} \put(50,74){\line(0,1){3}}
\put(50,77){\circle*{1}} \put(49,68.5){\makebox(0,0)[bl]{\(4\)}}

\put(50,77){\circle{2.5}} \put(50,77){\circle{3.5}}
\put(48,70){\line(-1,0){09}} \put(39,70){\line(-2,1){5}}
\put(48.4,71.2){\line(-1,2){3}}

\put(09,16){\line(1,0){6}} \put(09,16){\line(1,2){3}}
\put(15,16){\line(-1,2){3}} \put(12,22){\line(0,1){3}}
\put(12,25){\circle*{1}} \put(11,16.5){\makebox(0,0)[bl]{\(3\)}}

\put(12,25){\circle{2.5}} \put(12,25){\circle{3.5}}
\put(10,18){\line(-2,3){6.5}}
\put(13.5,18.5){\line(1,2){05}} \put(18.5,28.5){\line(-1,1){04}}
\put(14,18){\line(4,-1){14}}
\put(14,16){\line(1,-2){3}}


\put(36,19){\oval(3.2,4)} \put(36,19){\circle*{2}}
\put(34,13){\makebox(0,0)[bl]{\(13\)}}
\put(34.4,18.6){\line(-1,0){20.6}}

\put(39,10){\line(0,1){6}} \put(39.1,10){\line(0,1){6}}
\put(39,20){\line(0,1){6}} \put(39.1,20){\line(0,1){6}}
\put(39,30){\line(0,1){6}} \put(39.1,30){\line(0,1){6}}
\put(39,40){\line(0,1){6}} \put(39.1,40){\line(0,1){6}}
\put(32,48){\line(0,1){6}} \put(32.1,48){\line(0,1){6}}
\put(32,58){\line(0,1){6}} \put(32.1,58){\line(0,1){6}}
\put(32,68){\line(0,1){6}} \put(32.1,68){\line(0,1){6}}

\end{picture}
%
\begin{picture}(55,80)

\put(05,00){\makebox(0,0)[bl]{Fig. 22.
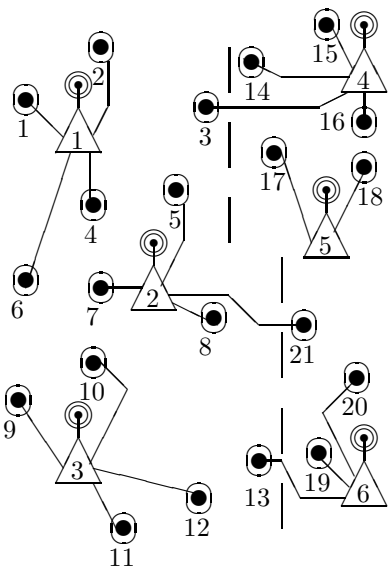
 Joint assignment \(S^{2}\)}}


\put(14,32){\oval(3.2,4)} \put(14,32){\circle*{2}}
\put(12,27){\makebox(0,0)[bl]{\(10\)}}

\put(15,42){\oval(3.2,4)} \put(15,42){\circle*{2}}
\put(13,37){\makebox(0,0)[bl]{\(7\)}}

\put(19,39){\line(1,0){6}} \put(19,39){\line(1,2){3}}
\put(25,39){\line(-1,2){3}} \put(22,45){\line(0,1){3}}
\put(22,48){\circle*{1}} \put(21,39.5){\makebox(0,0)[bl]{\(2\)}}

\put(22,48){\circle{2.5}} \put(22,48){\circle{3.5}}
\put(24.5,40){\line(2,-1){6}}
\put(23,42.3){\line(1,2){3}} \put(26,48.3){\line(0,1){4.8}}
\put(20.5,42){\line(-1,0){4.5}}

\put(05,43){\oval(3.2,4)} \put(05,43){\circle*{2}}
\put(03,38){\makebox(0,0)[bl]{\(6\)}}

\put(11,60){\line(-1,-3){6}}

\put(05,67){\oval(3.2,4)} \put(05,67){\circle*{2}}
\put(03.8,62){\makebox(0,0)[bl]{\(1\)}}

\put(15,74){\oval(3.2,4)} \put(15,74){\circle*{2}}
\put(13.8,69){\makebox(0,0)[bl]{\(2\)}}

\put(09,60){\line(1,0){6}} \put(09,60){\line(1,2){3}}
\put(15,60){\line(-1,2){3}} \put(12,66){\line(0,1){3}}
\put(12,69){\circle*{1}} \put(11,60.5){\makebox(0,0)[bl]{\(1\)}}

\put(12,69){\circle{2.5}} \put(12,69){\circle{3.5}}
\put(10,62){\line(-1,1){5}}
\put(13.5,60){\line(0,-1){6}}
\put(14,62.3){\line(1,2){2}} \put(16,66.3){\line(0,1){5.8}}

\put(14,53){\oval(3.2,4)} \put(14,53){\circle*{2}}
\put(12.8,48){\makebox(0,0)[bl]{\(4\)}}

\put(04,27){\oval(3.2,4)} \put(04,27){\circle*{2}}
\put(02,22){\makebox(0,0)[bl]{\(9\)}}

\put(18,10){\oval(3.2,4)} \put(18,10){\circle*{2}}
\put(16,5){\makebox(0,0)[bl]{\(11\)}}

\put(28,14){\oval(3.2,4)} \put(28,14){\circle*{2}}
\put(26,9){\makebox(0,0)[bl]{\(12\)}}

\put(30,38){\oval(3.2,4)} \put(30,38){\circle*{2}}
\put(28,33){\makebox(0,0)[bl]{\(8\)}}

\put(38,60){\oval(3.2,4)} \put(38,60){\circle*{2}}
\put(36,55){\makebox(0,0)[bl]{\(17\)}}

\put(50,58){\oval(3.2,4)} \put(50,58){\circle*{2}}
\put(49,53){\makebox(0,0)[bl]{\(18\)}}

\put(42,46){\line(1,0){6}} \put(42,46){\line(1,2){3}}
\put(48,46){\line(-1,2){3}} \put(45,52){\line(0,1){3}}
\put(45,55){\circle*{1}} \put(44,46.5){\makebox(0,0)[bl]{\(5\)}}

\put(45,55){\circle{2.5}} \put(45,55){\circle{3.5}}

\put(43,48){\line(-1,3){4}}

\put(46,49.4){\line(1,2){4.5}}



\put(49,30){\oval(3.2,4)} \put(49,30){\circle*{2}}
\put(47,25){\makebox(0,0)[bl]{\(20\)}}

\put(44,20){\oval(3.2,4)} \put(44,20){\circle*{2}}
\put(42,15){\makebox(0,0)[bl]{\(19\)}}

\put(42,37){\oval(3.2,4)} \put(42,37){\circle*{2}}
\put(40,32){\makebox(0,0)[bl]{\(21\)}}


\put(42,37){\line(-1,0){6}} \put(36,37){\line(-1,1){4}}
\put(32,41){\line(-1,0){8}}

\put(47,13){\line(1,0){6}} \put(47,13){\line(1,2){3}}
\put(53,13){\line(-1,2){3}} \put(50,19){\line(0,1){3}}
\put(50,22){\circle*{1}} \put(49,13.5){\makebox(0,0)[bl]{\(6\)}}

\put(50,22){\circle{2.5}} \put(50,22){\circle{3.5}}
\put(48,15.5){\line(-1,1){4.5}}
\put(48.7,17){\line(-1,2){4.2}} \put(44.5,25.4){\line(1,1){4}}

\put(25,55){\oval(3.2,4)} \put(25,55){\circle*{2}}
\put(24,50){\makebox(0,0)[bl]{\(5\)}}
\put(29,66){\oval(3.2,4)} \put(29,66){\circle*{2}}
\put(28,61){\makebox(0,0)[bl]{\(3\)}}
\put(29,66){\line(1,0){15}} \put(44,66){\line(2,1){4}}

\put(35,72){\oval(3.2,4)} \put(35,72){\circle*{2}}
\put(34,67){\makebox(0,0)[bl]{\(14\)}}

\put(45,77){\oval(3.2,4)} \put(45,77){\circle*{2}}
\put(43,72){\makebox(0,0)[bl]{\(15\)}}

\put(50,64){\oval(3.2,4)} \put(50,64){\circle*{2}}
\put(44,63){\makebox(0,0)[bl]{\(16\)}}

\put(50,64){\line(0,1){4}}

\put(47,68){\line(1,0){6}} \put(47,68){\line(1,2){3}}
\put(53,68){\line(-1,2){3}} \put(50,74){\line(0,1){3}}
\put(50,77){\circle*{1}} \put(49,68.5){\makebox(0,0)[bl]{\(4\)}}

\put(50,77){\circle{2.5}} \put(50,77){\circle{3.5}}
\put(48,70){\line(-1,0){09}} \put(39,70){\line(-2,1){5}}
\put(48.4,71.2){\line(-1,2){3}}

\put(09,16){\line(1,0){6}} \put(09,16){\line(1,2){3}}
\put(15,16){\line(-1,2){3}} \put(12,22){\line(0,1){3}}
\put(12,25){\circle*{1}} \put(11,16.5){\makebox(0,0)[bl]{\(3\)}}

\put(12,25){\circle{2.5}} \put(12,25){\circle{3.5}}
\put(10,18){\line(-2,3){6.5}}
\put(13.5,18.5){\line(1,2){05}} \put(18.5,28.5){\line(-1,1){04}}
\put(14,18){\line(4,-1){14}}
\put(14,16){\line(1,-2){3}}


\put(36,19){\oval(3.2,4)} \put(36,19){\circle*{2}}
\put(34,13){\makebox(0,0)[bl]{\(13\)}}

\put(36,19){\line(1,0){3}} \put(39,19){\line(1,-2){2.5}}
\put(41.5,14){\line(1,0){6}}


\put(39,10){\line(0,1){6}} \put(39.1,10){\line(0,1){6}}
\put(39,20){\line(0,1){6}} \put(39.1,20){\line(0,1){6}}
\put(39,30){\line(0,1){6}} \put(39.1,30){\line(0,1){6}}
\put(39,40){\line(0,1){6}} \put(39.1,40){\line(0,1){6}}
\put(32,48){\line(0,1){6}} \put(32.1,48){\line(0,1){6}}
\put(32,58){\line(0,1){6}} \put(32.1,58){\line(0,1){6}}
\put(32,68){\line(0,1){6}} \put(32.1,68){\line(0,1){6}}

\end{picture}
\end{center}

 The restructured problem is considered
 as a modification (change) of \(S^{1}\) into \(S^{*}\).
%
 To reduce the problem
 it is reasonable
 the select a subset of users
 (a ``change zone'' near borders between regions):
 ~\(\widetilde{A} = \{i\} =\{3,5,8,12,13,14,17,19,21\}\).
 Thus,
 it is necessary to assign each element of
 \(\widetilde{A}\) into an access point of ~\(B\).

 The considered simplified restructuring problem
 is based on set of change operations:
 (1) user 3, change of connection: ~\(1\rightarrow 4\)~
 (Boolean variable \(x_{1}\)),
 (2) user 13, change of connection: ~\(3\rightarrow 6\)~
 (Boolean variable \(x_{2}\)),
 (3) user 21, change of connection: ~\(5\rightarrow 2\)~
  (Boolean variable \(x_{3}\)).
 Table 7 contains estimates of change costs (expert judgment)
 and ``integrated profits'' of correspondence between users and access
 points from  (\cite{lev10a,levpet10}).

 The problem is:
 \[ \max ~(~ c_{3,4}~x_{1}+c_{13,6}~x_{2}+c_{21,2}~x_{3}~)\]
 \[s.t.~~ (~
 (h^{-}_{3,1}+h^{+}_{3,4}) ~x_{1} +
 (h^{-}_{13,3}+h^{+}_{13,6}) ~x_{2} +
 (h^{-}_{21,51}+h^{+}_{21,2}) ~x_{3}~)
 \leq \widehat{h}. \]
 The reassignment
 ~\( S^{*} \)~ is depicted in Fig. 23
 (i.e., \(x_{1}=0, x_{1}=1, x_{3}=1\), \(\widehat{h}=5\)).

\begin{center}
 {\bf Table 7.} User \(i\) - access points \(j\): \(h^{-}_{ij}\), \(h^{+}_{ij}\), \(c_{ij}\),\\
\begin{tabular}{| c| l  c c c c c c|}
\hline
 \(i\)&Access point \(j\): &\(1\)&\(2\)&\(3\)&\(4\)&\(5\)&\(6\)\\

\hline

 \(3\)& &\(3,2,2\)&\(2,1,3\)&\(1,0,3\)&\(3,1,3\)&\(2,1,0\)&\(1,1,0\)\\
 \(5\)& &\(2,1,1\)&\(1,3,1\)&\(1,2,1\)&\(3,2,1\)&\(1,1,1\)&\(1,1,1\)\\
 \(8\)& &\(1,1,3\)&\(1,1,3\)&\(1,1,3\)&\(1,1,0\)&\(1,1,3\)&\(2,2,2\)\\
 \(12\)& &\(2,2,3\)&\(1,2,3\)&\(1,2,3\)&\(3,1,0\)&\(2,1,0\)&\(1,1,0\)\\
 \(13\)& &\(1,1,3\)&\(1,1,3\)&\(1,1,3\)&\(2,1,0\)&\(2,2,1\)&\(1,1,3\)\\
 \(14\)& &\(1,1,1\)&\(2,2,2\)&\(1,2,0\)&\(1,1,1\)&\(1,1,1\)&\(1,1,0\)\\
 \(17\)& &\(1,1,2\)&\(1,1,1\)&\(1,0,1\)&\(3,1,1\)&\(1,1,1\)&\(1,1,1\)\\
 \(19\)& &\(1,1,0\)&\(1,1,3\)&\(1,2,3\)&\(3,2,0\)&\(1,1,3\)&\(1,1,2\)\\
 \(21\)& &\(1,1,0\)&\(1,2,3\)&\(1,1,2\)&\(3,1,1\)&\(1,1,1\)&\(1,1,1\)\\

\hline
\end{tabular}
\end{center}

\begin{center}
\begin{picture}(55,81)
\put(05,00){\makebox(0,0)[bl]{Fig. 23.
 Joint assignment \(S^{*}\)}}


\put(14,32){\oval(3.2,4)} \put(14,32){\circle*{2}}
\put(12,27){\makebox(0,0)[bl]{\(10\)}}

\put(15,42){\oval(3.2,4)} \put(15,42){\circle*{2}}
\put(13,37){\makebox(0,0)[bl]{\(7\)}}

\put(19,39){\line(1,0){6}} \put(19,39){\line(1,2){3}}
\put(25,39){\line(-1,2){3}} \put(22,45){\line(0,1){3}}
\put(22,48){\circle*{1}} \put(21,39.5){\makebox(0,0)[bl]{\(2\)}}

\put(22,48){\circle{2.5}} \put(22,48){\circle{3.5}}
\put(24.5,40){\line(2,-1){6}}
\put(23,42.3){\line(1,2){3}} \put(26,48.3){\line(0,1){4.8}}
\put(20.5,42){\line(-1,0){4.5}}

\put(05,43){\oval(3.2,4)} \put(05,43){\circle*{2}}
\put(03,38){\makebox(0,0)[bl]{\(6\)}}

\put(11,60){\line(-1,-3){6}}

\put(05,67){\oval(3.2,4)} \put(05,67){\circle*{2}}
\put(03.8,62){\makebox(0,0)[bl]{\(1\)}}

\put(15,74){\oval(3.2,4)} \put(15,74){\circle*{2}}
\put(13.8,69){\makebox(0,0)[bl]{\(2\)}}

\put(09,60){\line(1,0){6}} \put(09,60){\line(1,2){3}}
\put(15,60){\line(-1,2){3}} \put(12,66){\line(0,1){3}}
\put(12,69){\circle*{1}} \put(11,60.5){\makebox(0,0)[bl]{\(1\)}}

\put(12,69){\circle{2.5}} \put(12,69){\circle{3.5}}
\put(10,62){\line(-1,1){5}}
\put(13.5,60){\line(0,-1){6}}
\put(14,62.3){\line(1,2){2}} \put(16,66.3){\line(0,1){5.8}}

\put(14,53){\oval(3.2,4)} \put(14,53){\circle*{2}}
\put(12.8,48){\makebox(0,0)[bl]{\(4\)}}

\put(04,27){\oval(3.2,4)} \put(04,27){\circle*{2}}
\put(02,22){\makebox(0,0)[bl]{\(9\)}}

\put(18,10){\oval(3.2,4)} \put(18,10){\circle*{2}}
\put(16,5){\makebox(0,0)[bl]{\(11\)}}

\put(28,14){\oval(3.2,4)} \put(28,14){\circle*{2}}
\put(26,9){\makebox(0,0)[bl]{\(12\)}}

\put(30,38){\oval(3.2,4)} \put(30,38){\circle*{2}}
\put(28,33){\makebox(0,0)[bl]{\(8\)}}

\put(38,60){\oval(3.2,4)} \put(38,60){\circle*{2}}
\put(36,55){\makebox(0,0)[bl]{\(17\)}}

\put(50,58){\oval(3.2,4)} \put(50,58){\circle*{2}}
\put(49,53){\makebox(0,0)[bl]{\(18\)}}

\put(28.5,66){\line(-3,-1){14}}

\put(42,46){\line(1,0){6}} \put(42,46){\line(1,2){3}}
\put(48,46){\line(-1,2){3}} \put(45,52){\line(0,1){3}}
\put(45,55){\circle*{1}} \put(44,46.5){\makebox(0,0)[bl]{\(5\)}}

\put(45,55){\circle{2.5}} \put(45,55){\circle{3.5}}

\put(43,48){\line(-1,3){4}}

\put(46,49.4){\line(1,2){4.5}}


\put(49,30){\oval(3.2,4)} \put(49,30){\circle*{2}}
\put(47,25){\makebox(0,0)[bl]{\(20\)}}

\put(44,20){\oval(3.2,4)} \put(44,20){\circle*{2}}
\put(42,15){\makebox(0,0)[bl]{\(19\)}}

\put(42,37){\oval(3.2,4)} \put(42,37){\circle*{2}}
\put(40,32){\makebox(0,0)[bl]{\(21\)}}


\put(42,37){\line(-1,0){6}} \put(36,37){\line(-1,1){4}}
\put(32,41){\line(-1,0){8}}

\put(47,13){\line(1,0){6}} \put(47,13){\line(1,2){3}}
\put(53,13){\line(-1,2){3}} \put(50,19){\line(0,1){3}}
\put(50,22){\circle*{1}} \put(49,13.5){\makebox(0,0)[bl]{\(6\)}}

\put(50,22){\circle{2.5}} \put(50,22){\circle{3.5}}
\put(48,15.5){\line(-1,1){4.5}}
\put(48.7,17){\line(-1,2){4.2}} \put(44.5,25.4){\line(1,1){4}}

\put(25,55){\oval(3.2,4)} \put(25,55){\circle*{2}}
\put(24,50){\makebox(0,0)[bl]{\(5\)}}
\put(29,66){\oval(3.2,4)} \put(29,66){\circle*{2}}
\put(28,61){\makebox(0,0)[bl]{\(3\)}}

\put(35,72){\oval(3.2,4)} \put(35,72){\circle*{2}}
\put(34,67){\makebox(0,0)[bl]{\(14\)}}

\put(45,77){\oval(3.2,4)} \put(45,77){\circle*{2}}
\put(43,72){\makebox(0,0)[bl]{\(15\)}}

\put(50,64){\oval(3.2,4)} \put(50,64){\circle*{2}}
\put(44,63){\makebox(0,0)[bl]{\(16\)}}

\put(50,64){\line(0,1){4}}

\put(47,68){\line(1,0){6}} \put(47,68){\line(1,2){3}}
\put(53,68){\line(-1,2){3}} \put(50,74){\line(0,1){3}}
\put(50,77){\circle*{1}} \put(49,68.5){\makebox(0,0)[bl]{\(4\)}}

\put(50,77){\circle{2.5}} \put(50,77){\circle{3.5}}
\put(48,70){\line(-1,0){09}} \put(39,70){\line(-2,1){5}}
\put(48.4,71.2){\line(-1,2){3}}

\put(09,16){\line(1,0){6}} \put(09,16){\line(1,2){3}}
\put(15,16){\line(-1,2){3}} \put(12,22){\line(0,1){3}}
\put(12,25){\circle*{1}} \put(11,16.5){\makebox(0,0)[bl]{\(3\)}}

\put(12,25){\circle{2.5}} \put(12,25){\circle{3.5}}
\put(10,18){\line(-2,3){6.5}}
\put(13.5,18.5){\line(1,2){05}} \put(18.5,28.5){\line(-1,1){04}}
\put(14,18){\line(4,-1){14}}
\put(14,16){\line(1,-2){3}}


\put(36,19){\oval(3.2,4)} \put(36,19){\circle*{2}}
\put(34,13){\makebox(0,0)[bl]{\(13\)}}

\put(36,19){\line(1,0){3}} \put(39,19){\line(1,-2){2.5}}
\put(41.5,14){\line(1,0){6}}


\put(39,10){\line(0,1){6}} \put(39.1,10){\line(0,1){6}}
\put(39,20){\line(0,1){6}} \put(39.1,20){\line(0,1){6}}
\put(39,30){\line(0,1){6}} \put(39.1,30){\line(0,1){6}}
\put(39,40){\line(0,1){6}} \put(39.1,40){\line(0,1){6}}
\put(32,48){\line(0,1){6}} \put(32.1,48){\line(0,1){6}}
\put(32,58){\line(0,1){6}} \put(32.1,58){\line(0,1){6}}
\put(32,68){\line(0,1){6}} \put(32.1,68){\line(0,1){6}}

\end{picture}
\end{center}
%

\subsection{Morphological clique problem}

 Morphological clique problem is a basis of
 Hierarchical Morphological Multicriteria Design (HMMD)
 (combinatorial synthesis)
 (e.g.,
 \cite{lev98,lev06,lev15}).
 A brief description of HMMD is the following.
 An examined modular system consists
 of components and their compatibility (IC).
 Basic assumptions are:
 ~(a) a tree-like structure of the system;
 ~(b) a composite estimate for system quality
     that integrates components (subsystems, parts) qualities and
    qualities of IC (compatibility) across subsystems;
 ~(c) monotonic criteria for the system and its components;
 ~(d) quality estimates of system components and IC are evaluated by
 coordinated ordinal scales.
 The designations are:
  ~(1) design alternatives (DAs) for
  nodes of the model (i.e., components);
  ~(2) priorities of DAs (\(r=\overline{1,k}\);
      \(1\) corresponds to the best level of quality);
  ~(3) an ordinal compatibility estimate for each pair of DAs
  (\(w=\overline{0,l}\); \(l\) corresponds to the best level of quality).
 The phases of HMMD are:
  ~{\it 1.} design of the tree-like system model;
  ~{\it 2.} generation of DAs for each nodes (i.e., system component);
  ~{\it 3.} hierarchical selection and composition of DAs into composite
    DAs for the corresponding higher level of the system
    hierarchy.
 Let \(S\) be a system consisting of \(m\)  components:
 \(P(1),...,P(i),...,P(m)\).
 The problem is:

~~

 {\it Find composite design alternative}
 ~ \(S=S(1)\star ...\star S(i)\star ...\star S(m)\)~
 ({\it one representative design alternative} \(S(i)\)
 {\it for each system component/part} ~\(P(i)\), \(i=\overline{1,m}\))
 {\it with non-zero}~ IC
 {\it estimates between the representative design alternatives.}

~~

 A discrete ``space'' of the integrated system excellence is based
 on the following vector:
 ~\(N(S)=(w(S);n(S))\),
 ~where \(w(S)\) is the minimum of pairwise compatibility
 between DAs which correspond to different system components
 (i.e.,
 \(~\forall ~P_{j_{1}}\) and \( P_{j_{2}}\),
 \(1 \leq j_{1} \neq j_{2} \leq m\))
 in \(S\),
 ~\(n(S)=(n_{1},...,n_{r},...n_{k})\),
 ~where \(n_{r}\) is the number of DAs of the \(r\)th quality in \(S\)
 ~(\(\sum^{k}_{r=1} n_{r} = m\))
 (Fig. 11).
 Thus, synthesis problem is:
 \[\max n(S), ~ \max w(S)  ~~~ s.t. ~~ w(S)\geq 1 ~~~~~~~~
 or ~~~~~~~~
 \max N(S) ~~~ s.t. ~~ w(S)\geq 1.\]
 As a result,
 composite solutions which are nondominated by \(N(S)\)
 (i.e., Pareto-efficient solutions) are searched for.

 In the simplified numerical example
 (synthesis of four-component team for a start-up company
  \cite{lev15route}),
 ordinal scale \([1,2,3]\) is used for quality of DAs
 and ordinal scale \([0,1,2,3]\) is used for compatibility estimates.
%
%
 The basic simplified hierarchical structure
 of the considered team:

~~

 {\bf 1.} Team  ~\(T = L \star R \star I  \star K\):

 {\it 1.1.} Project leader \(L\):
 basic leader \(L_{1}\),
 the 2nd leader \(L_{2}\),
 extended group of leaders   \(L_{3}\);

 {\it 1.2.} Researcher \(R\):
 basic researcher (models, algorithms)
  \(R_{1}\),
 the 2nd researcher (models, algorithms)
  \(R_{2}\),
 the 3rd  researcher (models, algorithms)
  \(R_{3}\),
 a group of researchers  (models, algorithms)
  \(R_{4} = R_{1} \& R_{2} \),
 extended group of researchers
(including
 applications in R\&D and engineering,
  educational technology)
  \(R_{5} =  R_{1} \& R_{2} \& R_{3}\);

 {\it 1.3.} Engineer-programmer \(E\):
 none \(E_{1}\),
 engineer \(E_{2}\),
 group of engineers   \(E_{3}\),
 extended group of engineers
 (including specialist in Web-design)
  \(E_{4}\);

 {\it 1.4.} Specialist in marketing  \(M\):
 none \(M_{1}\),
 the 1st specialist \(M_{2}\).
 the 2nd specialist \(M_{3}\).
 group of  specialists \(M_{4} = M_{2} \&  M_{3} \).

~~

 Initial system structure for \(\tau_{0} \) is depicted in Fig. 24
 (including ordinal priorities of DAs),
 system structure for \(\tau_{1} \) is depicted in Fig. 25
 (including ordinal priorities of DAs),
 ordinal compatibility estimates for \(\tau_{0} \)
 are shown in Table 8,
 ordinal compatibility estimates for \(\tau_{1} \)
 are shown in Table 9.

\begin{center}
\begin{picture}(70,43)
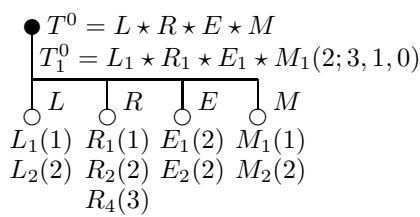
\put(04.5,00){\makebox(0,0)[bl]{Fig. 24. Team structure
  (\(\tau_{0} \))}}

\put(01,21){\makebox(0,8)[bl]{\(L_{1}(1)\)}}
\put(01,17){\makebox(0,8)[bl]{\(L_{2}(2)\)}}

\put(11,21){\makebox(0,8)[bl]{\(R_{1}(1)\)}}
\put(11,17){\makebox(0,8)[bl]{\(R_{2}(2)\)}}
\put(11,13){\makebox(0,8)[bl]{\(R_{4}(3)\)}}

\put(21,21){\makebox(0,8)[bl]{\(E_{1}(2)\)}}
\put(21,17){\makebox(0,8)[bl]{\(E_{2}(2)\)}}

\put(31,21){\makebox(0,8)[bl]{\(M_{1}(1)\)}}
\put(31,17){\makebox(0,8)[bl]{\(M_{2}(2)\)}}


\put(04,31){\line(1,0){30}}

\put(04,31){\line(0,-1){04}} \put(14,31){\line(0,-1){04}}
\put(24,31){\line(0,-1){04}} \put(34,31){\line(0,-1){04}}


\put(04,26){\circle{2}} \put(14,26){\circle{2}}
\put(24,26){\circle{2}} \put(34,26){\circle{2}}

\put(06,27){\makebox(0,8)[bl]{\(L\) }}
\put(16,27){\makebox(0,8)[bl]{\(R\) }}
\put(26,27){\makebox(0,8)[bl]{\(E\) }}
\put(36,27){\makebox(0,8)[bl]{\(M\) }}


\put(04,31){\line(0,1){07}} \put(04,38){\circle*{2}}

\put(06,37){\makebox(0,8)[bl]{\(T^{0} = L \star R \star E
  \star M\)}}

\put(05,32){\makebox(0,8)[bl]{\(T^{0}_{1} = L_{1} \star R_{1}
 \star E_{1} \star M_{1} (2;3,1,0) \) }}


\end{picture}
%
\begin{picture}(50,43)
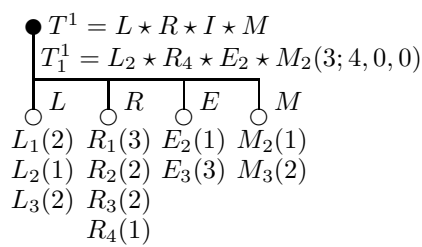
\put(04.5,00){\makebox(0,0)[bl]{Fig. 25. Team structure
 (\(\tau_{1} \))}}

\put(01,21){\makebox(0,8)[bl]{\(L_{1}(2)\)}}
\put(01,17){\makebox(0,8)[bl]{\(L_{2}(1)\)}}
\put(01,13){\makebox(0,8)[bl]{\(L_{3}(2)\)}}

\put(11,21){\makebox(0,8)[bl]{\(R_{1}(3)\)}}
\put(11,17){\makebox(0,8)[bl]{\(R_{2}(2)\)}}
\put(11,13){\makebox(0,8)[bl]{\(R_{3}(2)\)}}
\put(11,09){\makebox(0,8)[bl]{\(R_{4}(1)\)}}

\put(21,21){\makebox(0,8)[bl]{\(E_{2}(1)\)}}
\put(21,17){\makebox(0,8)[bl]{\(E_{3}(3)\)}}

\put(31,21){\makebox(0,8)[bl]{\(M_{2}(1)\)}}
\put(31,17){\makebox(0,8)[bl]{\(M_{3}(2)\)}}


\put(04,31){\line(1,0){30}}

\put(04,31){\line(0,-1){04}} \put(14,31){\line(0,-1){04}}
\put(24,31){\line(0,-1){04}} \put(34,31){\line(0,-1){04}}


\put(04,26){\circle{2}} \put(14,26){\circle{2}}
\put(24,26){\circle{2}} \put(34,26){\circle{2}}

\put(06,27){\makebox(0,8)[bl]{\(L\) }}
\put(16,27){\makebox(0,8)[bl]{\(R\) }}
\put(26,27){\makebox(0,8)[bl]{\(E\) }}
\put(36,27){\makebox(0,8)[bl]{\(M\) }}


\put(04,31){\line(0,1){07}} \put(04,38){\circle*{2}}

\put(06,37){\makebox(0,8)[bl]{\(T^{1} = L \star R \star I
 \star  M\)}}

\put(05,32){\makebox(0,8)[bl]{\(T^{1}_{1} = L_{2} \star R_{4}
\star E_{2} \star M_{2} (3;4,0,0) \) }}


\end{picture}
\end{center}

 Optimal solutions are the following:

 (a) for \(\tau_{0}\):~
 \(T^{0}_{1} = L_{1} \star R_{1} \star E_{1} \star M_{1}\),
 \( N(T^{0}_{1}) = (2;3,1,0) \),

 (b) for \(\tau_{1}\):~
 \(T^{1}_{1} = L_{2} \star R_{4} \star E_{2} \star M_{2}\),
 \( N(T^{1}_{1}) = (3;4,0,0) \).


 Here, the restructuring problem is considered as
  one-stage restructuring:

~~

 Find a solution \( T^{*}\) while taking into account the
 following:

  (i) \( H(T^{0} \rightarrow  T^{*}) \rightarrow \min \),
%
  ~(ii) \(\rho ( T^{*}, T^{1} )  \rightarrow \min  \).

~~

 It is assumed the following (for simplification):

 (a)   transformation cost
 \( H(T^{0} \rightarrow  T^{*})\)
 equals the number of change operations (by DAs);

 (b) proximity \(\rho ( T^{*}, T^{1} )\) equals
 a two-component vector
 \((\rho_{1},\rho_{2})\)
 (e.g., \cite{lev98}):
 \(\rho_{1}\) is the number of improvement steps by elements,
 \(\rho_{2}\) is  the number of improvement steps by compatibility.

 Two restructuring solutions are considered
 (evaluation of solution quality \(N(T)\) is calculated for \(\tau_{1}\)):

 (i)
 \(T^{*1}_{1} = L_{1} \star R_{3} \star E_{2} \star M_{2}\),
 \( N(T^{*1}_{1}) = (1;2,2,0) \),
 \( H(T^{0} \rightarrow  T^{*1}_{1}) = 2\),
 \(\rho ( T^{*1}_{1}, T^{1}) = (2,2) \);

 (ii)
 \(T^{*1}_{2} = L_{1} \star R_{1} \star E_{2} \star M_{2}\),
 \( N(T^{*1}_{2}) = (2;2,1,1) \),
 \( H(T^{0} \rightarrow  T^{*1}_{2}) =2 \),
 \(\rho ( T^{*1}_{2}, T^{1}) = (3,1) \).


 Further, additional stage is examined for
 \( \tau_{2}\)
 (Fig. 26, Table 10)
 and two-stage restructuring problem
 is considered for time moments:
 \( \{\tau_{0}, \tau_{1},\tau_{2} \} \).
 {\it Scheme 3} (composition \& selection solving process)
  above is used for the designing the solution trajectory.

 First, new combinatorial synthesis problem has to be solved
 for \(\tau_{2}\) (Fig. 26, Table 10).

 The solution is (Fig. 26):
 \(T^{2}_{1} = L_{3} \star R_{5} \star E_{4} \star M_{4}\),
 \(N(T^{2}_{1}) = (3;4,0,0)\).

 Second,
 the restructuring problem is examined
 (the second stage)
 for two initial solutions
 (\i.e., for \(\tau_{1}\)):~
 \(T^{*1}_{1} = L_{1} \star R_{3} \star E_{2} \star M_{2}\),
 and
 \(T^{*1}_{2} = L_{1} \star R_{1} \star E_{2} \star M_{2}\),
 This restructuring problem is considered as
  one-stage restructuring for the second stage
 (for two solutions
 \( T^{*1}_{1}\),\( T^{*1}_{2}\)):

~~

 Find a solution \( T^{2*}\) while taking into account the
 following (\(i=1,2\)):

  (i) \( H(T^{*1}_{i} \rightarrow  T^{2}_{1}) \rightarrow \min \),
%
  ~(ii) \(\rho ( T^{2}_{1}, T^{*1}_{i} )  \rightarrow \min  \).

~~

 As a result,
 the following restructuring solutions considered
  (for \(\tau_{2}\)):

 (i) for \(T^{*1}_{1}\):
 \(T^{*2}_{1} = L_{3} \star R_{3} \star E_{2} \star M_{4}\),
 \( N(T^{*2}_{1}) = (2;2,2,0) \),
 \( H(T^{*1}_{1} \rightarrow  T^{*2}_{1}) = 2\),
 \(\rho ( T^{*2}_{1}, T^{2}) = (2,1) \);

 (ii) for \(T^{*1}_{2}\):
 \(T^{*2}_{2} = L_{3} \star R_{5} \star E_{4} \star M_{3}\),
 \( N(T^{*2}_{2}) = (1;3,1,0) \),
 \( H(T^{1*}_{2}) \rightarrow  T^{*2}_{2}) = 4\),
 \(\rho ( T^{*2}_{2}, T^{2}) = (1,2) \).

 Third,
 composition of solution trajectories.
 The alternative trajectories are:
 ~\(S^{rest}_{1} = <T^{0}, T^{*1}_{1}, T^{*2}_{1} >\) and
 ~\(S^{rest}_{2} = <T^{0}, T^{*1}_{2}, T^{*2}_{2} >\).
 Estimates (i.e., integrated estimate of proximity and integrated estimate of transformation cost)
 of the trajectories are:

 (a) \(\widetilde{H}(S^{rest}_{1})=4\), \(\widetilde{\rho}(S^{rest}_{1})=(4,3)\);

 (b) \(\widetilde{H}(S^{rest}_{2})=6\), \(\widetilde{\rho}(S^{rest}_{2})=(4,3)\).

\begin{center}
{\bf Table 8.} Compatibility estimates (\(\tau_{0}\)) \\
\begin{tabular}{| l | c c c c c c c|}
\hline
 &  \(R_{1}\)&\(R_{2}\)&\(R_{4}\)&\(E_{1}\)&\(E_{2}\)&\(M_{1}\)&\(M_{2}\)\\
\hline
 \(L_{1}\) &\(2\)&\(2\)&\(1\)&\(3\)&\(2\)&\(3\)&\(1\)\\
 \(L_{2}\) &\(2\)&\(2\)&\(2\)&\(2\)&\(2\)&\(2\)&\(2\)\\

 \(R_{1}\) &&&&\(3\)&\(3\)&\(3\)&\(1\)\\
 \(R_{2}\) &&&&\(1\)&\(2\)&\(2\)&\(1\)\\
 \(R_{4}\) &&&&\(1\)&\(2\)&\(3\)&\(3\)\\

 \(E_{1}\) &&&&&&\(3\)&\(1\)\\
 \(E_{2}\) &&&&&&\(1\)&\(2\)\\
\hline
\end{tabular}
\end{center}

\begin{center}
{\bf Table 9.} Compatibility estimates (\(\tau_{1} \))\\

\begin{tabular}{| l | c c c c c c c c|}
\hline
 &  \(R_{1}\)&\(R_{2}\)&\(R_{3}\)&\(R_{4}\)&\(E_{2}\)&\(E_{3}\)&\(M_{2}\)&\(M_{3}\)\\
\hline
 \(L_{1}\) &\(3\)&\(2\)&\(2\)&\(2\)&\(2\)&\(3\)&\(2\)&\(2\)\\
 \(L_{2}\) &\(1\)&\(3\)&\(2\)&\(3\)&\(3\)&\(3\)&\(3\)&\(2\)\\
 \(L_{3}\) &\(2\)&\(2\)&\(2\)&\(2\)&\(2\)&\(2\)&\(2\)&\(2\)\\

 \(R_{1}\) &&&&&\(3\)&\(2\)&\(2\)&\(2\)\\
 \(R_{2}\) &&&&&\(3\)&\(2\)&\(3\)&\(2\)\\
 \(R_{3}\) &&&&&\(1\)&\(3\)&\(3\)&\(3\)\\
 \(R_{4}\) &&&&&\(3\)&\(3\)&\(3\)&\(3\)\\

 \(E_{2}\) &&&&&&&\(3\)&\(2\)\\
 \(E_{3}\) &&&&&&&\(2\)&\(2\)\\

\hline
\end{tabular}
\end{center}

\begin{center}
\begin{picture}(50,43)
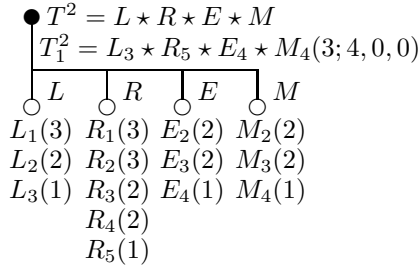
\put(04.5,00){\makebox(0,0)[bl]{Fig. 26. Team structure
     (\(\tau_{2} \))}}

\put(01,21){\makebox(0,8)[bl]{\(L_{1}(3)\)}}
\put(01,17){\makebox(0,8)[bl]{\(L_{2}(2)\)}}
\put(01,13){\makebox(0,8)[bl]{\(L_{3}(1)\)}}

\put(11,21){\makebox(0,8)[bl]{\(R_{1}(3)\)}}
\put(11,17){\makebox(0,8)[bl]{\(R_{2}(3)\)}}
\put(11,13){\makebox(0,8)[bl]{\(R_{3}(2)\)}}
\put(11,09){\makebox(0,8)[bl]{\(R_{4}(2)\)}}
\put(11,05){\makebox(0,8)[bl]{\(R_{5}(1)\)}}

\put(21,21){\makebox(0,8)[bl]{\(E_{2}(2)\)}}
\put(21,17){\makebox(0,8)[bl]{\(E_{3}(2)\)}}
\put(21,13){\makebox(0,8)[bl]{\(E_{4}(1)\)}}

\put(31,21){\makebox(0,8)[bl]{\(M_{2}(2)\)}}
\put(31,17){\makebox(0,8)[bl]{\(M_{3}(2)\)}}
\put(31,13){\makebox(0,8)[bl]{\(M_{4}(1)\)}}


\put(04,31){\line(1,0){30}}

\put(04,31){\line(0,-1){04}} \put(14,31){\line(0,-1){04}}
\put(24,31){\line(0,-1){04}} \put(34,31){\line(0,-1){04}}


\put(04,26){\circle{2}} \put(14,26){\circle{2}}
\put(24,26){\circle{2}} \put(34,26){\circle{2}}

\put(06,27){\makebox(0,8)[bl]{\(L\) }}
\put(16,27){\makebox(0,8)[bl]{\(R\) }}
\put(26,27){\makebox(0,8)[bl]{\(E\) }}
\put(36,27){\makebox(0,8)[bl]{\(M\) }}


\put(04,31){\line(0,1){07}} \put(04,38){\circle*{2}}

\put(06,37){\makebox(0,8)[bl]{\(T^{2} = L \star R \star E
 \star  M\)}}

\put(05,32){\makebox(0,8)[bl]{\(T^{2}_{1} = L_{3} \star R_{5}
\star E_{4} \star M_{4} (3;4,0,0) \) }}

\end{picture}
\end{center}

\begin{center}
{\bf Table 10.} Compatibility estimates  (\(\tau_{2} \))\\

\begin{tabular}{| l | c c c c c c c c c c c|}
\hline
 &  \(R_{1}\)&\(R_{2}\)&\(R_{3}\)&\(R_{4}\)&\(R_{5}\)&\(E_{2}\)&\(E_{3}\)&\(E_{4}\)&\(M_{2}\)&\(M_{3}\)&\(M_{4}\)\\
\hline
 \(L_{1}\) &\(3\)&\(0\)&\(2\)&\(2\)&\(1\) &\(3\)&\(3\)&\(2\) &\(3\)&\(2\)&\(2\)\\
 \(L_{2}\) &\(0\)&\(3\)&\(2\)&\(2\)&\(1\) &\(3\)&\(2\)&\(2\) &\(3\)&\(2\)&\(2\)\\
 \(L_{3}\) &\(0\)&\(2\)&\(2\)&\(3\)&\(3\) &\(2\)&\(2\)&\(3\) &\(2\)&\(1\)&\(3\)\\

 \(R_{1}\) &&&&&&\(3\)&\(2\)&\(2\) &\(3\)&\(2\)&\(2\)\\
 \(R_{2}\) &&&&&&\(1\)&\(3\)&\(2\) &\(2\)&\(3\)&\(2\)\\
 \(R_{3}\) &&&&&&\(2\)&\(3\)&\(2\) &\(2\)&\(3\)&\(2\)\\
 \(R_{4}\) &&&&&&\(1\)&\(3\)&\(3\) &\(2\)&\(3\)&\(3\)\\
 \(R_{5}\) &&&&&&\(1\)&\(2\)&\(3\) &\(1\)&\(2\)&\(3\)\\

 \(E_{2}\) &&&&&&&&&\(3\)&\(2\)&\(2\)\\
 \(E_{3}\) &&&&&&&&&\(2\)&\(3\)&\(1\)\\
 \(E_{4}\) &&&&&&&&&\(1\)&\(2\)&\(3\)\\

\hline
\end{tabular}
\end{center}

 Fourth,
 the best solution restructuring trajectory is
  (selected by Pareto rule) (Fig. 27):

 \(S^{rest}_{1} = <T^{0}_{1}, T^{*1}_{1}, T^{*2}_{1} >\).

 Table 11 contains ordinal estimates of compatibility
 (expert judgment) between
 DAs for the composite system at time stages.
 The final Pareto-efficient system trajectory is
 (hierarchical combinatorial synthesis) (Fig. 27):
 ~\(\alpha = < S^{1}_{1},  S^{2}_{1}, S^{3}_{1}  > \).

\begin{center}
{\bf Table 11.} Local (one-stage) estimates  \\
\begin{tabular}{| l | c c c c |}
\hline
 &
  \(S^{2}_{1}\)&\(S^{2}_{2}\)&\(S^{3}_{1}\)&\(S^{3}_{2}\)\\

\hline

 \(S^{1}_{1}\)  &\(3\)&\(0\) &\(3\)&\(0\) \\


  \(S^{2}_{1}\)  &\( \)&\( \) &\(3\)&\(2\)\\
  \(S^{2}_{2}\)  &\( \)&\( \) &\(3\)&\(3\) \\

\hline
\end{tabular}
\end{center}

\begin{center}
\begin{picture}(82,37)
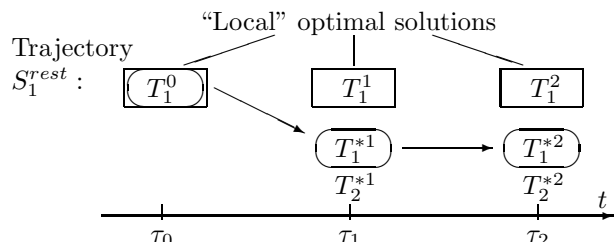
\put(03.5,00){\makebox(0,0)[bl] {Fig. 27. Illustration of
 restructuring trajectory}}

\put(12,8){\vector(1,0){68}} \put(78,9){\makebox(0,8)[bl]{\(t\)}}

\put(18.5,4){\makebox(0,8)[bl]{\(\tau_{0}\)}}
\put(43.5,4){\makebox(0,8)[bl]{\(\tau_{1}\)}}
\put(68.5,4){\makebox(0,8)[bl]{\(\tau_{2}\)}}

\put(20,7.5){\line(0,1){2}} \put(45,7.5){\line(0,1){2}}
\put(70,7.5){\line(0,1){2}}


\put(00,28.5){\makebox(0,8)[bl]{Trajectory}}
\put(00,24){\makebox(0,8)[bl]{\(S^{rest}_{1}:\)}}


\put(18,23){\makebox(0,8)[bl]{\(T^{0}_{1}\)}}

\put(15,22.5){\line(1,0){11}} \put(15,27.5){\line(1,0){11}}
\put(15,22.5){\line(0,1){5}} \put(26,22.5){\line(0,1){5}}

\put(20.5,25){\oval(10,5)}

\put(27,25){\vector(2,-1){12}}


\put(25,32){\makebox(0,8)[bl]{``Local'' optimal  solutions}}

\put(35,32){\line(-3,-1){12}} \put(45.5,32){\line(0,-1){4}}
\put(56,32){\line(3,-1){12}}


\put(44,23){\makebox(0,8)[bl]{\(T^{1}_{1}\)}}

\put(40,22.5){\line(1,0){11}} \put(40,27.5){\line(1,0){11}}
\put(40,22.5){\line(0,1){5}} \put(51,22.5){\line(0,1){5}}


\put(43,15){\makebox(0,8)[bl]{\(T^{*1}_{1}\)}}
\put(43,10){\makebox(0,8)[bl]{\(T^{*1}_{2}\)}}

\put(45.5,17){\oval(10,5)}

\put(52,17){\vector(1,0){12}}


\put(69,23){\makebox(0,8)[bl]{\(T^{2}_{1}\)}}

\put(65,22.5){\line(1,0){11}} \put(65,27.5){\line(1,0){11}}
\put(65,22.5){\line(0,1){5}} \put(76,22.5){\line(0,1){5}}


\put(68,15){\makebox(0,8)[bl]{\(T^{*2}_{1}\)}}
\put(68,10){\makebox(0,8)[bl]{\(T^{*2}_{2}\)}}

\put(70.5,17){\oval(10,5)}

\end{picture}
\end{center}

\subsection{Restructuring in clustering}

 Now, one-stage and multi-stage restructuring for
 clustering/classificaiton is described
 (based on \cite{lev15c}).
 The one-stage restructuring process in clustering problem
  is depicted in Fig. 28.

\begin{center}
\begin{picture}(80,56)
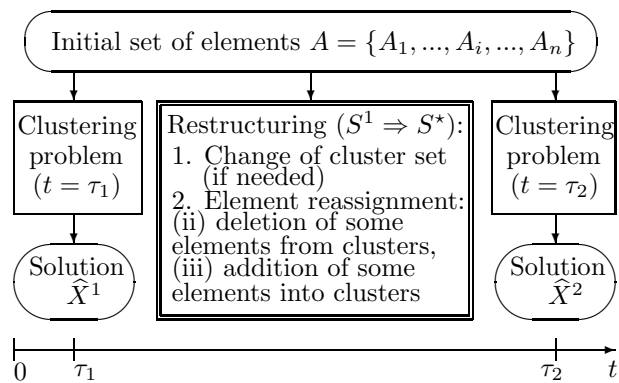
\put(05,00){\makebox(0,0)[bl]{Fig. 28. Restructuring in
 clustering problem}}

\put(00,10){\vector(1,0){80}}

\put(00,8.5){\line(0,1){3}} \put(8,8.5){\line(0,1){3}}
\put(72,8.5){\line(0,1){3}}

\put(00,06){\makebox(0,0)[bl]{\(0\)}}
\put(8,06){\makebox(0,0)[bl]{\(\tau_{1}\)}}
\put(70,06){\makebox(0,0)[bl]{\(\tau_{2}\)}}


\put(79,06.3){\makebox(0,0)[bl]{\(t\)}}


\put(00,28){\line(1,0){17}} \put(00,43){\line(1,0){17}}
\put(00,28){\line(0,1){15}} \put(17,28){\line(0,1){15}}

\put(0.5,38){\makebox(0,0)[bl]{Clustering}}
\put(2.0,34){\makebox(0,0)[bl]{problem}}
\put(2.5,30){\makebox(0,0)[bl]{(\(t=\tau_{1})\)}}


\put(8,28){\vector(0,-1){4}}

\put(8,19){\oval(16,10)}

\put(02,20){\makebox(0,0)[bl]{Solution}}

\put(07,16){\makebox(0,0)[bl]{\(\widehat{X}^{1}\)}}


\put(19,14){\line(1,0){42}} \put(19,43){\line(1,0){42}}
\put(19,14){\line(0,1){29}} \put(61,14){\line(0,1){29}}

\put(19.5,14.5){\line(1,0){41}} \put(19.5,42.5){\line(1,0){41}}
\put(19.5,14.5){\line(0,1){28}} \put(60.5,14.5){\line(0,1){28}}

\put(20.4,38){\makebox(0,0)[bl]{Restructuring
 (\(S^{1} \Rightarrow S^{\star }\)):}}

\put(21,34){\makebox(0,0)[bl]{1. Change of cluster set }}
\put(25,31){\makebox(0,0)[bl]{(if needed) }}

\put(21,28){\makebox(0,0)[bl]{2. Element reassignment: }}
\put(21,25){\makebox(0,0)[bl]{(ii) deletion of some}}
\put(21,22){\makebox(0,0)[bl]{elements from clusters,}}
\put(21,19){\makebox(0,0)[bl]{(iii) addition of some}}
\put(21,16.5){\makebox(0,0)[bl]{elements into clusters}}


\put(39.5,51){\oval(76,08)}

\put(05,49){\makebox(0,0)[bl]{Initial set of elements \(A =
 \{A_{1},...,A_{i},...,A_{n}\}\)}}


\put(8,47){\vector(0,-1){4}} \put(39.5,47){\vector(0,-1){4}}
\put(72,47){\vector(0,-1){4}}



\put(63.5,28){\line(1,0){16.5}} \put(63.5,43){\line(1,0){16.5}}
\put(63.5,28){\line(0,1){15}} \put(80,28){\line(0,1){15}}

\put(64,38){\makebox(0,0)[bl]{Clustering}}
\put(65.5,34){\makebox(0,0)[bl]{problem }}
\put(66,30){\makebox(0,0)[bl]{(\(t=\tau_{2})\)}}


\put(72,28){\vector(0,-1){4}}

\put(72,19){\oval(16,10)}

\put(66,20){\makebox(0,0)[bl]{Solution}}

\put(71,16){\makebox(0,0)[bl]{\(\widehat{X}^{2}\)}}

\end{picture}
\end{center}

~~~

 {\bf  Example for restructuring in clustering.}
 Initial information involves the following:

 (i) set of elements  \(A = \{1,2,3,4,5,6,7,8,9\}\);

 (ii) initial solution 1 (\(t=\tau_{1}\)):   \(\widehat{X}^{1} \ \{X^{1}_{1},X^{1}_{2},X^{1}_{3}\}\),
  clusters
 \(X^{1}_{1} = \{1,3,8\}\),
 \(X^{1}_{2} = \{2,4,7\}\),
 \(X^{1}_{3} = \{5,6,9\}\);

 (iii)  solution 2 (\(t=\tau_{2}\)):
  \(\widehat{X}^{2} = \{X^{2}_{1},X^{1}_{2},X^{2}_{3}\}\),
  clusters
 \(X^{2}_{1} = \{2,3\}\),
 \(X^{2}_{2} = \{5,7,8\}\),
 \(X^{2}_{3} = \{1,4,6,9\}\);

 (v) general set of considered possible change operations
 (each element can be replaced,
 the number of solution clusters is not changed):

 \(O_{11}\): none,
 \(O_{12}\):
 deletion of element \(1\) from  cluster \(X^{1}\),
 addition of element \(1\) into cluster \(X^{2}\),
 \(O_{13}\):
 deletion of element \(1\) from  cluster \(X^{1}\),
 addition of element \(1\) into cluster \(X^{3}\);

  \(O_{21}\): none,
  \(O_{22}\):
 deletion of element \(2\) from  cluster \(X^{2}\),
 addition of element \(2\) into cluster \(X^{1}\),
  \(O_{23}\):
 deletion of element \(2\) from  cluster \(X^{2}\),
 addition of element \(2\) into cluster \(X^{3}\);

  \(O_{31}\): none,
 \(O_{32}\):
 deletion of element \(3\) from cluster \(X^{1}\),
 addition of element \(3\) into cluster \(X^{2}\);
  \(O_{33}\):
 deletion of element \(3\) from cluster \(X^{1}\),
 addition of element \(3\) into cluster \(X^{3}\);

  \(O_{41}\): none,
 \(O_{42}\):
 deletion of element \(4\) from cluster \(X^{2}\),
 addition of element \(4\) into cluster \(X^{1}\),
 \(O_{43}\):
 deletion of element \(4\) from cluster \(X^{2}\),
 addition of element \(4\) into cluster \(X^{3}\);

  \(O_{51}\): none,
 \(O_{52}\):
 deletion of element \(5\) from cluster \(X^{3}\),
 addition of element \(5\) into cluster \(X^{1}\),
 \(O_{53}\):
 deletion of element \(5\) from cluster \(X^{3}\),
 addition of element \(5\) into cluster \(X^{2}\);

  \(O_{61}\): none,
 \(O_{62}\):
 deletion of element \(6\) from cluster \(X^{3}\),
 addition of element \(6\) into cluster \(X^{1}\),
 \(O_{63}\):
 deletion of element \(6\) from cluster \(X^{3}\),
 addition of element \(6\) into cluster \(X^{2}\);

  \(O_{71}\): none,
 \(O_{72}\):
 deletion of element \(7\) from cluster \(X^{2}\),
 addition of element \(7\) into cluster \(X^{1}\),
 \(O_{73}\):
 deletion of element \(7\) from cluster \(X^{2}\),
 addition of element \(7\) into cluster \(X^{3}\);

  \(O_{81}\): none,
 \(O_{82}\):
 deletion of element \(8\) from cluster \(X^{1}\),
 addition of element \(8\) into cluster \(X^{2}\),
 \(O_{83}\):
 deletion of element \(8\) from cluster \(X^{1}\),
 addition of element \(8\) into cluster \(X^{3}\);

  \(O_{91}\): none,
 \(O_{92}\):
 deletion of element \(9\) from cluster \(X^{3}\),
 addition of element \(9\) into cluster \(X^{1}\),
 \(O_{93}\):
 deletion of element \(9\) from cluster \(X^{3}\),
 addition of element \(9\) into cluster \(X^{2}\).

 In this case, optimization model (multiple choice problem) is:
 \[\max ~\sum_{i=1}^{n} ~\sum_{j=1}^{3}   c(O_{ij}) x_{ij}
 ~~~ s.t.~ \sum_{i=1}^{n}  ~\sum_{j=1}^{3}  a(O_{ij}) x_{ij}  ~\leq~ b,
 ~~~ x_{ij} \in \{0,1\},  \]
 where
 \(a(O_{ij}) \) is the cost of operation \(O_{ij}\),
 \(c(O_{ij}) \) is a ``local'' profit of operation \(O_{ij}\)
  as influence on closeness of obtained solution \(X^{\star}\)
   to clustering solution \(X^{2}\).
 Generally, it is necessary to examine quality
 parameters of clustering solution as basis for
 objective function(s).

 Evidently, the compressed
 set of change operations can be analyzed:

 \(O_{1}\):
 deletion of element \(1\) from cluster \(X^{1}\),
 addition of element \(1\) into cluster \(X^{3}\);

 \(O_{2}\):
 deletion of element \(2\) from cluster \(X^{2}\),
 addition of element \(2\) into cluster \(X^{1}\);

 \(O_{3}\):
 deletion of element \(4\) from cluster \(X^{2}\),
 addition of element \(4\) into cluster \(X^{3}\);

 \(O_{4}\):
 deletion of element \(5\) from cluster \(X^{3}\),
 addition of element \(5\) into cluster \(X^{2}\);

 \(O_{5}\):
 deletion of element \(8\) from cluster \(X^{1}\),
 addition of element \(8\) into cluster \(X^{2}\).

 In this case,
 optimization model is knapsack problem:
%
%
 \[\max ~\sum_{j=1}^{9}    c(O_{j}) x_{j}
 ~~~ s.t.~ \sum_{j=1}^{9}    a(O_{j}) x_{j}  ~\leq~ b,
 ~~~ x_{j} \in \{0,1\},  \]
 where
 \(a(O_{j}) \) is the cost of operation \(O_{j}\),
 \(c(O_{j}) \) is a ``local'' profit of operation \(O_{j}\)
  as influence on closeness of obtained solution \(X^{\star}\)
   to clustering solution \(X^{2}\).

 Finally, let us point out
 an illustrative example of clustering solution
 (Fig. 29):

  \(\widehat{X}^{\star} \ \{X^{\star}_{1},X^{\star}_{2},X^{\star}_{3}\}\),
  clusters
 \(X^{\star}_{1} = \{1,2,3\}\),
 \(X^{\star}_{2} = \{7,8\}\),
 \(X^{\star}_{3} = \{4,5,6,9\}\).

\begin{center}
\begin{picture}(34,45)
\put(05.5,00){\makebox(0,0)[bl]{Fig. 29.
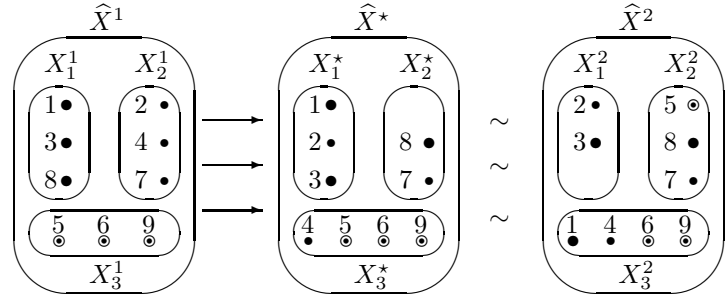
 Example:  restructuring of clustering solution}}

\put(10,40){\makebox(0,0)[bl]{\(\widehat{X}^{1}\)}}
\put(12,22){\oval(24,34)}

\put(04,33.5){\makebox(0,0)[bl]{\(X^{1}_{1}\)}}
\put(06,25){\oval(08,15)}

\put(07,20){\circle*{1.3}} \put(04,19){\makebox(0,8)[bl]{\(8\)}}
\put(07,25){\circle*{1.3}} \put(04,24){\makebox(0,8)[bl]{\(3\)}}
\put(07,30){\circle*{1.3}} \put(04,29){\makebox(0,8)[bl]{\(1\)}}

\put(16,33.5){\makebox(0,0)[bl]{\(X^{1}_{2}\)}}
\put(18,25){\oval(08,15)}

\put(20,20){\circle*{1.0}} \put(16,19){\makebox(0,8)[bl]{\(7\)}}
\put(20,25){\circle*{1.0}} \put(16,24){\makebox(0,8)[bl]{\(4\)}}
\put(20,30){\circle*{1.0}} \put(16,29){\makebox(0,8)[bl]{\(2\)}}

\put(10,05.5){\makebox(0,0)[bl]{\(X^{1}_{3}\)}}
\put(12,13){\oval(20,6)}

\put(06,12){\circle*{0.7}} \put(06,12){\circle{1.4}}
\put(05,13){\makebox(0,8)[bl]{\(5\)}}

\put(12,12){\circle*{0.7}} \put(12,12){\circle{1.4}}
\put(11,13){\makebox(0,8)[bl]{\(6\)}}

\put(18,12){\circle*{0.7}} \put(18,12){\circle{1.4}}
\put(17,13){\makebox(0,8)[bl]{\(9\)}}


\put(25,28){\vector(1,0){8}} \put(25,22){\vector(1,0){8}}
\put(25,16){\vector(1,0){8}}

\end{picture}
%
\begin{picture}(34,45)

\put(10,40){\makebox(0,0)[bl]{\(\widehat{X}^{\star}\)}}
\put(12,22){\oval(24,34)}

\put(04,33.5){\makebox(0,0)[bl]{\(X^{\star}_{1}\)}}
\put(06,25){\oval(08,15)}

\put(07,20){\circle*{1.3}} \put(04,19){\makebox(0,8)[bl]{\(3\)}}
\put(07,25){\circle*{1.0}} \put(04,24){\makebox(0,8)[bl]{\(2\)}}
\put(07,30){\circle*{1.3}} \put(04,29){\makebox(0,8)[bl]{\(1\)}}

\put(16,33.5){\makebox(0,0)[bl]{\(X^{\star}_{2}\)}}
\put(18,25){\oval(08,15)}

\put(20,20){\circle*{1.0}} \put(16,19){\makebox(0,8)[bl]{\(7\)}}
\put(20,25){\circle*{1.3}} \put(16,24){\makebox(0,8)[bl]{\(8\)}}


\put(10,05.5){\makebox(0,0)[bl]{\(X^{\star}_{3}\)}}
\put(12,13){\oval(20,6)}

\put(04,12){\circle*{1.0}} \put(03,13){\makebox(0,8)[bl]{\(4\)}}

\put(09,12){\circle*{0.7}} \put(09,12){\circle{1.4}}
\put(08,13){\makebox(0,8)[bl]{\(5\)}}

\put(14,12){\circle*{0.7}} \put(14,12){\circle{1.4}}
 \put(13,13){\makebox(0,8)[bl]{\(6\)}}

\put(19,12){\circle*{0.7}} \put(19,12){\circle{1.4}}
\put(18,13){\makebox(0,8)[bl]{\(9\)}}


\put(28,27){\makebox(0,0)[bl]{\(\sim\)}}
\put(28,21){\makebox(0,0)[bl]{\(\sim\)}}
\put(28,14){\makebox(0,0)[bl]{\(\sim\)}}

\end{picture}
%
\begin{picture}(24,45)

\put(10,40){\makebox(0,0)[bl]{\(\widehat{X}^{2}\)}}
\put(12,22){\oval(24,34)}

\put(04,33.5){\makebox(0,0)[bl]{\(X^{2}_{1}\)}}
\put(06,25){\oval(08,15)}

\put(07,25){\circle*{1.3}} \put(04,24){\makebox(0,8)[bl]{\(3\)}}
\put(07,30){\circle*{1.0}} \put(04,29){\makebox(0,8)[bl]{\(2\)}}

\put(16,33.5){\makebox(0,0)[bl]{\(X^{2}_{2}\)}}
\put(18,25){\oval(08,15)}

\put(20,20){\circle*{1.0}} \put(16,19){\makebox(0,8)[bl]{\(7\)}}
\put(20,25){\circle*{1.3}} \put(16,24){\makebox(0,8)[bl]{\(8\)}}

\put(20,30){\circle*{0.7}} \put(20,30){\circle{1.4}}
\put(16,29){\makebox(0,8)[bl]{\(5\)}}

\put(10,05.5){\makebox(0,0)[bl]{\(X^{2}_{3}\)}}
\put(12,13){\oval(20,6)}

\put(04,12){\circle*{1.3}} \put(03,13){\makebox(0,8)[bl]{\(1\)}}

\put(09,12){\circle*{1.0}} \put(08,13){\makebox(0,8)[bl]{\(4\)}}

\put(14,12){\circle*{0.7}} \put(14,12){\circle{1.4}}
 \put(13,13){\makebox(0,8)[bl]{\(6\)}}

\put(19,12){\circle*{0.7}} \put(19,12){\circle{1.4}}
\put(18,13){\makebox(0,8)[bl]{\(9\)}}



\end{picture}
\end{center}

 Fig. 30 and Fig. 31 illustrate
 multistage classification and multistage clustering problems:

 {\it 1.} Multistage classification (Fig. 30):
 the same set of classes at each time stage (here: four classes \(L^{1}\), \(L^{2}\), \(L^{3}\), \(L^{4}\)),
 elements can belong to different classes at each stage.
 Here: elements \(1\), \(2\), \(3\);
  trajectory for element \(1\): \(J(1) = <L^{1},L^{1},L^{1}>\),
  trajectory for element \(2\): \(J(2) = <L^{2},L^{1},L^{2}>\),
  trajectory for element \(3\): \(J(3) = <L^{3},L^{4},L^{3}>\)).

 {\it 2.} Multistage clustering (Fig. 31):
 different set of clusters at each time stage can be examined,
  elements can belong to different clusters at each stage.
 Here: elements \(1\), \(2\), \(3\);
  trajectory for element \(1\): \(J(1) = <L^{1}_{1},L^{1}_{2},L^{1}_{3}>\),
  trajectory for element \(2\): \(J(2) = <L^{2}_{1},L^{1}_{2},L^{2}_{3}>\),
  trajectory for element \(3\): \(J(3) = <L^{3}_{1},L^{4}_{2},L^{5}_{3}>\)).

 In this problem,
 it is necessary to examine
 a set of change trajectories for each element.
 As a result,
 multi-stage restructuring problem
 has to be based on
 multiple choice model.
 Generally, this problem is very prospective.

 This kind of clustering (or classification) model/problem is close
 to multistage system design
  \cite{fortun10,hopcroft04,lev12clique,lev13tra,lev15}.

\begin{center}
\begin{picture}(150,71)
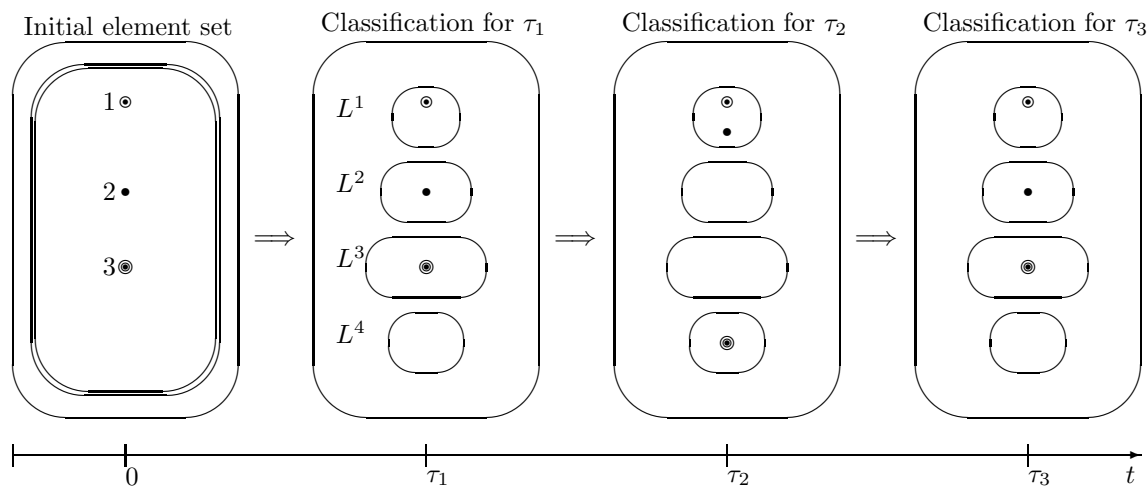
\put(36,00){\makebox(0,0)[bl]{Fig. 30. Illustration of multistage
 classification}}

\put(00,10){\vector(1,0){150}}

\put(00,8.5){\line(0,1){3}} \put(15,8.5){\line(0,1){3}}
\put(55,8.5){\line(0,1){3}} \put(95,8.5){\line(0,1){3}}
\put(135,8.5){\line(0,1){3}}

\put(15,06){\makebox(0,0)[bl]{\(0\)}}
\put(55,06){\makebox(0,0)[bl]{\(\tau_{1}\)}}
\put(95,06){\makebox(0,0)[bl]{\(\tau_{2}\)}}
\put(135,06){\makebox(0,0)[bl]{\(\tau_{3}\)}}

\put(148,06.3){\makebox(0,0)[bl]{\(t\)}}


\put(01.5,66){\makebox(0,0)[bl]{Initial element set}}

\put(15,40){\oval(30,50)}

\put(15,40){\oval(25,44)} \put(15,40){\oval(24,43)}

\put(15,57){\circle*{0.8}}  \put(15,57){\circle{1.4}}

\put(12,56){\makebox(0,0)[bl]{\(1\)}}

\put(15,45){\circle*{1.2}}

\put(12,44){\makebox(0,0)[bl]{\(2\)}}

\put(15,35){\circle*{0.7}}  \put(15,35){\circle{1.1}}
\put(15,35){\circle{1.7}}

\put(12,34){\makebox(0,0)[bl]{\(3\)}}



\put(32,38){\makebox(0,0)[bl]{\(\Longrightarrow \)}}
\put(72,38){\makebox(0,0)[bl]{\(\Longrightarrow \)}}
\put(112,38){\makebox(0,0)[bl]{\(\Longrightarrow \)}}


\put(41,66){\makebox(0,0)[bl]{Classification for \(\tau_{1}\)}}

\put(55,40){\oval(30,50)}

\put(55,57){\circle*{0.8}}  \put(55,57){\circle{1.4}}

\put(55,45){\circle*{1.2}}

\put(55,35){\circle*{0.7}}  \put(55,35){\circle{1.1}}
\put(55,35){\circle{1.7}}

\put(43,25){\makebox(0,0)[bl]{\(L^{4}\)}}
\put(43,35){\makebox(0,0)[bl]{\(L^{3}\)}}
\put(43,45){\makebox(0,0)[bl]{\(L^{2}\)}}
\put(43,55){\makebox(0,0)[bl]{\(L^{1}\)}}

\put(55,25){\oval(10,08)} \put(55,35){\oval(16,08)}
\put(55,45){\oval(12,08)} \put(55,55){\oval(09,08)}


\put(81,66){\makebox(0,0)[bl]{Classification for \(\tau_{2}\)}}

\put(95,40){\oval(30,50)}

\put(95,57){\circle*{0.8}}  \put(95,57){\circle{1.4}}

\put(95,53){\circle*{1.2}}

\put(95,25){\circle*{0.7}}  \put(95,25){\circle{1.1}}
\put(95,25){\circle{1.7}}

\put(95,25){\oval(10,08)} \put(95,35){\oval(16,08)}
\put(95,45){\oval(12,08)} \put(95,55){\oval(09,08)}


\put(121,66){\makebox(0,0)[bl]{Classification for \(\tau_{3}\)}}

\put(135,40){\oval(30,50)}

\put(135,57){\circle*{0.8}}  \put(135,57){\circle{1.4}}

\put(135,45){\circle*{1.2}}

\put(135,35){\circle*{0.7}}  \put(135,35){\circle{1.1}}
\put(135,35){\circle{1.7}}

\put(135,25){\oval(10,08)} \put(135,35){\oval(16,08)}
\put(135,45){\oval(12,08)} \put(135,55){\oval(09,08)}

\end{picture}
\end{center}

\begin{center}
\begin{picture}(150,71)
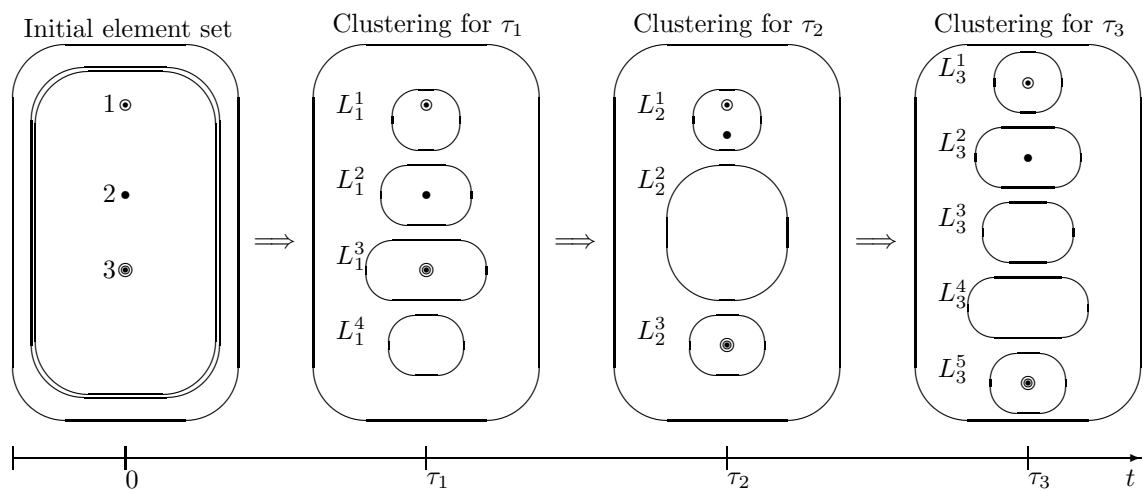
\put(38,00){\makebox(0,0)[bl]{Fig. 31. Illustration of multistage
 clustering}}

\put(00,10){\vector(1,0){150}}

\put(00,8.5){\line(0,1){3}} \put(15,8.5){\line(0,1){3}}
\put(55,8.5){\line(0,1){3}} \put(95,8.5){\line(0,1){3}}
\put(135,8.5){\line(0,1){3}}

\put(15,06){\makebox(0,0)[bl]{\(0\)}}
\put(55,06){\makebox(0,0)[bl]{\(\tau_{1}\)}}
\put(95,06){\makebox(0,0)[bl]{\(\tau_{2}\)}}
\put(135,06){\makebox(0,0)[bl]{\(\tau_{3}\)}}

\put(148,06.3){\makebox(0,0)[bl]{\(t\)}}


\put(01.5,66){\makebox(0,0)[bl]{Initial element set}}

\put(15,40){\oval(30,50)}

\put(15,40){\oval(25,44)} \put(15,40){\oval(24,43)}

\put(15,57){\circle*{0.8}}  \put(15,57){\circle{1.4}}

\put(12,56){\makebox(0,0)[bl]{\(1\)}}

\put(15,45){\circle*{1.2}}

\put(12,44){\makebox(0,0)[bl]{\(2\)}}

\put(15,35){\circle*{0.7}}  \put(15,35){\circle{1.1}}
\put(15,35){\circle{1.7}}

\put(12,34){\makebox(0,0)[bl]{\(3\)}}




\put(32,38){\makebox(0,0)[bl]{\(\Longrightarrow \)}}
\put(72,38){\makebox(0,0)[bl]{\(\Longrightarrow \)}}
\put(112,38){\makebox(0,0)[bl]{\(\Longrightarrow \)}}


\put(42.5,66){\makebox(0,0)[bl]{Clustering for \(\tau_{1}\)}}

\put(55,40){\oval(30,50)}

\put(43,25){\makebox(0,0)[bl]{\(L^{4}_{1}\)}}
\put(43,35){\makebox(0,0)[bl]{\(L^{3}_{1}\)}}
\put(43,45){\makebox(0,0)[bl]{\(L^{2}_{1}\)}}
\put(43,55){\makebox(0,0)[bl]{\(L^{1}_{1}\)}}

\put(55,25){\oval(10,08)} \put(55,35){\oval(16,08)}
\put(55,45){\oval(12,08)} \put(55,55){\oval(09,08)}

\put(55,57){\circle*{0.8}}  \put(55,57){\circle{1.4}}

\put(55,45){\circle*{1.2}}

\put(55,35){\circle*{0.7}}  \put(55,35){\circle{1.1}}
\put(55,35){\circle{1.7}}


\put(82.5,66){\makebox(0,0)[bl]{Clustering for \(\tau_{2}\)}}

\put(95,40){\oval(30,50)}

\put(83,25){\makebox(0,0)[bl]{\(L^{3}_{2}\)}}
\put(83,45){\makebox(0,0)[bl]{\(L^{2}_{2}\)}}
\put(83,55){\makebox(0,0)[bl]{\(L^{1}_{2}\)}}

\put(95,25){\oval(10,08)} \put(95,40){\oval(16,18)}


\put(95,55){\oval(09,08)}

\put(95,57){\circle*{0.8}}  \put(95,57){\circle{1.4}}

\put(95,53){\circle*{1.2}}

\put(95,25){\circle*{0.7}}  \put(95,25){\circle{1.1}}
\put(95,25){\circle{1.7}}


\put(122.5,66){\makebox(0,0)[bl]{Clustering for \(\tau_{3}\)}}

\put(135,40){\oval(30,50)}

\put(123,20){\makebox(0,0)[bl]{\(L^{5}_{3}\)}}
\put(123,30){\makebox(0,0)[bl]{\(L^{4}_{3}\)}}
\put(123,40){\makebox(0,0)[bl]{\(L^{3}_{3}\)}}
\put(123,50){\makebox(0,0)[bl]{\(L^{2}_{3}\)}}
\put(123,60){\makebox(0,0)[bl]{\(L^{1}_{3}\)}}

\put(135,20){\oval(10,08)} \put(135,30){\oval(16,08)}
\put(135,40){\oval(12,08)} \put(135,50){\oval(14,08)}
\put(135,60){\oval(09,08)}

\put(135,60){\circle*{0.8}}  \put(135,60){\circle{1.4}}

\put(135,50){\circle*{1.2}}

\put(135,20){\circle*{0.7}}  \put(135,20){\circle{1.1}}
\put(135,20){\circle{1.7}}

\end{picture}
\end{center}

\subsection{Restructuring in sorting}

 One-stage restructuring for sorting problem
 can be considered as well.
 Let \(A = \{A_{1},...,A_{i},...,A_{n}\}\) be a initial element
 set.
 Solution is a result of dividing set \{A\}
 into \(k\) linear ordered subsets (ranking):
 \(\widehat{R} = \{R_{1},...,R_{j},...,R_{k} \}\),
 \(R_{j} \subseteq A ~ \forall j=\overline{1,k} \),
 \(|R_{j_{1}} \& R_{j_{2}}|=0 ~ \forall j_{1},j_{2} \).
  Linear order is:
  \(R_{1} \rightarrow ... \rightarrow R_{j} \rightarrow ... \rightarrow  R_{k}\),
 \( A_{i_{1}} \rightarrow A_{i_{2}} \) if
 \(A_{i_{1}} \in R_{j_{1}}\), \(A_{i_{2}} \in R_{j_{2}}\),
 \( j_{1} < j_{2}  \).

 Generally, the sorting problem (or multicriteria ranking)
 consists in transformation of set \(A \)
 into ranking \(R\):~ \(A \Rightarrow R \)
 while taking into account multicriteria estimates of elements
 and/or expert judgment
 (e.g., \cite{roy96,zop02}).
 In Fig. 32,
 illustration for restructuring in sorting problem is depicted.
 The problem is:
 \[\min \delta (\widehat{R}^{2}, \widehat{R}^{\star})
  ~~~ s.t. ~~ a (\widehat{R}^{1} \rightarrow \widehat{R}^{\star}) < b,\]
 where
 \(\widehat{R}^{\star}\) is solution,
 \(\widehat{R}^{1}\) is initial (the ``first'') ranking,
 \(\widehat{R}^{2}\) is the ``second'' ranking,
 \(\delta (\widehat{R}^{\star}, \widehat{R}^{2})\) is proximity
 between solution \(\widehat{R}^{\star }\)
 and the ``second'' ranking \(\widehat{R}^{\star }\)
 (e.g., structural proximity or  proximity by quality parameters for rankings),
 \(a(\widehat{R}^{1} \rightarrow \widehat{R}^{\star})\) is the
 cost of transformation of the ``first'' ranking \( \widehat{R}^{1} \)
 into solution \(\widehat{R}^{\star}\)
 (e.g., editing ``distance''),
 \(b\) is constraint for the transformation cost.
 Evidently, multi-stage restructuring problems
 (with change trajectories of elements)
 are prospective as well.

\begin{center}
\begin{picture}(38,43)
\put(12,00){\makebox(0,0)[bl]{Fig. 32.
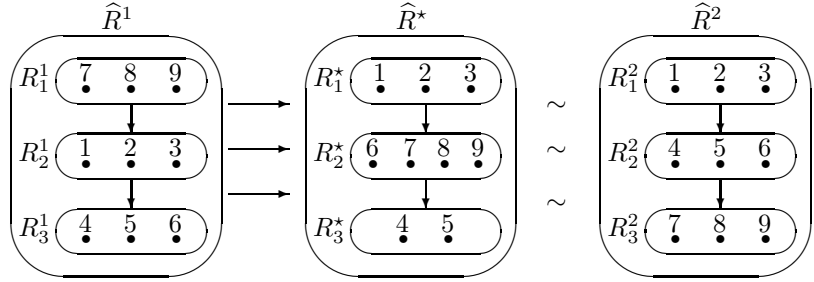
 Example:  restructuring in sorting problem}}

\put(12,38){\makebox(0,0)[bl]{\(\widehat{R}^{1}\)}}
\put(14,21){\oval(28,32)}

\put(01,29.5){\makebox(0,0)[bl]{\(R^{1}_{1}\)}}

\put(16,31){\oval(20,6)}

\put(10,30){\circle*{1.0}} \put(09,31){\makebox(0,8)[bl]{\(7\)}}
\put(16,30){\circle*{1.0}} \put(15,31){\makebox(0,8)[bl]{\(8\)}}
\put(22,30){\circle*{1.0}} \put(21,31){\makebox(0,8)[bl]{\(9\)}}

\put(16,28){\vector(0,-1){4}}

\put(01,19.5){\makebox(0,0)[bl]{\(R^{1}_{2}\)}}

\put(16,21){\oval(20,6)}

\put(10,20){\circle*{1.0}} \put(09,21){\makebox(0,8)[bl]{\(1\)}}
\put(16,20){\circle*{1.0}} \put(15,21){\makebox(0,8)[bl]{\(2\)}}
\put(22,20){\circle*{1.0}} \put(21,21){\makebox(0,8)[bl]{\(3\)}}

\put(16,18){\vector(0,-1){4}}

\put(01,09.5){\makebox(0,0)[bl]{\(R^{1}_{3}\)}}

\put(16,11){\oval(20,6)}

\put(10,10){\circle*{1.0}} \put(09,11){\makebox(0,8)[bl]{\(4\)}}
\put(16,10){\circle*{1.0}} \put(15,11){\makebox(0,8)[bl]{\(5\)}}
\put(22,10){\circle*{1.0}} \put(21,11){\makebox(0,8)[bl]{\(6\)}}


\put(29,28){\vector(1,0){8}} \put(29,22){\vector(1,0){8}}
\put(29,16){\vector(1,0){8}}

\end{picture}
%
\begin{picture}(38,43)

\put(12,38){\makebox(0,0)[bl]{\(\widehat{R}^{\star}\)}}
\put(14,21){\oval(28,32)}

\put(01,29.5){\makebox(0,0)[bl]{\(R^{\star}_{1}\)}}

\put(16,31){\oval(20,6)}

\put(10,30){\circle*{1.0}} \put(09,31){\makebox(0,8)[bl]{\(1\)}}
\put(16,30){\circle*{1.0}} \put(15,31){\makebox(0,8)[bl]{\(2\)}}
\put(22,30){\circle*{1.0}} \put(21,31){\makebox(0,8)[bl]{\(3\)}}

\put(16,28){\vector(0,-1){4}}

\put(01,19.5){\makebox(0,0)[bl]{\(R^{\star}_{2}\)}}

\put(16,21){\oval(20,6)}

\put(09,20){\circle*{1.0}} \put(08,21){\makebox(0,8)[bl]{\(6\)}}
\put(14,20){\circle*{1.0}} \put(13,21){\makebox(0,8)[bl]{\(7\)}}

\put(18.5,20){\circle*{1.0}}
\put(17.5,21){\makebox(0,8)[bl]{\(8\)}}

\put(23,20){\circle*{1.0}} \put(22,21){\makebox(0,8)[bl]{\(9\)}}

\put(16,18){\vector(0,-1){4}}

\put(01,09.5){\makebox(0,0)[bl]{\(R^{\star}_{3}\)}}

\put(16,11){\oval(20,6)}

\put(13,10){\circle*{1.0}} \put(12,11){\makebox(0,8)[bl]{\(4\)}}
\put(19,10){\circle*{1.0}} \put(18,11){\makebox(0,8)[bl]{\(5\)}}


\put(32,27){\makebox(0,0)[bl]{\(\sim\)}}
\put(32,21){\makebox(0,0)[bl]{\(\sim\)}}
\put(32,14){\makebox(0,0)[bl]{\(\sim\)}}

\end{picture}
%
\begin{picture}(28,43)
\put(12,38){\makebox(0,0)[bl]{\(\widehat{R}^{2}\)}}
\put(14,21){\oval(28,32)}

\put(01,29.5){\makebox(0,0)[bl]{\(R^{2}_{1}\)}}

\put(16,31){\oval(20,6)}

\put(10,30){\circle*{1.0}} \put(09,31){\makebox(0,8)[bl]{\(1\)}}
\put(16,30){\circle*{1.0}} \put(15,31){\makebox(0,8)[bl]{\(2\)}}
\put(22,30){\circle*{1.0}} \put(21,31){\makebox(0,8)[bl]{\(3\)}}

\put(16,28){\vector(0,-1){4}}

\put(01,19.5){\makebox(0,0)[bl]{\(R^{2}_{2}\)}}

\put(16,21){\oval(20,6)}

\put(10,20){\circle*{1.0}} \put(09,21){\makebox(0,8)[bl]{\(4\)}}
\put(16,20){\circle*{1.0}} \put(15,21){\makebox(0,8)[bl]{\(5\)}}
\put(22,20){\circle*{1.0}} \put(21,21){\makebox(0,8)[bl]{\(6\)}}

\put(16,18){\vector(0,-1){4}}

\put(01,09.5){\makebox(0,0)[bl]{\(R^{2}_{3}\)}}

\put(16,11){\oval(20,6)}

\put(10,10){\circle*{1.0}} \put(09,11){\makebox(0,8)[bl]{\(7\)}}
\put(16,10){\circle*{1.0}} \put(15,11){\makebox(0,8)[bl]{\(8\)}}
\put(22,10){\circle*{1.0}} \put(21,11){\makebox(0,8)[bl]{\(9\)}}

\end{picture}
\end{center}

\subsection{Spanning trees problems}

 Let us present the restructuring approach for basic spanning trees
 problems from \cite{lev11restr}.
 Restructuring problems for
 minimal spanning tree problem and for Steiner tree problem
 are described as follows (Fig. 33, Fig. 34).
 The following numerical examples are presented:

 {\bf I.} Initial graph (Fig. 33): ~\(G=(A,E)\), where ~\(A=\{1,2,3,4,5,6,7\}\),

 \(E=\{(1,2),(1,4),(1,5),(1,6),(2,3),(2,6),(3,6),(4,5),(4,6),(5,6),(5,7),\)

 \((6,7)\}\).

 {\bf II.} Spanning trees (Fig. 33):

 (i) \(T^{1} = (A,E^{1})\), where
 \(E^{1}=\{(1,2),(1,4),(1,6),(3,5),(5,6),(6,7)\}\),

 (ii) \(T^{2} = (A,E^{2})\), where
 \(E^{2}=\{(1,2),(2,3),(2,6),(4,6),(5,6),(6,7)\}\),

 (iii) \(T^{*} = (A,E^{*})\), where
 \(E^{*}=\{(1,2),(1,4),(2,3),(2,6),(3,5),(6,7)\}\).

  Here the edge changes are (\(T^{1} \rightarrow T^{*}\) as \(E^{1} \rightarrow E^{*}\)):

  \(E^{1*-} = \{(1,6),(5,6)\}\)
  and
  ~\(E^{1*+} = \{(2,3),(2,6)\}\).

\begin{center}
\begin{picture}(95,47)
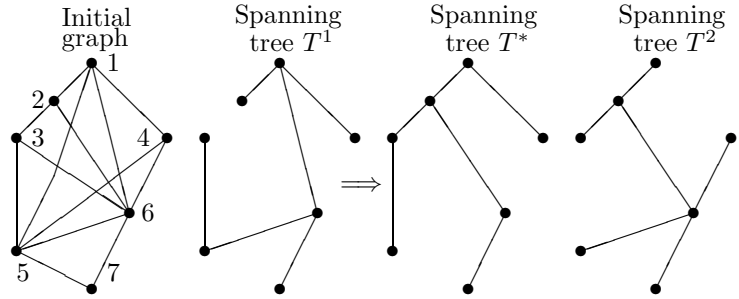
\put(018.1,00){\makebox(0,0)[bl]{Fig. 33. Restructuring of
 spanning tree}}


\put(6,42.9){\makebox(0,0)[bl]{Initial}}
\put(6,39.6){\makebox(0,0)[bl]{graph}}


\put(12,37){\makebox(0,0)[bl]{\(1\)}}
\put(02,32){\makebox(0,0)[bl]{\(2\)}}
\put(02,27){\makebox(0,0)[bl]{\(3\)}}
\put(16,27){\makebox(0,0)[bl]{\(4\)}}
\put(16.6,17){\makebox(0,0)[bl]{\(6\)}}
\put(00,09){\makebox(0,0)[bl]{\(5\)}}
\put(12,09){\makebox(0,0)[bl]{\(7\)}}


\put(00,28){\circle*{1.3}}

\put(00,28){\line(1,1){5}}

\put(00,28){\line(0,-1){15}}

\put(00,28){\line(3,-2){15}}


\put(05,33){\circle*{1.3}}

\put(05,33){\line(1,1){5}}

\put(10,38){\line(1,-4){5}}

\put(10,38){\line(1,-1){10}}

\put(10,38){\circle*{1.3}}

\put(20,28){\circle*{1.3}}

\put(20,28){\line(-4,-3){20}}

\put(20,28){\line(-1,-2){5}}

\put(15,18){\line(-2,3){10}}

\put(15,18){\circle*{1.3}}

\put(15,18){\line(-3,-1){15}}

\put(15,18){\line(-1,-2){5}}

\put(00,13){\line(1,2){5}} \put(05,23){\line(1,3){5}}

\put(00,13){\circle*{1.3}}

\put(10,8){\line(-2,1){10}}

\put(10,8){\circle*{1.3}}






\put(29,42.6){\makebox(0,0)[bl]{Spanning}}
\put(31,39.6){\makebox(0,0)[bl]{tree \(T^{1}\)}}

\put(25,28){\circle*{1.3}}


\put(25,28){\line(0,-1){15}}



\put(30,33){\circle*{1.3}}

\put(30,33){\line(1,1){5}}

\put(35,38){\line(1,-4){5}}

\put(35,38){\line(1,-1){10}}

\put(35,38){\circle*{1.3}}

\put(45,28){\circle*{1.3}}



\put(40,18){\circle*{1.3}}

\put(40,18){\line(-3,-1){15}}

\put(40,18){\line(-1,-2){5}}


\put(25,13){\circle*{1.3}}


\put(35,8){\circle*{1.3}}





\put(43,21){\makebox(0,0)[bl]{\(\Longrightarrow\)}}




\put(55,42.6){\makebox(0,0)[bl]{Spanning}}
\put(57,39.6){\makebox(0,0)[bl]{tree \(T^{*}\)}}

\put(50,28){\circle*{1.3}}

\put(50,28){\line(1,1){5}}

\put(50,28){\line(0,-1){15}}


\put(55,33){\circle*{1.3}}

\put(55,33){\line(1,1){5}}

\put(60,38){\circle*{1.3}}

\put(70,28){\circle*{1.3}}


\put(70,28){\line(-1,1){10}}

\put(65,18){\line(-2,3){10}}

\put(65,18){\circle*{1.3}}


\put(65,18){\line(-1,-2){5}}

\put(50,13){\circle*{1.3}}

\put(60,8){\circle*{1.3}}


\put(80,42.6){\makebox(0,0)[bl]{Spanning}}
\put(82,39.6){\makebox(0,0)[bl]{tree \(T^{2}\)}}

\put(75,28){\circle*{1.3}}

\put(75,28){\line(1,1){5}}




\put(80,33){\circle*{1.3}}

\put(80,33){\line(1,1){5}}



\put(85,38){\circle*{1.3}}

\put(95,28){\circle*{1.3}}

%

\put(95,28){\line(-1,-2){5}}

\put(90,18){\line(-2,3){10}}

\put(90,18){\circle*{1.3}}

\put(90,18){\line(-3,-1){15}}

\put(90,18){\line(-1,-2){5}}


\put(75,13){\circle*{1.3}}


\put(85,8){\circle*{1.3}}

\end{picture}
\end{center}

 {\bf III.} Steiner trees (Fig. 34, set of possible Steiner vertices is \(Z = \{a,b,c,d\}\)):

 (i) \(S^{1} = (A^{1},E^{1})\), where ~\(A^{1}=A\bigcup Z^{1}\), ~\(Z^{1}=\{a,b\}\),

 \(E^{1}=\{(1,2),(1,a),(a,4),(a,6),(3,5),(b,5),(b,6),(b,7)\}\),

 (ii) \(S^{2} = (A^{2},E^{2})\), where ~\(A^{2}=A\bigcup Z^{2}\), ~\(Z^{2}=\{a,b,d\}\),

 \(E^{2}=\{(3,4),(1,d),(3,d),(a,d),(a,4),(a,6),(b,6),(b,5)),(b,7)\}\),

 (iii) \(S^{*} = (A^{*},E^{*})\), where where ~\(A^{*}=A\bigcup Z^{*}\), ~\(Z^{*}=\{a,c\}\),

 \(E^{*}=\{(1,2),(1,a),(a,4),(a,6),(c,3),(c,5),(c,6),(6,7)\}\).

 Thus, the restructuring problem for spanning tree is
 (Fig. 33, a simple version):
  \[\min \rho ( T^{*} , T^{2})
  ~~~~ s.t. ~~~
   H(S^{1} \Rightarrow S^{*}) = ( \sum_{i \in  E^{1*-} } h^{-}_{i} +  \sum_{i \in E^{1*+}} h^{+}_{i} ~) \leq \widehat{h},\]
 where \(\widehat{h}\) is a constraint for the change cost,
 \(h^{-}(i)\) is a cost of deletion of element
 (i.e., edge) \(i \in E^{1}\), and
  \(h^{+}(i)\) is a cost of addition of element
  (i.e., edge) \(i  \in E \backslash  E^{1}\).

 The restructuring problem for Steiner tree is
 (Fig. 34, a simple version):
  \[\min \rho ( S^{*} , S^{2})\]
  \[s.t. ~~
   H(S^{1} \Rightarrow S^{*}) = ( \sum_{i \in  E^{1*-} } h^{-}_{i} +  \sum_{i \in E^{1*+}} h^{+}_{i}
   ~)  +
  ( \sum_{i \in  Z^{1*-} } w^{-}_{i} +  \sum_{i \in Z^{1*+}} w^{+}_{i} ~)
    \leq \widehat{h},\]
 where \(\widehat{h}\) is a constraint for the change cost,
 \(h^{-}(i)\) is a cost of deletion of element
 (i.e., edge) \(i \in E^{1}\),
  \(h^{+}(i)\) is a cost of addition of element
  (i.e., edge)
  \(i  \in  \widehat{E}^{*} \subseteq E \backslash  E^{1} \),
  \(w^{-}(j)\) is a cost of deletion of
  Steiner vertex \(j \in Z^{1}\),
  \(w^{+}(j)\) is a cost of addition of Steiner vertex
  \(j  \in \widehat{Z}^{*} \subseteq  Z \backslash Z^{1}\).

\begin{center}
\begin{picture}(70,47)
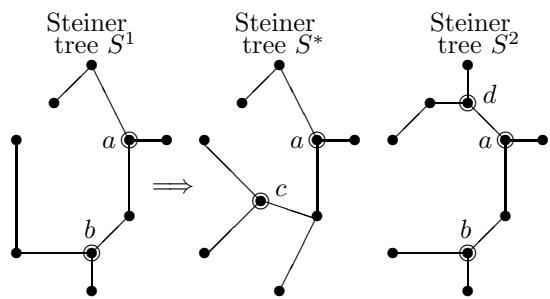
\put(07,00){\makebox(0,0)[bl]{Fig. 34. Restructuring of Steiner
tree}}

\put(04,42.6){\makebox(0,0)[bl]{Steiner}}
\put(05,39.6){\makebox(0,0)[bl]{tree \(S^{1}\)}}

\put(00,28){\circle*{1.3}}


\put(00,28){\line(0,-1){15}}


\put(05,33){\circle*{1.3}}

\put(05,33){\line(1,1){5}}

\put(10,38){\circle*{1.3}}

\put(20,28){\circle*{1.3}}

\put(15,18){\circle*{1.3}}

\put(15,28){\circle*{1.3}} \put(15,28){\circle{2}}
\put(15,28){\line(1,0){5}} \put(15,28){\line(0,-1){10}}

\put(15,28){\line(-1,2){5}}

\put(11.4,27){\makebox(0,0)[bl]{\(a\)}}


\put(00,13){\circle*{1.3}}

\put(10,13){\circle*{1.3}} \put(10,13){\circle{2}}
\put(10,13){\line(-1,0){10}} \put(10,13){\line(0,-1){5}}

\put(10,13){\line(1,1){5}}

\put(09,15){\makebox(0,0)[bl]{\(b\)}}

\put(10,8){\circle*{1.3}}


\put(18,21){\makebox(0,0)[bl]{\(\Longrightarrow\)}}


\put(29,42.6){\makebox(0,0)[bl]{Steiner}}
\put(30,39.6){\makebox(0,0)[bl]{tree \(S^{*}\)}}

\put(25,28){\circle*{1.3}}


\put(30,33){\circle*{1.3}}

\put(30,33){\line(1,1){5}}

\put(35,38){\circle*{1.3}}

\put(45,28){\circle*{1.3}}

\put(40,18){\circle*{1.3}}

\put(40,28){\circle*{1.3}} \put(40,28){\circle{2}}
\put(40,28){\line(1,0){5}} \put(40,28){\line(0,-1){10}}

\put(40,28){\line(-1,2){5}}

\put(36.4,27){\makebox(0,0)[bl]{\(a\)}}

\put(25,13){\circle*{1.3}}

\put(32.5,20){\circle*{1.3}} \put(32.5,20){\circle{2}}

\put(32.5,20){\line(3,-1){7.5}} \put(32.5,20){\line(-1,-1){7.5}}
\put(32.5,20){\line(-1,1){7.5}}

\put(34.4,20.6){\makebox(0,0)[bl]{\(c\)}}

\put(40,18){\line(-1,-2){5}}


\put(35,8){\circle*{1.3}}


\put(55,42.6){\makebox(0,0)[bl]{Steiner}}
\put(56,39.6){\makebox(0,0)[bl]{tree \(S^{2}\) }}

\put(50,28){\circle*{1.3}}

\put(50,28){\line(1,1){5}}


\put(55,33){\circle*{1.3}}

\put(60,38){\circle*{1.3}}

\put(60,33){\circle*{1.3}} \put(60,33){\circle{2}}
\put(60,33){\line(0,1){5}} \put(60,33){\line(-1,0){5}}

\put(60,33){\line(1,-1){5}}

\put(62,33){\makebox(0,0)[bl]{\(d\)}}

\put(70,28){\circle*{1.3}}

\put(65,28){\circle*{1.3}} \put(65,28){\circle{2}}
\put(65,28){\line(1,0){5}} \put(65,28){\line(0,-1){10}}

\put(61.4,27){\makebox(0,0)[bl]{\(a\)}}

\put(65,18){\circle*{1.3}}

\put(50,13){\circle*{1.3}}

\put(60,13){\circle*{1.3}} \put(60,13){\circle{2}}
\put(60,13){\line(-1,0){10}} \put(60,13){\line(0,-1){5}}

\put(60,13){\line(1,1){5}}

\put(59,15){\makebox(0,0)[bl]{\(b\)}}

\put(60,8){\circle*{1.3}}

\end{picture}
\end{center}

\section{Conclusion}

 In the paper,
 a restructuring approach in combinatorial optimization is
 examined.
 The restructuring problems are formulated as the following:
 (i) one-stage problem formulation
 (one-criterion statements, multicriteria statements),
 (ii) multi-stage problem formulation
 (one-criterion statements, multicriteria statements),
 The suggested restructuring approach is applied
 for several combinatorial optimization problems
 (e.g., knapsack problem, multiple choice problem,
 assignment problem, minimum spanning tree, Steiner tree problem,
 clustering problem, sorting problem).

 In the future,
 it may be prospective
  to consider the following research directions:

 {\it 1.} application of the suggested restructuring approach to
 other combinatorial optimization problems
 (e.g., covering, graph coloring);

 {\it 2.} examination of restructuring problems with changes of basic element sets
 (i.e., \(A^{1} \neq  A^{2}\));

 {\it 3.} study and usage of various types of proximity between
 obtained solution(s) and goal solution(s)
 (i.e., \( \rho (S^{*},S^{2})\));

  {\it 4.} examination of the restructuring problems under uncertainty
  (e.g., stochastic models, fuzzy sets based models, problems with multi-set based estimates);

 {\it 5.} further studies of dynamical restructuring problems
  including restructuring over changing set(s)
  (one-stage restructuring, multi-stage restructuring);

 {\it 6.} reformulation of restructuring problem(s)
 as satisfiability model(s);

 {\it 7.} analysis of restructuring problem(s)
 in case of changing the set of problem elements and/or their
 interconnection (i.e., while taking into account
 dynamical sets based methods, dynamical graph based methods);

 {\it 8.} usage of various AI techniques in solving procedures;
 and

 {\it 9.} application of the suggested restructuring approaches
  in engineering/CS/management education.

\section{Acknowledgments}

 The research materials presented in the article
 were partially supported by The Russian Foundation for
 Basic Research,
 project 15-07-01241
 ``Reconfiguration of Solutions in Combinatorial Optimization''
 (principal investigator: Mark Sh. Levin).



\begin{thebibliography}{350}

 \bibitem {adib10} M.A. Adibi, M. Zandieh, M. Amiri,
 Multi-objective scheduling of dynamic job shop using veriable
 neighborhood search.
 ESwA 37(1), 282--287, 2010.

  \bibitem {aiex05} R.M. Aiex, M.G.C. Resende,
  P.M. Pardalos, G. Toraldo,
  GRASP with path relinking for three-index assignment.
  INFORMS J. on Computing 17(2), 224--247, 2005. 

 \bibitem {albers03} S. Albers,
 Online algorithms: a survey.
 Math. Program. Ser. B 97(1-2), 3--26, 2003.

 \bibitem {albers99} S. Albers, S. Leonardi,
 On-line algorithms.
 ACM Comput. Surv. 31(3es) (4), 1999.

  \bibitem {alva12} R. Alvares-Valdes, F. Parreno, J.M. Tamarit,
  A GRASP/Path relinking algorithm for tw- and three-dimansional
  bin-size bin packing problems.
  Comp. and Oper. Res. 49(12), 3081--3090, 2012.

  \bibitem {amal11} E. Amaldi, G. Galbiati, F. Maffioli,
 On minimum reload cost paths, tours, and flows.
 Networks 57(3), 254--260, 2011.

  \bibitem {ames01} P.R. Amestoy, I.S. Duff,
  J.Y. L'Excellent, J. Koster,
 A fully asynchronous multifrontal solver using
 distributed dynamic scheduling.
 SIAM J. on Matrix Analysis and Applications
  23(1), 15--41, 2011.

  \bibitem {arain08} F.M. Arain,
   IT-based approach for effective management of project changes:
   A change management system (CMS).
   Advanced Engineering Informatics 22(4), 457--472, 2008.

 \bibitem {archetti03} C. Archetti, L. Bertazzi,
 M.G. Speranza,
 Reoptimizing the traveling salesman problem.
  Networks 42(3), 154-159, 2003.

  \bibitem {arch06} C. Archetti, L. Bertazzi, M.G. Speranza,
 Reoptimizing the 0-1 knapsack problem.
 Technical Report 267, University of Brescia, 2006.

 \bibitem {arch13} C. Archetti, G. Guastaroba, M.G. Speranza,
 Reoptimizing the rural postman problem.
 Comp. \& Oper. Res. 40(5), 1306--1313, 2013.

  \bibitem {ash97} G.R. Ash,
  Dynamic Routing in Telecommunications Networks.
  McGraw-Hill Professional, 1997.

 \bibitem {asch99} N. Ascheuer, M. Grotschel, S.O. Krumke,
 J. Rambau,
 Combinatorial online optimization. In:
 P. Kall, H.-J. Luthi (eds),
 Operations Research Proceedigns 1998,
 Springer, 21-37,  1999.

  \bibitem {aus09} G. Austello, B. Escoffer,
 J. Monnot, V. Paschos,
 Reoptimization of minimum and maximum traveling
 salesman's tours.
 J. of Discr. Algorithms 7(4), 453-463, 2009.

  \bibitem {ayd00} M.E. Aydin, E. Oztemel,
  Dynamic job-shop shceduling using reinforcement
  learning agents.
  Robotics and Autonomous Systems
  33(2), 169--178, 2000.

  \bibitem {aytug05} H. Aytug, M.A. Lawley, K. McKay, S. Mohan,
  R. Uzsoy,
  Execting production schedules in the face of
  uncertainty: A review and some future directions.
 EJOR 161(1), 86--110, 2005.

  \bibitem {berm81} O. Berman,
  Repositioning of distinguishable urban service
  units on networks.
   Comp. \& Oper. Res. 8(2), 105--118, 1981.

   \bibitem {bi08} Z.M. Bi, S. Lang, W. Shen, L. Wang,
  Reconfigurable manufacturing systems: the state of the art.
   Int. J. of Production Res. 46(4), 967--992, 2008.

 \bibitem {bilo08} D. Bilo, H.-J. Bockenhauer, J. Hromkovic,
 R. Kralovic, T. Momke, P. Widmayer, A. Zych,
 Reoptimization of Steiner trees.
 In:  J. Gudmundsson  (ed),
  Proc. of Scandinavian Workshop on Algorithm Theory
 SWAT'08, LNCS 5124, Springer,  258-269, 2008.

 \bibitem {bilo08a} D. Bilo, P. Widmayer, A. Zych,
 Reoptimization of weighted graph and covering problems.
 In: E. Bampis, M.  Skutella (eds),
 Proc. of 6th Int. Workshop on Approximation and
 Online Algorithms WAOA'08,
 LNCS 5426, Springer,  201-213, 2009.

 \bibitem {bilo11} D. Bilo, H.-J. Bockenhauer, D. Komm,
R. Kralovic, T. Momke, S. Seibert, A. Zych,
 Reoptimization of the Shortest Common Superstring Problem.
 Algorithmica 61(2), 227-–251, 2011.

 \bibitem {bock08a} H.-J. Bockenhauer, J.  Hromkovic,
  T. Momke, P. Widmayer,
 On the hardness of reoptimization.
 In: 34th International Conf. on Current Trends in Theory
 and Practice of Computer Science SOFSEM, 50–-65, 2008.

  \bibitem {bock09} S. Bocker, S. Briesemeister,
 Q.B.A. Bui, A. Truss,
 Going wieghted: Parametrized algorithms
 for cluster editing.
 Theor. Comp. Sci. 410(52), 5467--5480, 2009.

 \bibitem {bock11} S. Bocker, P. Damaschke,
 Even faster parameterized cluster deletion and cluster editing.
 Information Proceessing Letters 111(14), 717--721, 2011. 

 \bibitem {boj07} B. Bojduj, D. Taylor, F. Kurfess,
 Optimization of dyhnamic combinatorial optimization
 problems through truth maintenance. In:
 H.G. Okuno, M. Ali (eds), New Trends in Applied Artificial
 Intelligence,
  LNAI 4570, Springer, 984--991, 2007.

 \bibitem {bon02} K. Bondalapati, V.K. Prasana,
  Reconfigurable computing systems.
  Proc. of the IEEE 90(7), 1201--1217, 2002.

  \bibitem {boria10} N. Boria, V.P. Paschos,
 Fast reoptimization for the minimum spanning tree problem.
 J. of Discr. Algorithms 8(3), 296-310, 2010.

 \bibitem {borod98} A. Borodin, R. El-Yaniv,
 Online Computaiton and Competitive Analysis.
 Cambridge Univ. Press, New York, 1998.

 \bibitem {bose01} P. Bose, J. Czyzowicz, L. Gasieniec,
 E. Kranakis, D. Krizanc, A.  Pelc, M.V. Martin,
 Strategies for hotlink assignments.
  In:
  Int. Symp. on Algorithms and Computation (ISAAC'00),
 LNCS 1969, Springer,  23--34, 2001.

  \bibitem {brot03} L. Brot, G. Laporte, F. Semet,
 Ambulanve location and relocation models.
 EJOR 147(3), 451--463, 2003. 

  \bibitem {cas10} M. Caserta, S. Schwarze, S. Voss,
 Container rehandling at maritime container terminals.
 In:
 J.W. Bose (ed),
 Handbook of Terminal Planning.
 Operations Researcj/Computer Science
 Interfaces Series 49,
 Springer, 247--269, 2010.

   \bibitem {cas11} M. Caserta, S. Voss,
   M. Sniedovich
 Applying the corridor method
 to a blocks relocation problem.
 OE Spektrum, 33, 915--929, 2011.

   \bibitem {cas12} M. Caserta,  S. Schwarze, S. Voss,
 A mathematical formulation
 and complexity consideration
 for the blocks relocation problem.
 EJOR 219, 96--104, 2012.

 \bibitem {chaud94} B.B. Chaudhri,
 Dynamic clustering for time incremental data.
 Pattern Recognition Letters 13, 27--34, 1994.

 \bibitem {chenh98} H.-K. Chen, C.F. Hsueh,
 A model and an algorithm for the dynamic user-optimal route
 choice problem
 Transportation Research Part B:
 Methodological 32(3), 219--234, 1998.

 \bibitem {chen04} W.P. Chen, J.C. Hou, L. Sha,
 Dynamic clustering for acoustic target tracking in wireless
 sensor networks.
 IEEE Trans. on Mobile Computing
 3(3), 258--271, 2004.

 \bibitem {chur92}  L.K. Church, R. Uzsoy,
 Analysis of periodic and event-driven rescheduling policies in
 dynamic shops.
 Int. J. of Computer-Integrated Manufacturing
 5(3), 153--163, 1992.

 \bibitem {coello07} C. Coello Coello, G.B. Lamont, D.A. Van Veldhuizen,
 Evolutionary Algorithms for Solving Multi--Objective
 Problems.
 2nd ed., Springer, 2007.

  \bibitem {cow02} P. Cowling, M. Jojansson,
  Using real time information for effective
  dynamic scheduling.
  EJOR 139(2), 230--244, 2002.

 \bibitem {cox73}  D.C. Cox, D.O. Reudink,
 Increasing channel occupancy in large-scale mobile
 radio systems:
 Dynamic channel reassignment.
 IEEE Trans. on Vehicular Technology 22(4), 218--222, 1973.

 \bibitem {czy03} J. Czyzowicz, E. Kranakis, D. Krizanc,
 A. Pelc, M.V. Martin,
 Evaluation of hotlink structure for improving web performance.
 J. Web Eng. 1(2), 93--127, 2003.

  \bibitem {dam10} P. Damaschke, 
 Fixed parameter enumerability of
 cluster editing and related problems.
 Theory of Comput. Syst. 46, 261--283, 2010.

 \bibitem {de06}
 F. De Carvalho, R. De Souza,
 M. Chavent, Y. Lechevallier,
 Adaptative Hausdorff distances
 and dynamic clustering of symbolic interval data.
 Pattern Recog. Letters 27(3), 167--179, 2006.

 \bibitem {deb09} K. Deb,
 Multi-Objective Optimization Using Evolutionary Algorithms.
 Wiley, 2009.

  \bibitem {dehne06} F. Dehne, M.A. Langston, X. Luo, S. Pitre,
 P. Shaw, Y. Zhang,
 The cluster editing problem: implementations and experiemnts.
 In:
 H.L. Bodlaender, M.A. Langston (eds),
 Proc. IWPEC 2006,
 LNCS 4169, Springer, 13--24, 2006.

 \bibitem {dini08} G. Dini, M. Pelagatti, I.M. Savino,
  An algorithm for reconnecting wireless sensor netowkr
  partitions. In:
  Proc. 5th Eur. Conf. on Wireless Sensor Networks EWSN 2008,
  253--267, 2008.

 \bibitem {ehr10} M. Ehrgott,
 Multicriteria Optimization.
 2nd ed., Springer, 2010.

 \bibitem {ehr00} M. Ehrgott, X. Gandibleux,
 A survey and annotated bibliography of multiobjective
 combinatorial optimization.
 OR-Spektrum 22(4), 425--460, 2000.

 \bibitem {esc09} B. Escoffier, M. Milanic, V.T.  Paschos,
 Simple and fast reoptimizations for the Steiner tree problem.
 Algorithmic Oper. Res. 4(2), 86-–94, 2009.

 \bibitem {esw76} K.P. Eswaran, R.E. Tarjan,
 Augmentation problems.
 SIAM J. on Computing 5(4), 653--665, 1976.

 \bibitem {exp14} C. Exposito-Izquierdo, E. Lalla-Ruiz,
 B. Melian-Batista, J.M. Moreno-Vega
 A study of rescheduling strategies for the quay crane scheduling
 problem under random disruptions.
 Inteligencia Artificial 17(54), 35--47, 2014.

  \bibitem {far05}
  H. Farria, Jr., S. Binato, M.G.C. Resende, D.M. Falcao,
  Power transmission network design by greedy randomized adaptive
  path relinking.
  IEEE Trans. on Power Systems 20(1), 43-49, 2005.

  \bibitem {fortun10} S. Fortunato,
 Community detection in graphs.
  Electronic preprint, 103 p.,
 Jan. 25, 2010.
 http://arxiv.org/abs/0906.0612v2 [physics.soc-ph]

  \bibitem {fuhr01} S. Fuhrmann, S.O. Krumke,
  H.-C. Wirth,
 Multiple hotlink assignment.
 In: A. Brandstadt, V.B. Le (eds)
  Proc. of WG 2001, LNCS 2204,
 Springer,  189--200, 2001.

 \bibitem {gal95} G. Galambos, G.J. Woeginger,
 On-line bin-packing - a restricted survey.
 ZOR - Mathematical Methods of Operations Research
 42, 25--45, 1995.

 \bibitem {gam12} I. Gamvros, L. Gouveira, S. Raghavan,
 Reload cost trees and network design.
 Networks 59(4), 365--379, 2012.

 \bibitem {gar79} M.R. Garey, D.S. Johnson,
  Computers and Intractability. The
  Guide to the Theory of NP-Completeness.
  W.H. Freeman and Company, San Francisco, 1979.

 \bibitem {gham04} I. Ghamlouche, T.G. Crainic,
 M. Gendreau,
 Path relinking, cycle-based neighbourhoods and
 capacitated multicommodity network design.
 Annals of Oper. Res. 131(1-4), 109--133, 2004.

 \bibitem {glov00} F. Glover, M. Laguna, R. Marti,
 Fundamentals of scatter search and path relinking.
 Control and Cybernetics 39(3), 653--684, 2000.

  \bibitem {goz14} D. Gozupek, M. Shalom, A. Voloshin, S. Zaks,
 On the complexity of constructing minimum changeover cost
 arborescences.
 Theoretical Computer Science
 540-541, 40--52, 2014.

 \bibitem {goz13} D. Gozupek, S. Buhari, F> Alogoz,
 A spectrum switching delay-aware  scheduling algorithm
 for centralized cognitive radio networks.
 IEEE Trans. Mob. Comput. 12(7), 1270--1280, 2013.

  \bibitem {guna13} C. Gunasekara, K. Mehrotra, C.K. Mohan,
 Multi-objective restructuring in social networks.
 In: Proc. of The 2013 IEEE/ACM Int. Conf. on Advances in Social Network
 Analysis and Mining (ASONAM 2013), ACM, 277--281, 2013.

   \bibitem {guo09} J. Guo,
  A more effective linear kernelization for cluster editing.
  Theor. Comput. Sci. 410(8-10), 718--726, 2009.

 \bibitem {hen09} P. Van Hentenryck, R. Bent,
 Online Stochastic Combinatorial Optimization.
 The MIT Press, 2009.

 \bibitem {ho06} S.C. Ho, M. Gendreau,
  Path relinking for the vehicle routing problem.
  J. of Heuristics 12, 55--72, 2006.

  \bibitem {ho14} S.C. Ho, W.Y. Szeto,
  Solving a static repositioning problem in bike-sharing systems
  using iterated tabu search.
  Transp. Res. Part E
  69, 180--198, 2014.

  \bibitem {hopcroft04} J. Hopcroft, O. Khan,
  B. Kulis, B. Selman,
  Tracking evolving communities in large linked networks.
   PNAS 101(Suppl 1), 5249--5353, 2004.

  \bibitem {jain97} A.K. Jain, H.A. Elmaraghy,
 Produciton scheduling/rescheduling in flexible manufacturing.
 Int. J. of Production Research 35(1),
 281--309, 1997.

 \bibitem {jay14} J. Jayabharathy, S. Kanmani,
 Correlated concept based dynamic document clustering
 algorithms for newsgroups and scientific literature.
 Decision Analytics 1(3), 1-21, 2014.

 \bibitem {jov14} R. Jovanovic, S. Voss,
 A chain heuristic for the blocks relocation problem.
 Comp. and Ind. Eng. 75, 79--86, 2014.

 \bibitem {kho12} M.R. Khouadjia, B. Sarasola,
 E. Alba, L. Jourdan, E.G. Talbi,
 A comparative study between dynamic adapted pso and vns for the
 vehicle routing problem
 with dynamic requests.
 Applied Soft Computing 12(4), 1426--1439, 2012.

  \bibitem {kim06} K.H. Kim, G.-P. Hong,
 A heuristic rule for relocationing blocks.
 Com. and Oper. Res. 33, 940--954, 2005.

  \bibitem {kley98} A.J. Kleywegt, J.D. Papastavrou,
 he dynamic and stochastic knapsack problem.
 Operations Research 46(1), 17--35, 1998.

 \bibitem {kley01} A.J. Kleywegt, J.D. Papastavrou,
 he dynamic and stochastic knapsack problem
 with random sized items.
 Operations Research 49(1), 26--41, 2001.

 \bibitem {lag99} M. Laguna, R. Marti,
  GRASP and path relinking for 2-layer straight line crossing
  minimization.
  INFORMS J. on Computing 11(1), 44--52, 1999.

 \bibitem {lee01} J.-H. Lee, C.-D. Park, K.-Y. Chwa,
  An online relocation game on a graph.
   J. of Graph Algorithms and Applications
   5(5), 3--16, 2001.

  \bibitem{lev98} M.Sh. Levin,
   Combinatorial Engineering of Decomposable Systems.
  Kluwer Academic Publishers, Dordrecht, 1998.

 \bibitem{lev06} M.Sh. Levin,
  Composite Systems Decisions. Springer, New York, 2006.

 \bibitem {lev09}  M.Sh. Levin,
 Combinatorial optimization in  system configuration design.
  Automation and Remote Control 70(3), 519--561, 2009.

   \bibitem {lev10a} M.Sh. Levin,
 Towards communication network development
 (structural system issues, combinatorial models).
 In:  2010 IEEE Region 8 Int. Conf. SIBIRCON-2010,
  vol. 1, 204--208, 2010.

 \bibitem {lev11restr} M.Sh. Levin,
 Restructuring in combinatorial optimization.
 Electronic preprint, 11 p.,
 Febr. 8, 2011.
 http://arxiv.org/abs/1102.1745 [cs.DS]

  \bibitem {lev12a} M.Sh. Levin,
   Multiset estimates and combinatorial synthesis.
   Electronic preprint. 30 pp., May 9, 2012.
   http://arxiv.org/abs/1205.2046 [cs.SY]

  \bibitem {lev12hier} M.Sh. Levin,
  Towards design of system hierarchy (research survey).
    Electronic preprint. 36 p., Dec. 7, 2012.
    http://arxiv.org/abs/1212.1735 [math.OC]

  \bibitem {lev12clique} M.Sh. Levin,
   Clique-based fusion of graph streams in multi-function system testing.
    Informatica 23(3), 391--404, 2012.

   \bibitem {lev13tra} M.Sh. Levin,
  Towards multistage design of modular systems.
 Electronic preprint, 13 p., June 19, 2013.
 http://arxiv.org/abs/1306.4635 [cs.AI].

 \bibitem {lev15} M.Sh. Levin,
 Modular System Design and Evaluation, Sprigner, 2015.

 \bibitem {lev15c} M.Sh. Levin,
 Towards combinatorial clustering:
 preliminary research survey.
 Electr. prepr., 102 p., May 28, 2015.
 http://arxiv.org/abs/1505.07872 [cs.AI]

 \bibitem {lev15route} M.Sh. Levin,
 Discrete route/trajectory decision making problems.
 Electr. prepr., 25 p., Aug. 16, 2015.
 http://arxiv.org/abs/1508.03863 [cs.AI]

  \bibitem {levdan05} M.Sh. Levin, M.A. Danieli,
  Hierarchical decision making framework for evaluation and improvement
  of composite systems
  (example for building).
  Informatica (LI) 16(2), 213--240, 2005.

  \bibitem {levfim10} M.Sh. Levin, A.V. Fimin,
  Configuration of alarm wireless sensor  element.
  In:
  Proc. of 2nd Int. Conf. on Ultra Modern Telecommunication ICUMT-2010,
   Moscow,
 924--928, 2010.

   \bibitem {levpet10} M.Sh. Levin, M. Petukhov,
  Multicriteria assignment problem (selection of access points).
  In:
  Proc. of
  IEA/AIE 2010,
  LNCS 6097, part II, Springer, Cordoba, Spain, 277--287, 2010.

 \bibitem {levsaf11} M.Sh. Levin, A.V. Safonov,
 Improvement of regional telecommunications networks.
 J. of Communications Technology and Electronics
 56(6), 770--778, 2011.

 \bibitem {levand11} M.Sh. Levin, A. Andrushevich,
 A. Klapproth,
  Improvement of building automation system.
 In:  Proc. of 24th Int. Conf.  IEA/AIE 2011, LNCS 6704, part II,
  Springer, Heidelberg,
  459--468, 2011.

 \bibitem {li93} R.-K. Li, Y.-T. Shyu, S. Adiga,
 A heuristic rescheduling algorithm for computer-based production
 scheduling systems.
 Int. J. Prod. Res. 31, 1815--1826, 1993.

  \bibitem {lima11} K.R. Lima, Y. Wakabayashi,
 Convex recoloring of paths.
 Electr. Notes in Disc. Math. 37, 165--170, 2011.

 \bibitem {lod10}  W.A. Lodwick (ed),
 Fuzzy Optimizaiton: Recent Advances and Applications.
 Springer, 2010.

  \bibitem {man98} G. Manimaran, C. Murthy,
  An efficient dynamic scheduling algorithm for
  multiprocessor real-time systems.
 IEEE Trans. on Parallel and Distributed Systems
 9(3), 312--319, 1998.

  \bibitem {mann10} B. Mannaa,
 Cluster editing problem for points on the real line:
 A polynomial time algorithm.
 Inform. Process. Lett. 110(21), 961--965, 2010.

 \bibitem {mas13} R. Masson, T. Vidal, J. Michallet,
 P.H.W. Penna, V. Petrucci, A. Subramanian, H. Dubedout,
 An iterated local search heuristic for multi-capacity
 bin-packing and machine reassignment.
 ESwA 40(13), 5266--5275, 2013.

  \bibitem {mcdon99} A.B. McDonald, T.F. Znati,
 A mobnility-based framework for adaptive clustering
 in wireless ad hoc network.
 IEEE J. on Selected Areas in Communications 17(8),
 1466--1487, 1999.

   \bibitem {mel00} E. Melachrinoudis and H. Min,
  The dynamic relocation and phase-out of a hybrid,
  two-eshelon plant / warehousing facility:
  A multiple objective approach.
  EJOR 123(1), 1--15, 2000.

 \bibitem {mes12} M. Mestria, L.S. Ochi, S. de Lima Martins,
 GRASP with path relinking for the symmetric Euclidean clustered
 traveling salesman problem.
 Comp. and Oper. Res. 40(12), 3218--3229, 2012.

  \bibitem {mirkin79} B.G. Mirkin,
  Group choice. Halsted Press, New York, 1979.

  \bibitem {moraes12} R.E.N. Moraes, C.C. Ribeiro,
 Power optimization in ad hoc wireless network topology control
 with biconnectivity requirements.
 Comp. and Oper. Res. 40(12), 3188--3196, 2012.

  \bibitem {moran08} S. Moran, S. Snir,
 Convex recoloring of strings and trees:
 definitions, hardness results and algorithms.
 J. Comput. System Sci. 74(5), 850--869, 2008.

 \bibitem {mor04} R.W. Morrison,
 Designing Evolutionary Algorithms for Dynamic Environments.
 Berlin, Springer, 2004.

  \bibitem {nolt99} H. Noltmeier, H.-C.  Wirth,
 S.O.  Krumke,
  Network design and improvement.
  ACM Computing Surveys  32(3es), Art. no. 2, 1999.

 \bibitem {omran06} M.G. Omran, A. Salman, A.P. Engelbrecht,
 Dynamic clusterign using particle swarm optimizaiton with
 applicaiton in image segmentation.
 Pattern Analysis and Applications
 8(4), 332--344, 2006.


 \bibitem {ou09} D. Ouelhadj, S. Petrovic,
 A survey of dynamic scheduling in manufacturign systems.
 J. of Scheduling 12(4), 417--431, 2009.

 \bibitem {papa08} A. Papadogiannis, D. Gesbeert, E. Hardouin,
 A dynamic clusterign approach in wireless networks
 with multi-cell cooperative processing.
 In: IEEE Int. Conf. on Communicaitons ICC'08, 4033--4037, 2008.

 \bibitem {pard13} P.M. Pardalos, E.K. Aydogan, F. Gurbuz,
 O. Demirtas, B.B. Bakiri,
 Fuzzy combinatorial optimization problems. In:
 P.M. Pardalos, D.-Z. Du, R.L. Graham (eds),
 Handbook of Combinatorial Optimization, Springer,
 1357--1413, 2013.

  \bibitem {pareto71} V. Pareto,
  Mannual of Political Economy.
 (English translation), A. M. Kelley Publishers,
 New York, 1971.

 \bibitem {ped12} O. Pedrola, M. Ruiz, L. Velasco, D. Careglio,
 O.G. de Dios, J. Comellas,
 A GRASP with path relinking heuristic for the
 IP/MPLS-over-WSON multi-layer network optimization
 problem.
 Comp. and Oper. Res. 40(12), 3174--3187, 2012.

 \bibitem {pet13} M.E. Petering, M. Hussein,
 A new mixed integer program and extended look-ahead
 heuristic algorithm
 for the block relocation problem.
 EJOR 231, 120--130, 2013.

 \bibitem {pollock92} L.L. Pollock, M.L. Soffa,
 Incremental global reoptimization of programs.
 ACM Trans. on Programming Languages and Systems
  14(2), 173--200, 1992.

 \bibitem {qui11} J. Qiu, Y. Liu, G. Mohan, K.C. Chua,
 Fast spanning tree reconnection mechanism for resilient Metro
 Ethernet networks.
 Computer Networks 55(00), 2717–-2729, 2011.

  \bibitem {rah07} S. Rahmann, T. Wittkop, J. Baumbach,
  M. Martin, A. Truss, S. Bocker,
  Exact and heuristic algorithms for weighted cluster editing.
  In: Proc. Comput. Syst.  Bioinform. Conf.,
   6(1), 391--401, 2007.

 \bibitem {ran04} R. Rangsaritratsamee, W. Ferrell Jr., M. Kurz,
 Dynamic rescheduling that similtaneously considers efficiency and
 stability.
 Computers and Industrial Engineering 46(1), 1--15, 2004.

 \bibitem {reeves98}  C.R. Reeves, T. Yamada,
 Genetic algorithms, path relinking, and
 flowshop sequencing problem.
 Evolutionary Computation 6(1), 45--60, 1998.

  \bibitem {res05}  M.G.C. Resende, C.C. Ribeiro,
  GRASP with path relinking:
  Recent advances and applicaitons.
  In:
  Metaheuristics: Progress as Real Problem Solvers.
  Springer, 29--63, 2006.

 \bibitem {roh09} P. Rohlfshagen, X. Yao,
 The dynamic knapsack problem revised:
 A new benchmark problem for dynamic combinatorial optimisation.
 In:
 Applications of Evolutionary Computing, LNCS 5484, Springer,
 745--754, 2009.

  \bibitem {roy96} B. Roy,
 Multicriteria Methodology for Decision Aiding.
  Kluwer,
  Dordrecht, 1996.

 \bibitem {savel98} M. Savelsbergh, M. Sol,
 DRIVE: Dynamic routing of independent vehicles.
 Operations Research 46(4), 474--490, 1998.

 \bibitem {scha97} M.W. Schaffter,
 Scheduling with forbidden sets.
 Discrete Appl. Math. 72(1–2), 155-–166, 1997.

 \bibitem {sch15} F. Schulte, S. Voss,
 Decision support for environmental-friendly vehicle relocations
 in free-floating car sharing systems:
 The case of car2go.
 Procedia CIRP 30, 275--280, 2015.

  \bibitem {sham04} R. Shamir, R. Sharan, D. Tsur,
 Cluster graph modification problems.
 Disc. Appl. Math. 144(1-2), 173--182, 2004.

  \bibitem {sor12} K. Sorensen, P. Schittekat,
 Statistical analysis of distance-based path relinking
 for the capacitated vehicle routing problem.
 Comp. and Oper. Res. 40(12), 3197--3205, 2012.

 \bibitem {souf10} W. Souffriau, P. Vansteenwegen, G. Vanden
 Berghe, D. Van Oudheusden,
 A path relinking approach for the team orienteering problem.
 Comp. and Oper. Res. 37(11), 1853--1859, 2010.

 \bibitem {talbi09} E.-G. Talbi,
 Metaheuristics: From Design to Implementation.
 Wiley, 2009.

  \bibitem {tk06} V. T'Kindt, J.-C. Billaut,
 Vincent T'Kindt, Jean-Charles Billaut,
 Multicriteria Scheduling: Theory, Models and Algorithms.
 2nd ed., Springer, 2006.

 \bibitem {usb12} F.L. Usberti, P.M. Franca, A.L.M. Franca,
 GRASP with evolutionary path-relinking
 for the capacitated arc routing problem.
 Comp. and Oper. Res. 40(12), 3206--3217, 2012.

  \bibitem {val10} E. Vallada, R. Ruiz,
 Genetic algorothms with path relinking for the
 minimum tardness permutation flowshop problem.
 Omega 38(1), 57--67, 2010.

  \bibitem {vier03} G.E. Viera, J.W. Herrmann, E. Lin,
 Rescheduling manufacturing systems: a framework of strategies,
 policies, and methods.
 J. of Heuristics 6, 39--62, 2003.

  \bibitem {vos96} B. Vos, H. Akkermans,
  Capturing the dynamics of facility allocation.
  Int. J. of Operations \& Production Management
  16(11), 57--70, 1996.

  \bibitem {wang05} G. Wang, G. Cao, T.L. Porta, W. Zang,
 Sensor realocation in mobile sensor network.
 In: Proc. IEEE 24th
 Annu. Conf. of the Comp. and Commun. Soc.
 INFOCOM 2005,
 vol. 4, 2302--2312, 2005.

 \bibitem {wangz16} Z. Wang, Z. Lu, T. Ye,
 Multi-neighborhood local search optimization
 for machine reassignment problem.
 Comp. and Oper. Res. 68, 16--29, 2016.

   \bibitem {wirth01} H. Wirth, J. Steffan,
 Reload cost problems:
 Minimum diameter spanning trees.
 Discr. Appl. Math. 113(1), 73--85, 2001.

  \bibitem {wzorek06} M. Wzorek, P. Doherty,
 Reconfigurable path planning for an
 autonomous unmanned aerial vehicle.
 In:
 Int. Conf. on Hybrid Information Technology
 ICHIT'06, vol. 2, 242--249, 2006.

  \bibitem {yag06} M. Yagiura, T. Ibaraaki, F. Glover,
 A path relinking approach with ehection chains
 for the generalized assignment problem.
 EJOR 169(2), 548--569, 2006.

 \bibitem {yang07} S. Yang, Y.-S. Ong, Y. Jin (eds),
 Evolutionary Computation in Dynamic and Uncertaint Environments.
 Berlin, Springer, 2007.

 \bibitem {yang12} S. Yang, Y. Jiang, T.T. Nguyen,
 Metaheuristics for dynamic combinatorial optimizaiton problems.
 IMA J. of Management Mathematics 24, 451--480, 2013.

   \bibitem {yem08} O.A. Yemets, A.A. Roskladka,
 Combinatorial optimization under uncertainty.
 Cybernetics and Systems Analysis
 44(5), 655--663, 2008.

 \bibitem {yu07} M. Yu, K.K. Leung, A. Malvankar,
 A dynamic clustering and energy efficient routing technique
 for sensor networks.
 IEEE Trans. on Wireless Commmunications 6(8), 3069--3079, 2007.

 \bibitem {zeh15} E. Zehendner, M. Caserta, D. Feillet,
  S. Schwarze, S. Voss,
 An improved mathematical formulation for
 the blocks relocation problem.
 EJOR 245, 415--422, 2015.

 \bibitem {zhu12} W. Zhu, H. Qin, A. Lim,
 H. Zhang,
 Iterative deeping A' algorithm for the
 container relocation problem.
 IEEE Trans. on Automatic Science and Engineering
  9, 710--722, 2012.

  \bibitem {zop02} C. Zopounidis, M. Doumpos,
 Multicriteria classification and sorting methods:
 a literature review.
  EJOR
 138(2), 229--246, 2002.

 \bibitem {zweb93} M. Zweben, E. Davis, B. Daun, M.J. Deale,
 Scheduling and rescheduling with iterative repair.
 IEEE Trans. SMC, 23(6), 1588--1596, Dec. 1993.

\end{thebibliography}
\end{document}